 \documentclass[3p,times]{elsarticle}  

\journal{}

\usepackage{times}
\usepackage{helvet}
\usepackage{courier}

\usepackage[utf8]{inputenc}
\usepackage[T1]{fontenc}
\usepackage[english]{babel}

\usepackage{amsmath}
\usepackage{amssymb}
\usepackage{amsthm}           %
\usepackage{stmaryrd}
\usepackage[usenames,dvipsnames,svgnames,table]{xcolor}

\usepackage{xparse}
\usepackage[strict]{csquotes}
\usepackage{booktabs} %
\usepackage{tabularx} %
\usepackage{multicol} %
\usepackage{multirow} %
 \usepackage{rotating}
\usepackage{enumerate}
\usepackage{soul}
\usepackage{pifont} %
\usepackage{xifthen}
\usepackage{diagbox}

\usepackage{esvect}         %
\usepackage[kerning,spacing,tracking,expansion]{microtype}

\usepackage{graphicx}
\usepackage{tikz}
\usetikzlibrary{fadings}
\usetikzlibrary{shapes,decorations,positioning}
\usetikzlibrary{fadings,shapes,decorations,positioning,calc}
\usetikzlibrary{decorations.markings,decorations.pathreplacing}
\usetikzlibrary{shapes.geometric,automata,positioning}
\usetikzlibrary{shadows}\usetikzlibrary{calc}
\usetikzlibrary{decorations,decorations.pathmorphing,decorations.pathreplacing}
\usetikzlibrary{calc}
\usepackage{xspace}
\usepackage{array}
\usepackage[dvipsnames]{xcolor}

\usepackage{microtype}
\usepackage[list=true, labelformat=brace, position=top]{subcaption}

\usepackage{tocbasic}
\DeclareTOCStyleEntry[
  beforeskip=0.3em plus 1pt,
  pagenumberformat=\textbf
]{tocline}{section}

\usepackage[hidelinks]{hyperref}   

\newif\ifhideproofs

\ifhideproofs
\usepackage{environ}
\NewEnviron{hide}{}

\fi

\newcommand{\FARBE}[1]{#1}

\makeatletter
\newcommand{\textlabelmarker}[1]{%
	\protected@edef\@currentlabel{#1}%
	\phantomsection%
}
\newcommand{\textlabel}[2]{%
	\textlabelmarker{#1}%
	#1\label{#2}%
}
\makeatother

\newcommand{\cL}{\ensuremath{{\mathcal L}}\xspace}

\newcommand{\ol}[1]{\ensuremath{\overline{#1}}\xspace}

\renewcommand{\emptyset}{\varnothing}
\renewcommand{\epsilon}{\varepsilon}

\renewcommand{\leq}{\leqslant}

\renewcommand{\geq}{\geqslant}

\newcommand{\notA}{\overline{A}}

\newcommand{\notC}{\overline{C}}
\newcommand{\notB}{\overline{B}}
\newcommand{\nota}{\overline{a}}

\newcommand{\notb}{\overline{b}}
\newcommand{\condL}{\mbox{$(\cL \mid \cL)$}}

\newcommand{\Cn}{\textit{Cn}}

\makeatletter
\ifx\centernot\undefined
\newcommand*{\centernot}{%
	\mathpalette\@centernot
}
\fi
\def\@centernot#1#2{%
	\mathrel{%
		\rlap{%
			\settowidth\dimen@{$\m@th#1{#2}$}%
			\kern.5\dimen@
			\settowidth\dimen@{$\m@th#1=$}%
			\kern-.5\dimen@
			$\m@th#1\not$%
		}%
		{#2}%
	}%
}
\makeatother

\newcommand{\Th}{{\it Th}}

\newcommand{\Mod}{\mbox{\it Mod}\,}
\newcommand{\Bel}{\mbox{\it Bel}\,}

\newcommand{\naturals}{\mathbb{N}}

\newcommand{\defremarkblau}[2][\empty]{
 \ifthenelse{\equal{#1}{\empty}}
	{\begin{center}
	 \fcolorbox[named]{Blue}{White}{#2}
	 \end{center}}
	{\fcolorbox[named]{Blue}{White}{#2}}
}

\newcommand{\WEG}[1]{}

\newcommand{\kappaplus}{\gamma^+}
\newcommand{\kappaminus}{\gamma^-}

\makeatletter
\newcommand{\leqnomode}{\tagsleft@true\let\veqno\@@leqno}
\newcommand{\reqnomode}{\tagsleft@false\let\veqno\@@eqno}
\makeatother

\DeclareRobustCommand\nmmodelsSymb{\mathrel|\mkern-.5mu\joinrel\approx} %
\newcommand{\nmmodels}{\ensuremath{\nmmodelsSymb}} %
\newcommand{\notnmmodels}{\ensuremath{\centernot\nmmodelsSymb}} %

\newcommand{\ESmodels}{\ensuremath{\models}} %
\newcommand{\notESmodels}{\ensuremath{\not\models}} %

\newcommand{\selectOpIgnore}{\ensuremath{-_{\textit{ign}}}}
\newcommand{\strategyIgnore}{\ensuremath{\selectionStrategyContractProp_{\textit{ign}}}}

\newcommand{\selectOpRevocation}{\ensuremath{-_{\textit{rev}}}}
\newcommand{\strategyRevocation}{\ensuremath{\selectionStrategyContractProp_{\textit{rev}}}}

\newcommand{\selectOpMinimal}{\ensuremath{-_{\textit{min}}}}
\newcommand{\strategyMinimal}{\ensuremath{\selectionStrategyContractProp_{\textit{min}}}}

\newcommand{\selectOpNonMinimal}{\ensuremath{-_{\textit{non-min}}}}
\newcommand{\strategyNonMinimal}{\ensuremath{\selectionStrategyContractProp_{\textit{non-min}}}}
\newcommand{\cspContractProp}{\ensuremath{\cspContractPropof{\kappa}{A}}}

\newcommand{\cspContractPropof}[2]{\ensuremath{\mathit{CR}^{-}_{P}(#1,#2)}}

\newcommand{\solutionsOf}[1]{\ensuremath{\mathit{Sol}(#1)}}

\newcommand{\solutionsContractProp}{\ensuremath{\solutionsOf{\cspContractProp}}}

\newcommand{\kappaCirc}{\ensuremath{\kappa^{\circ}}}   %

\newcommand{\induzierteContractProp}[1]{\ensuremath{\kappaCirc_{#1}}}

\newcommand{\selectionStrategyContractProp}{\ensuremath{\sigma}}

\newcommand{\postSyntaxIndependent}{(\textbf{SI})}  %
\newcommand{\postMinimal}{(\textbf{Min})}  %
\newcommand{\postNonMinimal}{(\textbf{NonMin})}  %
\newcommand{\postIgnore}{(\textbf{Ignore})}  %
\newcommand{\postRevoke}{(\textbf{Revoke})}  %

\newcommand{\margsub}[2]{{#1}_{\downarrow #2}}
\newcommand{\embedsup}[2]{\ensuremath{{#1}_{\uparrow #2}}}
\newcommand{\margANDembed}[3]{\ensuremath{{{#1}_{\downarrow #2\uparrow #3}}}}

\newcommand{\embedANDembed}[3]{\ensuremath{{#1}_{\uparrow #2\uparrow #3}}}
\newcommand{\margANDmarg}[3]{\ensuremath{{#1}_{\downarrow #2\downarrow #3}}}
\newcommand{\margANDembedANDmarg}[4]{\ensuremath{{#1}_{\downarrow #2\uparrow #3\downarrow #4}}}
\newcommand{\margANDembedANDembed}[4]{\ensuremath{{#1}_{\downarrow #2\uparrow #3\uparrow #4}}}
\newcommand{\margANDembedANDmargANDembed}[5]{\ensuremath{{#1}_{\downarrow #2\uparrow #3\downarrow #4\uparrow #5}}}

\newcommand{\subseteqCNsig}[1]{\ensuremath{\subseteq_{#1}}}
\newcommand{\gleichCNsig}[1]{\ensuremath{=_{#1}}}

\newcommand{\cupdisjoint}{\dot{\cup}}

\newcommand{\revocation}{revocation}
\newcommand{\Revocation}{Revocation}

\newcommand{\xequiv}{\cong_\ell}
\newcommand{\postxE}{(Mult)}  %
\newcommand{\contract}[2]{\mbox{$#1\! -\! #2$}}
\renewcommand{\kappaCirc}{\ensuremath{\kappa^{-}}}

\newcounter{ocf@count}%
\newcounter{ocf@level}[ocf@count]%
\newcounter{ocftable@row}%
\newcounter{ocftable@col}[ocftable@row]%
\makeatletter%
\AtBeginDocument{%
	\@ifpackageloaded{array}{		
		\newcommand{\ocftable@tabular}{%
			\expandafter\newcolumntype%
				\expandafter{\expandafter Z\expandafter}%
					\expandafter{%
						\ocftable@coltype%
					}%
			\begin{tabular}{c*{\theocf@count}{|Z|c}}%
				\ocftable@clines%
				\ocftable@content%
			\end{tabular}%
		}%
	}{%
		\newcommand{\ocftable@tabular}{%
			\begin{tabular}{c*{\theocf@count}{|\ocftable@coltype|c}}%
				\ocftable@clines%
				\ocftable@content%
			\end{tabular}%
		}%
	}%
}%
\newenvironment*{ocftable}[2][c]{%
	\newcommand{\ocftable@coltype}{#1}
	\newcommand{\ocf@levels}{#2}%
	\setcounter{ocf@count}{0}%
	\setcounter{ocf@level}{0}%
	\newcommand*{\ocf@setlabel}[2]{%
		\expandafter\edef\csname ocf@label:##1\endcsname{##2}%
	}%
	\newcommand*{\ocf@getlabel}[1]{%
		\expandafter\csname ocf@label:##1\endcsname%
	}%
	\newcommand*{\ocf@setvalue}[3]{%
		\expandafter\edef\csname ocf@value:##1:##2\endcsname{##3}%
	}%
	\newcommand*{\ocf@getvalue}[2]{%
		\expandafter\csname ocf@value:##1:##2\endcsname%
	}%
	\newcommand*{\ocf}[2]{%
		\ocf@setlabel{\theocf@count}{##1}%
		\@for\next:=##2\do{%
			\ocf@setvalue{\theocf@count}{\theocf@level}{\next}%
			\stepcounter{ocf@level}%
		}%
		\stepcounter{ocf@count}%
	}%
}{%
	\setcounter{ocftable@row}{0}%
	\setcounter{ocftable@col}{0}%
	\def\ocftable@colspecs{c}%
	\setcounter{ocftable@col}{0}%
	\loop%
		\ifnum\theocftable@col<\theocf@count%
		\g@addto@macro\ocftable@colspecs{|\ocftable@coltype|c}%
		\stepcounter{ocftable@col}%
	\repeat%
	\def\ocftable@clines{}%
	\setcounter{ocftable@col}{0}%
	\loop%
		\ifnum\theocftable@col<\theocf@count%
		\stepcounter{ocftable@col}%
		\edef\clinei{\the\numexpr2*\theocftable@col\relax}%
		\edef\clinei{\clinei-\clinei}%
		\expandafter\g@addto@macro%
			\expandafter\ocftable@clines%
				\expandafter{%
					\expandafter\cline%
						\expandafter{%
							\clinei%
						}%
				}%
	\repeat%
	\def\ocftable@content{}%
	\def\ocftable@content@append##1{%
		\expandafter\g@addto@macro%
			\expandafter\ocftable@content%
				\expandafter{%
					##1%
				}%
	}%
	\setcounter{ocftable@row}{0}%
	\setcounter{ocftable@col}{0}%
	\loop%
		\ifnum\theocftable@row<\ocf@levels%
		\edef\rowlabel{$\scriptstyle{\the\numexpr\ocf@levels-\theocftable@row-1\relax}$\enspace}%
		\ocftable@content@append{\rowlabel}%
		{\loop%
			\ifnum\theocftable@col<\theocf@count%
			\edef\cellcontents{&\ocf@getvalue{\theocftable@col}{\theocftable@row}&}%
			\ocftable@content@append{\cellcontents}%
			\stepcounter{ocftable@col}%
		\repeat}%
		\g@addto@macro\ocftable@content{\\ \ocftable@clines}%
		\stepcounter{ocftable@row}%
	\repeat%
	\g@addto@macro\ocftable@content{\multicolumn{1}{c}{}}%
	\setcounter{ocftable@col}{0}%
	\loop\ifnum\theocftable@col<\theocf@count%
		\edef\ocflabel{\ocf@getlabel{\theocftable@col}}%
		\edef\cellcontents{%
			&\unexpanded{\multicolumn{1}{c}}{\ocflabel}%
			&\unexpanded{\multicolumn{1}{c}{}}%
		}%
		\ocftable@content@append{\cellcontents}%
		\stepcounter{ocftable@col}%
	\repeat%
	\ocftable@tabular
}%
\makeatother %
\let\Mod\undefined
\DeclareMathOperator{\Mod}{Mod}

\newcommand{\beliefssymbol}{\mathrm{Bel}}
\newcommand{\beliefsOf}[1]{\beliefssymbol(#1)} %
\newcommand{\modelsOf}[1]{\Mod(#1)} %
\newcommand{\modBelOf}[1]{\ensuremath{\modelsOf{\Bel(#1)}}} %
\newcommand{\modBeliefsOf}[1]{\ensuremath{\modelsOf{\Bel(#1)}}} %
\newcommand{\ModBelOf}[1]{\ensuremath{\modBelOf{#1}}} %
\newcommand{\forgetBy}[2]{{#1}^\circ_{\mkern-2mu{#2}}}
\newcommand{\condOf}[1]{\Cons(#1)}

\newcommand{\sigmaprime}{\ensuremath{\Sigma'}}

\newcommand{\propWCSigma}{{(wC\( {}_{\Sigma} \))}\xspace}
\newcommand{\propSCSigma}{{(sC\( {}_{\Sigma} \))}\xspace}
\newcommand{\propCPSigma}{{(CP\(_\Sigma \))}\xspace}

\newcommand{\propW}{{(W)}\xspace}
\newcommand{\propWC}{{(wC)}\xspace}
\newcommand{\propSC}{{(sC)}\xspace}
\newcommand{\propCP}{{(CP)}\xspace}
\newcommand{\propWE}{{(wE)}\xspace}
\newcommand{\propE}{{(E)}\xspace}
\newcommand{\propxE}{{(LE$^\text{ocf}$)}\xspace}
\newcommand{\propBE}{{(BE)}\xspace}
\newcommand{\propEBE}{{(EBE)}\xspace}
\newcommand{\propDE}{\propEBE} %

\newcommand{\propOI}{{(OI)}\xspace}
\newcommand{\propPP}{{(PP)}\xspace}

\newcommand{\propNP}{{(NP)}\xspace}
\newcommand{\propEP}{{(EP)}\xspace}
\newcommand{\propBP}{{(BP)}\xspace}

\newcommand{\propPPcond}{{(PP$^{cond}$)}\xspace}
\newcommand{\propNPcond}{{(NP$^{cond}$)}\xspace}
\newcommand{\propCF}{{(CF)}\xspace}

\newcommand{\Cons}{C}

\newcommand{\xmark}{\ding{55}}%

\newcommand{\agm}[1]{{(AGMes\kern.05em\raisebox{.7ex}{\small$\circ$}\kern-.05em#1)}}

\newcommand{\ignore}{{ignoration}\xspace}
\newcommand{\Ignore}{{Ig\-nor\-ation}\xspace}

\newcommand{\ocfMarg}{OCF-mar\-gin\-al\-ization\xspace}
\newcommand{\ocfCond}{forgetting by OCF-con\-di\-tion\-al\-ization\xspace}
\newcommand{\OcfCond}{Forgetting by OCF-con\-di\-tion\-al\-ization\xspace}

\newcommand{\signatureOf}[1]{\mathsf{Sig}(#1)}
\newcommand{\signatureMinOf}[1]{\ensuremath{\mathsf{Sig}_{\mathit{\min}}(#1)}}

\newcounter{thm}
\newtheorem{lemma}[thm]{Lemma}
\newtheorem{proposition}[thm]{Proposition}
\newtheorem{example}[thm]{Example}
\newtheorem{corollary}[thm]{Corollary}
\newtheorem{definition}[thm]{Definition}

{\bfseries}{\itshape}

\numberwithin{thm}{section}

\DeclareMathOperator{\varForgetSymb}{VarElem}
\newcommand{\varForget}[2]{\varForgetSymb(#1,#2)}

\DeclareMathOperator{\semmargSymb}{ModMg}
\DeclareMathOperator{\synmargSymb}{SynMg}
\DeclareMathOperator{\ewmargSymb}{EWSynMg}
\DeclareMathOperator{\bsmargSymb}{SynMg}
\DeclareDocumentCommand{\semmarg}{g m m}{\semmargSymb_{#1}(#3,#2)}
\DeclareDocumentCommand{\synmarg}{g m m}{\synmargSymb_{#1}(#3,#2)}
\DeclareDocumentCommand{\ewmarg}{g m m}{\ewmargSymb_{#1}(#3,#2)}
\DeclareDocumentCommand{\bsmarg}{g m m}{\bsmargSymb_{#1}(#3,#2)}

\newcommand{\postulateskip}{4.5ex}
\newcommand\totheright[1]{{%
        \unskip\nobreak\hfil\penalty50
        \hskip2em\hbox{}\nobreak\hfil\textbf{#1}%
        \parfillskip=0pt \finalhyphendemerits=0 \par}}

\newcommand{\setzePostulat}[4]{\item[]\hspace{-5ex}\begin{minipage}{\postulateskip}\raggedright
        #1
    \end{minipage}
    #3
    \totheright{\textbf{(#2)}}%
    \ifthenelse{\isempty{#4}}{}{\leavevmode\emph{Explanation:} #4}}

\newenvironment{postulate}{\begin{itemize}}{\end{itemize}}

\begin{document}

\begin{frontmatter}

\title{A General Framework of Epistemic Forgetting and its Instantiation by Ranking~Functions}

\author[ha]{Christoph Beierle}
\author[do]{Alexander Hahn}
\author[do]{Diana Howey}
\author[do]{Gabriele Kern-Isberner}
\author[ha]{Kai Sauerwald}

\address[ha]{Faculty of Mathematics and Computer Science,  University of Hagen, 58084 Hagen, Germany}
\address[do]{Department of Computer Science, Technology Dortmund University, 44221 Dortmund, Germany}

\begin{abstract}
Forgetting as a knowledge management operation deliberately ignores parts of the knowledge and beliefs of an agent, for various reasons. Forgetting has many facets, one may want to forget parts of the syntax, a proposition, or a conditional. In the literature, two main operators suitable for performing forgetting have been proposed and investigated in depth: First, variable elimination is a syntactical method that blends out certain atomic variables to focus on the rest of the language. It has been mainly used in the area of logic programming and answer set programming. Second, contraction in AGM belief revision theory effectively removes propositions from belief sets under logical deduction. Both operations rely mainly on classical logics. In this article, we take an epistemic perspective and study forgetting operations in epistemic states with richer semantic structures, but with clear links to propositional logic. This allows us to investigate what forgetting in the epistemic background means, thereby lifting well-known and novel forgetting operations to the epistemic level. We present five general types of epistemic forgetting and instantiate them with seven concrete forgetting operations for Spohn's ranking functions. We take inspiration from postulates of forgetting both from logic programming and AGM theory to propose a rich landscape of axioms for evaluating forgetting operations. Finally, we evaluate all concrete forgetting operations according to all postulates, leading to a novel comprehensive overview highlighting differences and commonalities among the forgetting operators.
\end{abstract}


\begin{highlights}
    \item Exploration and formal characterization of types of forgetting from epistemic states;
        
    \item
    Presentation of five general types of epistemic forgetting:
    Contraction, Ignoration, Revocation, Marginalization, and Conditionalization;
    
    \item Instantiation of the types of epistemic forgetting by seven forgetting operations for ranking functions: Marginalization, Lifted Marginalization, Conditionalization, c-Ignoration, c-Revision, Minimal c-Revision, and Non-Minimal c-Revisions;
    
    \item Adaption of ASP postulates and AGM postulates for evaluation of epistemic forgetting;
    
    \item Development of novel postulates for epistemic forgetting in a unifying framework;
    
    \item
    Evaluation of all concrete forgetting operations according to all postulates, leading to a novel comprehensive overview highlighting differences and commonalities among the forgetting operators.
\end{highlights}


\end{frontmatter}

\clearpage
\thispagestyle{empty}
\newlength{\parskipTMP}
\setlength{\parskipTMP}{\parskip}
\setlength{\parskip}{0.1mm}
\tableofcontents
\setlength{\parskip}{\parskipTMP}
\clearpage

\pagenumbering{arabic}  
\section{Introduction}
\label{sec:introduction}
\FARBE{Forgetting} \FARBE{in the sense of intentionally ignoring information} \FARBE{plays an essential role in the maintenance, restructuring, and efficient querying of knowledge management systems.}
In many works on forgetting, the focus is on how \FARBE{forgetting affects the knowledge \FARBE{of an agent}}.
In this \FARBE{article,}
we focus on the interaction between forgetting and the entire epistemic state, which encompasses the agent's knowledge but extends beyond it, by considering also additional extra-logical information. 
Prototypically, we consider forgetting in the context of \FARBE{Spohn's} ranking functions~\cite{Spohn88} \FARBE{over propositional logic.
Forgetting as a knowledge management} operation has received much less attention in the field of knowledge representation than operations like inference or revision.  For a broad overview on \FARBE{approaches to} forgetting, we refer the reader to \FARBE{Eiter and Kern-Isberner \cite{KernIsbernerEiter2018}} where also numerous references can be found.
\FARBE{Furthermore,}
at least from a cognitive view, forgetting plays an important role in restructuring and reorganizing a human's mind, and it is closely related to notions like relevance and independence which are crucial to knowledge representation and reasoning.

Two major technical types of forgetting have been presented and studied intensely in the literature: Forgetting of signature elements in the framework of logic programming, \FARBE{ particularly in answer set programming (ASP)}, also \FARBE{more broadly} known as variable elimination~\cite{KS_GoncalvesKnorrLeite2016},
and forgetting of propositional beliefs in the AGM framework of belief change~\cite{AlchourronGaerdenforsMakinson85} where contraction corresponds to forgetting. Both frameworks came up with a number of postulates that should describe rational properties of forgetting. At first sight, these two types of forgetting appear to be very different, both from their technical settings (forgetting of syntactic entities in a program vs.\ forgetting of propositional beliefs in a deductively closed set). Correspondingly, the proposed postulates from both frameworks do not match easily. In particular, while the overall goal of contraction is made very explicit by the postulate of success in AGM theory, the aim of forgetting in logic programs is more implicitly described by several postulates.
\FARBE{Kern-Isberner et al. \cite{KernIsbernerBockBeierleSauerwald2019a,KernIsbernerBockSauerwaldBeierle2019}}
started to set up a joint framework based on epistemic states in which
\FARBE{one is}
able to study and evaluate both main types of forgetting according to postulates adapted from both fields to the setting of epistemic states. As proofs of concept,
\FARBE{also forgetting operations} for each type in the epistemic framework of Spohn's ranking functions \cite{Spohn88} \FARBE{are presented and then evaluated}
according to the adapted postulates
\FARBE{\cite{KernIsbernerBockBeierleSauerwald2019a,KernIsbernerBockSauerwaldBeierle2019}.}
Indeed, those evaluations showed that forgetting by variable elimination vs.\ forgetting by AGM contraction differ in crucial characteristics. However, it became also clear that there are more forgetting operators like conditionalization which show characteristics of both types.

A most crucial insight from these works was to understand and explore the duality of syntactic variable elimination and semantic marginalization --- forgetting an atom in the syntax corresponds generally to semantically marginalize on the rest of atoms. This has been elaborated in detail
\FARBE{by Sauerwald et al.~\cite{KS_SauerwaldBeierleKernIsberner2024} where it is shown}
that a full correspondence can be reached in the context of deductively closed sets of formulas. This is essential for our epistemic approach to forgetting because belief sets of epistemic states are deductively closed in the framework presented here. In particular, it allows us to re-interpret the syntactic forgetting operators and postulates from the area of \FARBE{anwer set programming} in the  more semantic-based setting of epistemic states.

In this \FARBE{article,} we continue to explore different types of forgetting from a thoroughly epistemic perspective to understand epistemic forgetting operations better
while capturing the many facets of forgetting in a unifying framework,
and make basic techniques from \FARBE{forgetting in ASP} and AGM contraction more broadly applicable. We aim to preserve and elaborate the different characteristics of forgetting in both fields, on the one hand, while building bridges between both perspectives on forgetting, on the other hand. To these aims, first, we present a variety of epistemic forgetting operators with quite different \FARBE{characteristics, taking
inspiration from a general framework of forgetting in common-sense knowledge and belief management by Beierle et. al.~\cite{BeierleBockKernIsbernerRagniSauerwald2018KI,BeierleKernIsbernerSauerwaldBockRagni2019KIzeitschrift}.
Second,}
we modify postulates to make their basic intention more adequately expressible in a general epistemic \FARBE{framework.
Third,} we show correspondences between \FARBE{ASP}-motivated postulates and AGM-based postulates.
\FARBE{The epistemic perspective on which this article relies is mainly formalized by considering conditionals, i.e., plausible if-then-statements. Conditionals allow for expressing plausible beliefs beyond classical formulas and hence provide a much richer background for forgetting processes. Spohn's ranking functions \cite{spohn12} are a convenient means to represent conditionals and major features associated with their semantics. }

\FARBE{More precisely,} we present and investigate five general types of epistemic forgetting (Contraction, Ignoration, Revocation, Marginalization, and Conditionalization) which are instantiated by seven forgetting operations for ranking functions (Marginalization, Lifted Marginalization, Conditionalization, c-Ignoration, c-Revision, Minimal c-Revision, and Non-Minimal c-Revisions). We broaden
\FARBE{the} landscape of forgetting from the papers
\FARBE{cited above
\cite{KernIsbernerBockBeierleSauerwald2019a,KernIsbernerBockSauerwaldBeierle2019}
significantly by making use of selection strategies and strategic c-contractions \cite{BeierleKernIsberner2021FLAIRS} throughout this article and by} presenting a more fine-grained view on marginalization and contraction. 
Regarding postulates, in this
\FARBE{article} we also consider the postulate of  Conjunctive Factoring \propCF\ from the AGM framework,
and study Persistence axioms \FARBE{originating from forgetting in ASP \cite{KS_GoncalvesKnorrLeite2016}}. The epistemic perspective allows us to come up with \FARBE{additional} persistence postulates (Epistemic Persistence, \propEP, and Belief Persistence, \propBP). All concrete forgetting operations are evaluated according to all postulates, providing now a novel comprehensive overview that highlights differences, but also commonalities among the forgetting operators. Major findings of previously published work can be confirmed: Marginalization satisfies all postulates from the field of \FARBE{ASP} but does not fully comply with AGM postulates, while Minimal c-Contractions satisfy all AGM postulates while violating the \FARBE{ASP}-postulates. Therefore, these two operations can be regarded as  most typical implementations of variable elimination and AGM-like forgetting in an epistemic setting. The postulates also help us to make the crucial difference between forgetting in \FARBE{ASP} and AGM contractions more intelligible: while in \FARBE{the ASP environment}, the aim of forgetting is to preserve beliefs when forgetting (irrelevant) atoms, as best expressed by Consequence Persistence \propCP\ and other Persistence postulates \FARBE{\cite{KS_GoncalvesKnorrLeite2016}}, the intention of AGM-like forgetting is to give up (relevant) beliefs (cf.\ \agm{1} --- \agm{3}).

Considering Equivalence postulates also \FARBE{reveals} novel insights presented in this \FARBE{article.} In its epistemic version presented here, the \propE\ postulate claims that equivalent epistemic states should remain equivalent under forgetting. This is indeed a strong postulate that is also hard to fulfill in \FARBE{ASP} (cf., e.g., \FARBE{Gon{\c{c}}alves et al.}~\cite{KS_GoncalvesKnorrLeite2016a}). In that setting, there are two \FARBE{more} postulates regarding equivalence \FARBE{available in the literature}: Weak Equivalence  considers equivalence on the level of answer sets, whereas Strong Equivalence  relies on so-called HT-structures \FARBE{\cite{KS_GoncalvesKnorrLeite2016a}}.
\FARBE{Strong Equivalence}
comes close to epistemic \FARBE{Equivalence, while
Weak Equivalence}
focuses on belief sets only
and finds an adequate formalization as the \propWE\ postulate here. Again, for \propE\ and \propWE, we observe a clear distinction in the characteristics of variable elimination resp.\ marginalization and AGM-like forgetting -- marginalization satisfies both \propE\ and \propWE, but AGM contraction violates both. However, \FARBE{we present} a third aspect of equivalence under forgetting that we introduce as Belief Equivalence \propBE\ here: equivalence on the epistemic level can ensure equivalence after forgetting on the level of belief sets.
\FARBE{This postulate} can indeed bridge the dichotomy between the \FARBE{ASP} vs.\ AGM perspective on forgetting: it is one of the very few postulates that is fulfilled by all forgetting operations considered in this \FARBE{article.} Moreover, we also succeed in complying with the challenging Equivalence postulate in our ranking-based framework by making use of the arithmetic structure of ranking functions. We employ the notion of linear equivalence \FARBE{\cite{AH_KernIsberner-etal_2024-KR}} for ranking functions and show that in principle, all our forgetting operations can preserve linear equivalence, i.e., they satisfy the postulate \propxE.
So, beyond the many novel technical results presented in this article, the unifying epistemic perspective on different forms of forgetting provides a much more coherent understanding of forgetting and proves to allow for  overcoming limitations that are inherited by  specific frameworks.

\FARBE{In summary, the main contributions of this article are the following:
\begin{itemize}
\item Exploration and formal characterization of types of forgetting from epistemic states;

\item
Presentation of five general types of epistemic forgetting:
Contraction, Ignoration, Revocation, Marginalization, and Conditionalization;

\item Instantiation \FARBE{of} the types of epistemic forgetting by seven forgetting operations for ranking functions: Marginalization, Lifted Marginalization, Conditionalization, c-Ignoration, c-Revision, Minimal c-Revision, and Non-Minimal c-Revisions;

\item \FARBE{Adaption} of \FARBE{ASP} postulates and AGM postulates for evaluation of epistemic forgetting;

\item Development of novel postulates for epistemic forgetting in a unifying framework;

\item
Evaluation of all concrete forgetting operations according to all postulates, leading to a novel comprehensive overview highlighting differences and commonalities among the forgetting operators. 
\end{itemize}
}

\FARBE{This article revises and largely extends the work of two conference 
papers presented at the
\emph{32nd International FLAIRS Conference}
\cite{KernIsbernerBockBeierleSauerwald2019a} and at the 
\emph{16th International PRICAI Conference}
\cite{KernIsbernerBockSauerwaldBeierle2019}.
As outlined above, the extensions include in particular the identification of 
new subclasses of
forgetting operations, the development of additional novel postulates,
a much broader and more detailed evaluation of all
forgetting operations
with respect to all considered postulates, and a significantly extended overview of epistemic forgetting operators and their characteristics.  
}

\FARBE{%
\FARBE{All types of forgetting} addressed in this article can be classified as \emph{intentional forgetting}
\FARBE{\cite{BeierleTimm2019KIzeitschrift}}
as they generally presuppose the consideration of (conditional) beliefs and a mental process of focusing beyond the formal techniques. This is different from (unconscious and often undesired) forgetting processes we experience in everyday life when we are not able to remember things properly.}

\FARBE{The rest of this article is organized as follows:} \FARBE{In Section \ref{sec:prelim}, we recall formal preliminaries from propositional logic, conditionals, and Spohn's ranking functions, also called Ordinal Conditional Functions (OCF) \cite{Spohn88}. We also formalize basic lemmata here that will be useful throughout this article. Section \ref{sec:bg_marginlisation} investigates in depth the close connection between marginalization and variable forgetting. This will allow us to establish immediate links between syntactic forms of forgetting and epistemic forgetting operators. In Section \ref{sec:kindsOfForgetting}, we make prerequisites of epistemic states explicit on which our approaches rely and propose five (non-exclusive) general types of forgetting operators which are categorized according to abstract success postulates. Section \ref{sec:general_props} presents a collection of postulates to guide the well-behavedness of forgetting operators from various perspectives. These postulates are based on postulates from \FARBE{ASP} (see, in particular, \cite{KS_GoncalvesKnorrLeite2016a}) and AGM belief revision theory \cite{AlchourronGaerdenforsMakinson85} with suitable modifications for the setting of epistemic states. We also present some relationships between postulates from \FARBE{the ASP framework}, on the one hand, and AGM postulates, on the other hand, to establish first bridges between the two frameworks.
Section \ref{sec:instantiating} shows instantiations of all types of forgetting proposed in Section \ref{sec:kindsOfForgetting} in the framework of ranking functions. The rest of the article is mainly dedicated to evaluate all operators by all postulates. First, we elaborate on technical specialties of ranking functions that are relevant for the evaluation in Section \ref{sec_eval_ocf_specifics}. In particular, we present characterizations of major postulates for ranking functions here, to alleviate the evaluation of the operators. 
\FARBE{Moreover, we show how linear equivalence allows us to interpret equivalence statements in an arithmetic way, and to propose novel postulates for our epistemic framework.} 
Section \ref{sec_evaluation_ocf_overview} presents an overview on the evaluation of all operators, before we go into the details regarding forgetting by marginalization and conditionalization in Section  \ref{sec_evaluation_ocf_marg_cond}, and forgetting by strategic c-contractions in Section \ref{sec_evaluation_ocf_c-contractions}. Finally, we discuss related work in Section \ref{sec:related_work}, and summarize our main results and point out future work in Section \ref{sec:conclusion}. \FARBE{Counterexamples used for showing} results in Section \ref{sec_evaluation_ocf_c-contractions} are collected in an Appendix.
}

---------------------------------------------------------------------------------------

\section{Basics on Logics, Conditionals, and Ordinal Conditional Functions (OCF)}
\label{sec:prelim}

\FARBE{In this section, we present the \FARBE{required} background on propositional logic and ranking functions.}

\FARBE{\subsection{Propositional Logic}}
\FARBE{A non-empty \FARBE{finite} set of atoms $\Sigma=\{a,b,c, \ldots\}$
	is called a \emph{signature}.
	In the following, several notions will be parameterized by a signature $ \Sigma $. However, when \( \Sigma \) is clear from the context or irrelevant, we may omit the subscript $ \Sigma $ when we use them later in this article.
	
With $\mathcal{L}_\Sigma$ we denote the 
propositional language
generated from \( \Sigma \).
Formulas of \( \cL_{\Sigma} \) are denoted by large Latin letter \( A,B,C, \ldots \), sometimes with subscript.}
For conciseness of notation, we will omit the logical 
\textit{and}-connector, writing $AB$ instead of $A \wedge B$, 
and overlining formulas will indicate negation, i.e., $\overline{A}$ means $\neg A$. 
Furthermore, we require \( \cL \) to contain \( \top \) and \( \bot \), where \( \top \) is interpreted, as usually, as a tautology, and \( \bot \) as a contradiction.

Let $\Omega_\Sigma$ denote the 
set of all possible worlds (propositional interpretations) over $\Sigma$. 
As usual, $\omega \models_\Sigma A$ means that the propositional formula $A \in \mathcal{L}_\Sigma$ holds in the possible world $\omega \in \Omega_\Sigma$, and $ \Mod_\Sigma(A)=\{\FARBE{\omega \in \Omega_\Sigma} \mid \omega\models_\Sigma A \} $ denotes the set of all such possible worlds.
\FARBE{For two formulas $A,B \in \mathcal{L}_\Sigma$, $A$ \emph{entails} $B$, in symbols $A  \models_\Sigma B$, if $\Mod_\Sigma(A) \subseteq \Mod_\Sigma(B)$. }
\FARBE{	
	By slight abuse of notation, we will use $\omega$ both for the model and the corresponding complete conjunction containing all atoms either in positive or negative form. 
This will simplify notations a lot in the course of this article.}
With $ A \equiv_\Sigma B $ we denote semantic equivalence defined as usually, i.e. $ \Mod_\Sigma(A)=\Mod_\Sigma(B) $.
\FARBE{We say a formula \( A \) is \FARBE{\emph{contingent}} if \( A\not\equiv\top \) and \( A\not\equiv\bot \).}
With $ \Cn_\Sigma(A)=\{ F \in \mathcal{L}_\Sigma \mid A \models_\Sigma F \} $ we denote the set of all logical consequences of $ A $.

\FARBE{%
\FARBE{The notions mentioned above}
extend naturally to sets of formulas \( X \subseteq \mathcal{L}_\Sigma \), i.e., by defining \( \Mod_\Sigma(X)= \bigcap_{A \in X} \Mod_\Sigma(A) \).
\FARBE{For $A \in \mathcal{L}_\Sigma$ and \( X \subseteq \mathcal{L}_\Sigma \)
we then obtain}
\( X \models_\Sigma A \) if \( \Mod_\Sigma(X) \subseteq \Mod_\Sigma(A) \), for sets of formulas \( X_1,X_2 \subseteq \mathcal{L}_\Sigma \) we have \( X_1 \equiv_\Sigma X_2 \) if \FARBE{\( \Mod_\Sigma(X_1) = \Mod_\Sigma(X_2 ) \)}.
For a set of formulas $ L\subseteq \mathcal{L}_\Sigma $, we define $ \Cn_\Sigma(L)=\{ F \in \mathcal{L}_\Sigma \mid L \models_\Sigma F \} $ and say that \( L \) is \emph{deductively closed} if $ \Cn_\Sigma(L)=L $.}

\FARBE{For a deductively closed set of formulas \( \Cn_\Sigma(X) \) we denote its signature with \( \signatureOf{\Cn_\Sigma(X) }=\Sigma \).}
The following lemma shows that the signature of a deductively closed set can be expanded (and to a certain \FARBE{extent}, reduced) without loss of information.
\begin{lemma}
    \label{lem:cn-subsignature}
    Let $\Sigma' \subseteq \Sigma$ and $X,Y \subseteq \cL_{\Sigma'}$. Then the following statements hold:
    \begin{enumerate}
        \item $\Mod_{\Sigma}(X) \subseteq \Mod_{\Sigma}(Y)$ iff $\Mod_{\Sigma'}(X) \subseteq \Mod_{\Sigma'}(Y)$.
        \item $\Mod_{\Sigma}(\Cn_{\Sigma'}(X)) = \Mod_{\Sigma}(X)$.
        \item $\Cn_{\Sigma} (\Cn_{\Sigma'}(X)) = \Cn_{\Sigma}(X)$.
    \end{enumerate}
\end{lemma}
\begin{proof}
    For (1), note that $X,Y$ contain no symbols in $\cL_{\Sigma} \setminus \cL_{\Sigma'}$. Hence the truth value of those symbols is irrelevant for determining the truth value of $X$ or $Y$.    
    For (2), observe that $\Mod_{\Sigma}(\Cn_{\Sigma'}(X)) \subseteq \Mod_{\Sigma}(X)$ is necessary since $X \subseteq \Cn_{\Sigma'}(X)$. Furthermore, since $\Mod_{\Sigma'}(X) \subseteq \Mod_{\Sigma'}(A)$ holds for all $A \in \Cn_{\Sigma'}(X)$, it follows from the first statement that also $\Mod_{\Sigma}(X) \subseteq \bigcap_{A \in \Cn_{\Sigma'}\FARBE{(X)}} \Mod_{\Sigma}(A) \subseteq \Mod_{\Sigma}(\Cn_{\Sigma'}(X))$ must hold.
    The statement (3) is now an easy consequence of the second statement, since $A \in \Cn_{\Sigma}(\Cn_{\Sigma'}(X))$ iff \FARBE{$\Mod_{\Sigma}(\Cn_{\Sigma'}(X)) \subseteq \Mod_{\Sigma}(A)$}, and \FARBE{$\Mod_{\Sigma}(X) \subseteq \Mod_{\Sigma}(A)$} holds iff $A \in \Cn_{\Sigma}(X)$. \FARBE{Hence $A \in \Cn_{\Sigma}(\Cn_{\Sigma'}(X))$ iff $A \in \Cn_{\Sigma}(X)$.}
\end{proof}
    
If $A \in \cL_\Sigma$ is a formula, \FARBE{then \( \signatureOf{A} \) denotes the set of atoms that appear in \( A \), and} the minimal set of signature elements from $\Sigma$ to represent a formula which is equivalent to $A$ is denoted by $\signatureMinOf{A}$.
Parikh showed that \( \signatureMinOf{A} \) is unique and
\FARBE{well-defined \cite[Lem. 2]{Parikh99}.}
Note that $\signatureMinOf{A}=\emptyset$ holds if and only if \( A\equiv\bot \) or \( A\equiv\top \) holds. 
    Moreover, negation is a neutral operation regarding minimal signatures, i.e., \( \signatureMinOf{A} = \signatureMinOf{\neg A} \).
\FARBE{
	Furthermore, we will make use of the following property.
\begin{lemma}
	\label{lem:sigmin_conjunction}
\FARBE{For every two formulas \( A\) and \(C \) the following statements hold:
\begin{enumerate}
    \item \( \signatureMinOf{\neg A} = \signatureMinOf{A} \)
    \item \( \signatureMinOf{A\land C} \subseteq \signatureMinOf{A} \cup \signatureMinOf{C} \)
    \item \( \signatureMinOf{A \lor C} \subseteq \signatureMinOf{A} \cup \signatureMinOf{C} \)
\end{enumerate}}
\end{lemma}
\begin{proof}\FARBE{Statement 1 is \FARBE{immediate}.
    We show Statement 2.}
    For \( A \) and \( C \) there exist propositional formulas \( A_{\min}, C_{\min} \)  such that \( \signatureOf{A_{\min}}=\signatureMinOf{A_{\min}}=\signatureMinOf{A} \) and \( A_{\min} \equiv A \), and  \( \signatureOf{C_{\min}}=\signatureMinOf{C_{\min}}=\signatureMinOf{C} \)
    and \( C_{\min} \equiv C \).
	By the properties of propositional logic, we have that \(A_{\min} \land C_{\min} \equiv A\land C  \).
	Clearly, we have \( \signatureOf{A_{\min} \land C_{\min}} \subseteq \signatureMinOf{A} \cup \signatureMinOf{C} \).
	Because \( \signatureMinOf{B} \subseteq \signatureOf{B}  \) holds for every formula \( B \), the observations \(A_{\min} \land C_{\min} \equiv A\land C  \) and \( \signatureOf{A_{\min} \land C_{\min}} \subseteq \signatureMinOf{A} \cup \signatureMinOf{C} \) together imply the statement \( \signatureMinOf{A \land C} \subseteq \signatureMinOf{A} \cup \signatureMinOf{C} \).
    
\FARBE{We show that Statement 3 holds. 
    We have that \( \signatureMinOf{A \lor C} = \signatureMinOf{\neg (A \lor C)}  \) holds because of Statement 1.
    By employing De Morgan's laws, we have \( \neg (A \lor C) \equiv \neg A \land \neg C \) and consequently also, \( \signatureMinOf{A \lor C} = \signatureMinOf{\neg A \land \neg C}  \).
    Now, by employing Statement 1 and Statement 2, we obtain \( \signatureMinOf{\neg A \land \neg C} \subseteq \signatureMinOf{\neg A} \cup \signatureMinOf{C} = \signatureMinOf{A} \cup \signatureMinOf{C} \).
    The combination \FARBE{of} the latter observation with \( \signatureMinOf{A \lor C} = \signatureMinOf{\neg A \land \neg C}  \) yields Statement 3.}
\end{proof}}

The \emph{theory of $\omega$} is the set of all formulas of which a possible world $\omega$ is a model, i.e.,
\FARBE{$\Th(\omega) = \{ A \in \mathcal{L} \, | \, \omega  \models A \}$,} 
and correspondingly for sets $M$ of possible worlds, we have
\FARBE{$\Th(M) = \{ A \in \mathcal{L} \, | \, \omega  \models A \, \mbox{ for all } \omega \in M \} $.}
Clearly, $\Th(M) = \bigcap_{\omega \in M} \Th(\omega)$.
For convenience, we recall some basic statements on theories which will be very useful in this
 \FARBE{article}. 

\begin{lemma}
\label{Lemma_Th}
Let $\omega$ be a possible world, let $M, M_1$ and $M_2$ be sets of possible worlds. Then the following statements hold: 
\begin{enumerate}
 \item $\Th(\omega) = \Cn(\omega)$, where $\omega$ is seen as a complete conjunction on the right hand side of the equation.
 \item $\Th(M) = \Cn(\bigvee_{\omega \in M} \omega)$,  where again, $\omega$'s are considered both as possible worlds and complete conjunctions on the right hand side of the equation.
 \item $\Th(M_1) \subseteq \Th(M_2)$ iff  $M_2 \subseteq M_1$. 
 \item $\Th(M_1 \cup M_2) = \Th(M_1) \cap \Th(M_2)$.
\end{enumerate}
\end{lemma}

\begin{proof}
(1) is clear from the definitions of $\Th$ and $\Cn$, and (2) follows immediately \FARBE{from (1)}
by observing that $\Th(M) = \bigcap_{\omega \in M} \Th(\omega)$. For two sets $M_1$, $M_2$ of possible worlds, we have $M_2 \subseteq M_1$ iff $\bigvee_{\omega \in M_2} \omega \models \bigvee_{\omega \in M_1} \omega$, which is equivalent to $\Cn(\bigvee_{\omega \in M_1} \omega) \subseteq \Cn(\bigvee_{\omega \in M_2} \omega)$. Thus, by (2), also (3) holds. (4) holds since $\Th(M_1) \cap \Th(M_2) = \bigcap_{\omega \in M_1 \cup M_2} \Th(\omega) = \Th(M_1 \cup M_2)$.
\end{proof}

\FARBE{\subsection{Conditionals and Ordinal Conditional Functions}}
By introducing a new binary operator $|$, we obtain the set $(\mathcal{L}|\mathcal{L}) = \{ (B|A) \mid A,B \in \mathcal{L}\}$ of conditionals over $\mathcal{L}$. $(B|A)$ formalizes ``\textit{if $A$ then usually $B$}'' and establishes a plausible connection between the \emph{antecedent} $A$ and the \emph{consequent} $B$. 
Conditionals with tautological antecedents are taken as plausible statements about the world. 
Following \FARBE{de Finetti} \cite{deFinetti37orig}, a conditional $(B|A)$ can be \emph{verified} (\emph{falsified}) by a possible world $\omega$ iff $\omega \models AB$ ($\omega \models A \overline{B}$). 
If $\omega \not\models A$, then we say the conditional is \emph{not applicable} to $\omega$.
Because conditionals go well beyond classical logic, they require a richer setting for their semantics than classical logic.

\emph{Ordinal conditional functions (OCFs)}, (also called \emph{ranking
    functions})
    \FARBE{$\kappa: \Omega \to \naturals$} with
$\kappa^{-1}(0) \neq \emptyset$, were introduced
(in a more general form)
first by \cite{Spohn88}. They express degrees of plausibility of propositional formulas $A$
by specifying degrees of disbelief of their negations $\overline{A}$.
More formally, we have
$\kappa(A) := \min\{\kappa(\omega) \mid \omega \models A\}$, so that
$\kappa(A \vee B) = \min\{\kappa(A), \kappa(B)\}$. 
Hence, due to $\kappa^{-1}(0) \neq \emptyset$, at least one of $\kappa(A), \kappa(\overline{A})$ must be $0$.

\FARBE{$\beliefsOf{\kappa}$ denotes
the theory of propositional formulas that hold in all
models that are most plausible under $\kappa$,}
i.e.,
\FARBE{
\begin{equation}
 \label{eq_bel_th}
 \beliefsOf{\kappa} = \Th(\{ \omega \, | \, \kappa(\omega) = 0 \}).
\end{equation}
}
The following lemma is immediate with Lemma \ref{Lemma_Th}:

\begin{lemma}
 \label{Lemma_Bel}
 \FARBE{The following \FARBE{statements} hold for each OCF $\kappa$ and for each proposition \( A \in \cL \):
 \begin{itemize}
 	\item $\beliefsOf{\kappa} = \Cn(\bigvee_{\omega: \kappa(\omega) = 0} \omega)$.
 	\item $A \in \beliefsOf{\kappa}$ if and only if $\kappa(\ol{A}) > 0$.
 \end{itemize}}
\end{lemma}

OCFs can serve as representations of epistemic states providing semantics for conditionals:
A conditional $(B|A)$ is accepted in the epistemic state represented by $\kappa$,
written as $
\kappa \models (B|A) \text{, iff }
\kappa(AB) < \kappa(A\notB), $
i.e., iff the verification $AB$ of the conditional is strictly more plausible than its falsification $A\notB$.
Note that for a propositional formula $A$, we have $\kappa \models A$ iff $\kappa \models (A|\top)$.

\section{Marginalization and Variable Forgetting}\label{sec:bg_marginlisation}
Forgetting of signature elements in propositional logic has been considered by the notions of variable elimination \cite{KS_LinReiter1994} and marginalization \cite{KS_SauerwaldBeierleKernIsberner2024}.
In the course of this section, we present both approaches, show how they are interrelated and present the properties of these operators, which we will make use of.

\subsection{Marginalization of Models}
The notion of \FARBE{marginalization} used here is rooted in  probability theory~\cite{Pearl88} and probabilistic logic, where it is the operation on interpretations that computes the marginal probability of a set of atoms \( \sigmaprime \).

In the setting here, where classical propositional logic is used, marginalization is the reduction to the \( \sigmaprime \)-part of an interpretation. For a subset of signature elements $\Sigma^\prime\subseteq\Sigma$ and a world $\omega\in\Omega_\Sigma$ we denote the \emph{$\Sigma^\prime$-part of $\omega$} with $\omega^{\Sigma^\prime} \in \Omega_{\Sigma'}$, mentioning exactly the atoms from $\Sigma^\prime$ such that $\omega\models\omega^{\Sigma^\prime}$. We lift the notion of \( \sigmaprime \)-part to set of interpretations, this operation is called model marginalization \cite{KS_SauerwaldBeierleKernIsberner2024}.
\begin{definition}[model marginalization]\label{def:marginalization_prop}
	For a subset $ \Sigma'\subseteq \Sigma $ and a set $ M\subseteq \Omega_{\Sigma}$ we denote with  
\begin{equation*}
		\margsub{M}{\Sigma'} = \{ \omega^{\Sigma^\prime} \mid \omega \in M \}
\end{equation*}
the \emph{model marginalization of \( M \) from \( \Sigma \) to \( \sigmaprime \)}.
\end{definition}
When viewing a logic as an institution \cite{GoguenBurstall92JACM}, the marginalization of models to a subsignature as given in Definition~\ref{def:marginalization_prop} is just a special case of the general forgetful functor \( \Mod(\varphi) \) from \( \Sigma \)-models to \( \sigmaprime \)-models induced by any signature morphism \( \varphi \) from \( \sigmaprime \) to \( \Sigma \).
The special case of Definition~\ref{def:marginalization_prop} is given by the forgetful functor \( \Mod(\iota) \) induced by the signature inclusion \( \iota : \sigmaprime \to \Sigma \) (cf. also \cite{BeierleKernIsberner2012AMAI}).

In this article, we will make use of the fact that \( \signatureMinOf{A} \) is the set of those atoms that distinguish models of \( A \) from non-models of \( A \) by exactly one signature element.
	\begin{proposition}[\cite{KS_SauerwaldBeierleKernIsberner2024}]\label{prop:sigmin_semantically}
		For each propositional formula \( A \in \mathcal{L}_\Sigma \) we have
		\begin{equation*}
			\signatureMinOf{A} = \{ a \in \Sigma \mid  \ \exists \omega_1,\omega_2\in\Omega_\Sigma.\ \omega_1\models A \text{ and } \omega_2\not\models A \text{ and } \omega_1^{\Sigma\setminus\{a\}}=\omega_2^{\Sigma\setminus\{a\}}   \}.
		\end{equation*}
\end{proposition}
We close this \FARBE{subsection on marginalization} with an example.
\begin{example}
	For illustration, consider $\Sigma=\{a,b,c\}$ with $\omega_1=\overline{a}b\overline{c} \in \Omega_\Sigma$ and $\Sigma^\prime=\{a,b\}$.
	The $\Sigma^\prime$-part of $\omega_1$ is $\omega_1^{\Sigma^\prime}=\overline{a}b$.
	For $\omega_2=\overline{a}bc$ the $\Sigma^\prime$-part
        \FARBE{is the same.} %
        Because we have \( \omega_1 \models b \land \neg c \) \FARBE{and
\( \omega_2 \not\models b \land \neg c \),} the atom \( c \) is an element of \( \signatureMinOf{b \land \neg c} \) due to Proposition~\ref{prop:sigmin_semantically}.
\end{example}

\subsection{Variable Forgetting}
Forgetting in a logical setting is sometimes understood as removing a variable by an syntactic operation. The approach is rooted in the work of Lin and Reiter \cite{KS_LinReiter1994}, which established a whole line of research on that topic, e.g., \cite{KS_LangLiberatoreMarquis2003,Delgrande17}.
However, the technical notion was already considered by Boole~\cite{KS_Boole1854}.
\begin{definition}[variable forgetting \cite{KS_LangLiberatoreMarquis2003}]\label{def:variable_forgetting}
	Let \( A \in \mathcal{L}_\Sigma \)
	be a formula, let $ a \in \Sigma$ be an atom, and let \( \sigmaprime \subseteq \Sigma \) be a subsignature.
	The \emph{variable forgetting of \( a \) in \( A \)}, 
	\begin{equation*}
		\varForget{A}{a} = A[a/\top]\vee A[a/\bot] \ ,
	\end{equation*}
	arises from $A$ by replacing all occurrences of $a$ 
	by $\top$, yielding $A[a/\top]$,  and by replacing all occurrences of \( a \) by $\bot$, yielding $A[a/\bot]$.	
	The \emph{variable forgetting of \( \sigmaprime \) in \( A \)}, 
	denoted by $\varForget{A}{\sigmaprime}$, is	the result of successively eliminating all variables  
	of \( \sigmaprime \) in \( A \) that appear in \( \signatureOf{A} \).
\end{definition}    
It has been shown that $\varForget{A}{\sigmaprime}$ yields a syntactically equivalent formula (up to associativity of disjunction) for every order on \( \sigmaprime \) \cite{KS_LangLiberatoreMarquis2003}. Clearly, when one is interested in the exact syntactic result of \( \varForget{A}{\sigmaprime} \), order of execution matters. However, as we consider here semantic equivalences, we not investigate this further.

\subsection{Interrelation of Marginalization and Variable Forgetting}
Variable forgetting is a syntactic operation on formulas and model marginalization is a semantic operation on interpretations.  
Moreover, variable forgetting  takes the signature elements to be removed as a parameter, while for marginalization the posterior sub-signature is a parameter. 
\FARBE{In order to deal with this duality conveniently,
we define \emph{syntactic marginalization}} as the dual of variable forgetting.
\begin{definition}[syntactic marginalization, \cite{KS_SauerwaldBeierleKernIsberner2024}]
	Let \( A \in \mathcal{L}_\Sigma \)	and let \( \sigmaprime\subseteq\Sigma \).
	The \emph{syntactic marginalization of \( A \) (from \( \Sigma \)) to \( \sigmaprime \)}, written \( \synmarg{\Sigma}{\sigmaprime}{A} \), is \( \varForget{A}{\Sigma\setminus\sigmaprime} \).
\end{definition}
The syntactic marginalization of a formula to a reduced signature
is equivalent to the formula obtained by forgetting all variables that are not in the subsignature.

\begin{figure}[t]\centering
	\begin{tikzpicture}
		\node (start)  at (0,0){$A$};
		\node (ur) [text width=4cm] at (8,0) {\(\synmarg{\Sigma}{\sigmaprime}{A} \)\newline \(= \varForget{A}{\Sigma\setminus\sigmaprime} \)};
		\node (ll)  at (0,-1.8) {$\Mod_\Sigma(A)$};
		\node (lr) [text width=6.1cm] at (8,-1.8) {$ \margsub{\Mod_\Sigma(A)}{\Sigma'} = \Mod_{\sigmaprime}(\synmarg{\Sigma}{\sigmaprime}{A})$};
		
		\draw (start) edge[->] node [above,text width=4.7cm] {\small syntactic marginalization to \( \sigmaprime \)\newline (variable forgetting of \( \Sigma\setminus\sigmaprime \))} (ur);
		\draw (start) edge[->] node [left] {\small \( \Sigma \)-models} (ll);
		\draw (ur) edge[->] node [right] {\small \( \sigmaprime \)-models} (lr);
		\draw (ll) edge[->] node [below] {\small model marginalization to \( \sigmaprime \)} (lr);
	\end{tikzpicture}%
	\caption{Semantic compatibility between marginalization and variable forgetting.}\label{fig:marginalisation_diagram}
\end{figure}
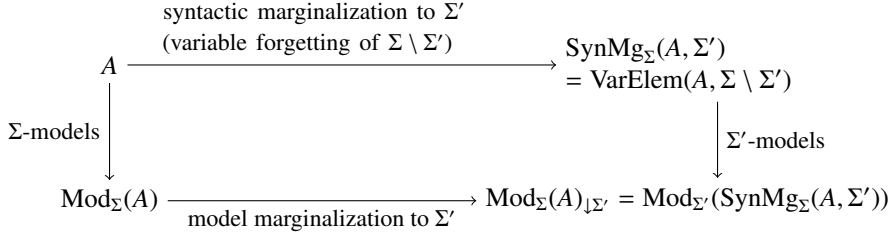
We obtain the following compatibility result for variable forgetting and marginalization which is closely related to known results on variable forgetting~\cite{KS_LangLiberatoreMarquis2003}.
Figure \ref{fig:marginalisation_diagram} provides an  illustration of these interrelations.
\begin{proposition}\label{prop:variableelemination_model_syntactic}
	For every $ A \in \mathcal{L}_\Sigma $ and \FARBE{every} $ \sigmaprime \subseteq \Sigma $ the following holds:
	\begin{align*}
		\margsub{\Mod_\Sigma(A)}{\Sigma'} & =  \Mod_{\sigmaprime}(\synmarg{\Sigma}{\sigmaprime}{A}) \\
		& = \Mod_{\sigmaprime}(\varForget{A}{\Sigma\setminus\sigmaprime})
	\end{align*}
\end{proposition}

Marginalization is sensitive to the syntactic structure.
In particular, marginalization does not comply with negation and conjunction \cite{KS_SauerwaldBeierleKernIsberner2024}.
For instance, when one has \( A \equiv \neg B \), then, in general, \( \synmarg{\Sigma}{\sigmaprime}{A} \equiv \neg\synmarg{\Sigma}{\sigmaprime}{B} \) does not hold.
However, marginalization and thus also variable forgetting are compatible with disjunction, which is shown in the following proposition.
\begin{proposition}[\cite{KS_SauerwaldBeierleKernIsberner2024}]%
    \label{prop:variableelemention_disjunction}
    For each $ A \equiv_\Sigma A_1 \lor \ldots \lor A_n $ with $ A, A_1,\ldots A_n \in \mathcal{L}_\Sigma $ and each $ \sigmaprime\subseteq \Sigma $ the following holds:
    \begin{equation*}
        \synmarg{\Sigma}{\sigmaprime}{A} \equiv_{\sigmaprime} 	\synmarg{\Sigma}{\sigmaprime}{A_1} \lor \ldots \lor \synmarg{\Sigma}{\sigmaprime}{A_n}  
    \end{equation*}
\end{proposition}

\subsection{Marginalization for Deductively Closed Sets}
\begin{figure}[t]\centering
	\begin{tikzpicture}
		\node (start)  at (0,0){$X$ with \( \Cn_\Sigma(X)=X \)};
		\node [anchor=east] (lr) at (8.5,-1.75) {$\synmarg{\Sigma}{\sigmaprime}{A}$};
		\node [anchor=east] (lm) at (7.75,-3.5) {$\synmarg{\Sigma}{\sigmaprime}{X} = Cn_{\sigmaprime}(\synmarg{\Sigma}{\sigmaprime}{A}) =   X \cap \mathcal{L}_{\Sigma'}$};
		\node [anchor=center] (ur) at (start.center -| lr.north) {\( A \)};
		
		\draw (start) edge[dotted,-] node [above,text width=2.6cm] {\small \( \Sigma \)-equivalence \( \equiv_{\Sigma} \)} (ur);
		\coordinate (intermed) at (start.south |- lm.north);
		
		\coordinate (lrs) at ([xshift=0.25cm]lr.south);
		\coordinate (lrsm) at ([xshift=-0.25cm]lr.south);
		
		\draw ([xshift=-0.25cm]start.south) edge[->] node [text width=1.5cm,align=right,left]{\small intersection with \( \mathcal{L}_{\Sigma'} \)}  ([xshift=-0.25cm]intermed);
		\draw ([xshift=0.25cm]start.south) edge[->] node [text width=3cm,align=left,right]{\small syntactic\\marginalization to \( \sigmaprime \)} ([xshift=0.25cm]intermed);
		\draw (ur) edge[->] node [text width=3cm,align=left,right] {\small syntactic\\marginalization to \( \sigmaprime \)} (ur.south |- lr.north);
		\draw (lrs) edge[->] node [right] {\small \( \sigmaprime \)-closure \( \Cn_{\sigmaprime} \)} (lrs |- lm.north);
		\draw (lrsm) edge[dotted,-] node [left] {\small \( \Sigma \)-equivalence \( \equiv_{\Sigma} \)} (lrsm |- lm.north);
	\end{tikzpicture}%
	\caption{%
        Relations provided by Proposition~\ref{prop:bsmargrepresentation_extended} for a deductively closed set \( X \), an equivalent formula \( A \), and their  syntactic marginalizations.}
	\label{fig:marg_belief_sets}
\end{figure}
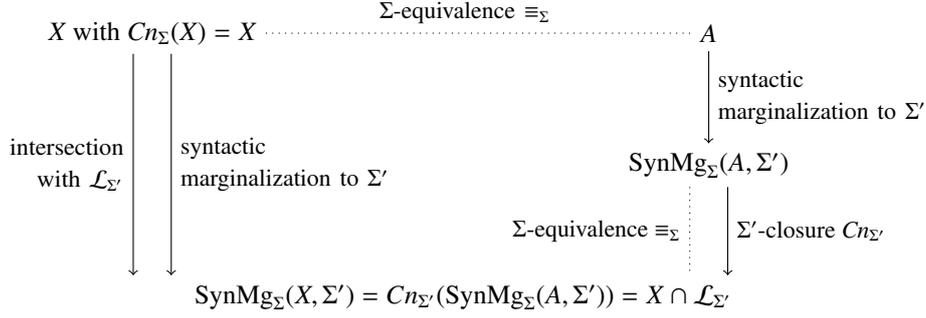

\FARBE{For the rest of the article, we will mainly deal with deductively closed sets and formulas equivalent to them.
	When one uses marginalization on both representations, one has to use a stable notion of (syntactic) marginalization that applies to deductively closed sets and commutes with the marginalization of the equivalent formula.
It has been discovered that plain element-wise marginalization behaves, in general, not very nicely, as plain element-wise marginalization does not always yield a deductively closed set \cite{KS_SauerwaldBeierleKernIsberner2024}.
Because of that, syntactic marginalization of deductively closed sets is defined as the deductive closure of the element-wise marginalization.}
\begin{definition}[\cite{KS_SauerwaldBeierleKernIsberner2024}]\label{def:synmarg_dedsets}
	Let $ X \subseteq \mathcal{L}_\Sigma $ be a deductively closed set \FARBE{and let} \( \sigmaprime\subseteq \Sigma \). 
	The \emph{syntactic marginalization of \( X \) from \( \Sigma \) to \( \sigmaprime \)}, written \( \bsmarg{\Sigma}{\sigmaprime}{X} \), is
	\begin{equation*}
		\bsmarg{\Sigma}{\sigmaprime}{X} =  Cn_{\sigmaprime}(\{ \synmarg{\Sigma}{\sigmaprime}{A} \mid A \in X \})\ ,
	\end{equation*}
	the deductive closure of the element-wise syntactic marginalization of \( X \) to \( \sigmaprime \).
\end{definition}
\FARBE{We want to remark again that syntactic marginalization from Definition~\ref{def:synmarg_dedsets} is different from plain element-wise marginalization on a set of formulas. The difference between both operations has been studied more deeply by Sauerwald et al. \cite{KS_SauerwaldBeierleKernIsberner2024}, where, e.g., it is shown that plain element-wise marginalization of a set of formulas does not commute with the marginalization of an equivalent formula.    
	It also has been shown that syntactic marginalization of deductively closed sets from Definition~\ref{def:synmarg_dedsets} \FARBE{satisfies} the following representation theorem, which we employ heavily in this article.} 
\begin{proposition}[Extended Representation Theorem for Marginalization, \cite{KS_SauerwaldBeierleKernIsberner2024}]\label{prop:bsmargrepresentation_extended}
	Let $ X \subseteq \mathcal{L}_\Sigma $ be a deductively closed set and $ A\in \mathcal{L}_\Sigma $ be a formula representing \( X \), i.e. \( X \equiv_\Sigma A \), then the following holds:
	\begin{equation*}
		\synmarg{\Sigma}{\Sigma'}{X} = Cn_{\Sigma'}(\synmarg{\Sigma}{\Sigma'}{A})  =  X \cap \mathcal{L}_{\Sigma'}
	\end{equation*}
\end{proposition}
By Proposition \ref{prop:bsmargrepresentation_extended} we obtain different characterizations for syntactic marginalization for deductively closed sets and formulas equivalent to them.
Figure~\ref{fig:marg_belief_sets} \FARBE{gives} an illustration of Proposition \ref{prop:bsmargrepresentation_extended}. 
Another approach to forgetting is due to Delgrande \cite{Delgrande17} in which he proposes to understand forgetting of variables as a reduction to a sublanguage, i.e. \( \mathit{forget}(X,\Sigma') =  X \cap \mathcal{L}_{\Sigma\setminus\Sigma'} \).
Proposition~\ref{prop:bsmargrepresentation_extended} shows that the notion of syntactic marginalization for deductively closed sets used here complies with the approach by Delgrande.
\FARBE{In summary, Proposition~\ref{prop:bsmargrepresentation_extended} is a witness for the stability of the notion of syntactic marginalization for deductively closed sets given in Definition~\ref{def:synmarg_dedsets}.}

\subsection{\FARBE{Relevance for Epistemic States}}
\FARBE{In the setting} of epistemic states, which we will introduce in the subsequent section, each epistemic state \( \Psi \) is equipped with a belief set \( \beliefsOf{\Psi} \), which is a deductively closed set of formulas.
For a belief set \( \beliefsOf{\Psi} \), the signature \( \signatureOf{\beliefsOf{\Psi}} \) will be implicitly given by the signature
\FARBE{\( \signatureOf{\Psi} \) of the epistemic state \( \Psi \).}

This gives the rationale to ease the notion for syntactic marginalization of a deductively closed set \( X \) to a subsignature \( \Sigma' \subseteq \signatureOf{X} \) to the following compact notation
\begin{equation}\label{eq_margsub_for_deductively_closed_set}
	\margsub{X}{\Sigma'} = \synmarg{\signatureOf{X}}{\Sigma'}{X}
\end{equation}
which we will use frequently in the rest of this article.

Proposition~\ref{prop:variableelemination_model_syntactic} shows nicely that (model and syntactic) marginalization and variable forgetting of a formula comply with each other semantically. Because of this,
\FARBE{in the following sections, we will}  %
present results primarily from the viewpoint of syntactic marginalization. 
Whenever \( \Sigma \) is given by the context,
for a formula \( A \in \mathcal{L}_\Sigma \), we write
\begin{align}
	\margsub{A}{\Sigma'} & = \synmarg{\Sigma}{\Sigma'}{A}
\end{align}
for the syntactic marginalization of $A$ to the subsignature 
\( \Sigma' \subseteq \Sigma \).
Due to Proposition~\ref{prop:variableelemination_model_syntactic}, we can always switch freely to the semantic operation of model-marginalization.

In the next section, we broaden the view on forgetting by considering the richer setting of epistemic states, which go beyond sets of formulas, and by taking further different forgetting operations into account.
\section{Forgetting in Epistemic States}
\label{sec:kindsOfForgetting}
In this section, we describe the general framework in which we consider forgetting in epistemic states, specify the forgetting operators to be considered in this article \FARBE{on an abstract level}, and recall technical details for realizing forgetting operators in epistemic states represented by OCFs.

\label{subsec_kindsofforgetting}

The paper 
\cite{BeierleKernIsbernerSauerwaldBockRagni2019KIzeitschrift} discusses various kinds of forgetting and specifies characteristic properties for each of the forgetting operations on epistemic states. We recall and expand this discussion here. As in \cite{BeierleKernIsbernerSauerwaldBockRagni2019KIzeitschrift}, an abstract model of  epistemic states (also called belief states) is used in which each epistemic state $ \Psi $ is presupposed to make use of a logical language $ \mathcal{L} $ over a signature $ \Sigma $ and is equipped with
an acceptance (or inference)
relation $ \ESmodels $ with respect to formulas.
		With $ \signatureOf{\Psi} $, we
denote the signature of \FARBE{the logical language $\cL$ on which the epistemic state $ \Psi $ is based}. In particular, if not explicitly stated otherwise, we assume $ \signatureOf{\Psi}=\Sigma $.
The relation $ \Psi \ESmodels \varphi$ holds if an agent with belief state $\Psi$
believes or accepts $\varphi$, where  
$\varphi$ can be a statement from $\cL$, or a conditional from $(\cL|\cL)$. The connection between
the acceptance of both types of statements
is given by $\Psi\ESmodels A$ iff $\Psi\ESmodels (A|\top)$. 
\FARBE{Note that $\ESmodels$ is meant to be a semantic relation, and we presuppose that each state $\Psi$ is equipped with a suitable semantic structure to implement these properties.}
\FARBE{This semantic structure is defined suitably over interpretations of $\cL$ resp.\ $ \signatureOf{\Psi} $, called \emph{possible worlds}.}
	With $\beliefsOf{\Psi}$ we denote the set of all propositional beliefs accepted by $ \Psi $, i.e. $A\in \beliefsOf{\Psi}$ if and only if $ \Psi\ESmodels A $.
Furthermore, we presuppose that $\beliefsOf{\Psi}$ \FARBE{is consistent and that it} is a deductively closed set with respect to \( \signatureOf{\Psi} \), i.e., \( \beliefsOf{\Psi}=\Cn_{\signatureOf{\Psi}}(\beliefsOf{\Psi}) \).
Indeed, validation of conditionals (or rules) is a crucial characteristic feature of epistemic states that distinguishes them from flat belief sets, i.e., logical theories. One may even think of an epistemic state as being (basically) specified by the conditionals that are believed on its base; this view can be found in works on non-monotonic reasoning and belief revision \cite{GoldszmidtPearl96}, and logic programming \cite{EiterSabbatini_etal2001}. Moreover, conditionals can be easily related to total preorders which play an important role in belief revision \cite{KatsunoMendelzon91a}. 

The set of all (conditional) inferences \FARBE{from $\Psi$} will be denoted by $\Cons(\Psi)$ with $\Cons(\Psi)=\{(B|A)  \in \condL \mid \Psi\ESmodels (B|A)\}$.
Using the conditional inference relation,
we define the \emph{entailment relation $\nmmodels$} between belief states $\Psi_1, \Psi_2$ by  $\Psi_1 \nmmodels \Psi_2$ iff $\Cons(\Psi_2) \subseteq \Cons(\Psi_1)$.
Equivalence among epistemic states is defined by the entailment relation $\nmmodels$ on epistemic states with $\Psi_1 \cong \Psi_2$ iff $\Psi_1 \nmmodels \Psi_2$ and $\Psi_2 \nmmodels \Psi_1$.
\FARBE{This is equivalent to stating that $\Cons(\Psi_1) = \Cons(\Psi_2)$.}
Note that for $\Psi_1 \nmmodels \Psi_2$ to hold, we must have $\signatureOf{\Psi_2} \subseteq \signatureOf{\Psi_1}$. Consequently, equivalent epistemic states must be defined over the same signature.

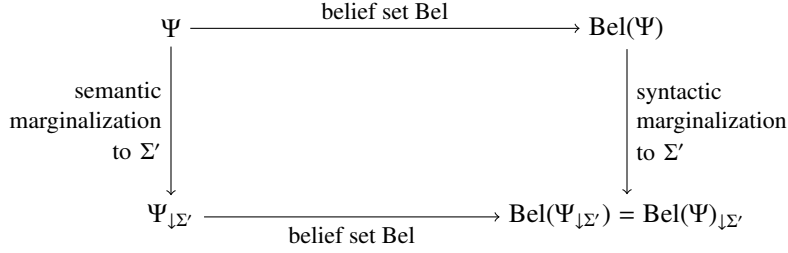
\begin{figure}\centering
    \begin{tikzpicture}
        \node (start)  at (0,0){$\Psi$};
        \node [text width=1cm] (ur) at (6,0) {\( \beliefsOf{\Psi} \)};
        \node (ll)  at (0,-2.5) {$\margsub{\Psi}{\Sigma^\prime}$};
        \node (lr) at (6,-2.5) {$\beliefsOf{\margsub{\Psi}{\Sigma^\prime}} = \margsub{\beliefsOf{\Psi}}{\Sigma^\prime}$};
        
        \draw (start) edge[->] node [above] {\small belief set \( \beliefssymbol \)} (ur);
        \draw (start) edge[->] node [left,text width=2cm,align=right] {\small \FARBE{semantic}\\ marginalization\\ to \( \Sigma' \)} (ll);
        \draw (ur) edge[->] node [right,text width=2cm] {\small syntactic\\ marginalization\\ to \( \Sigma' \)} (lr);
        \draw (ll) edge[->] node [below] {\small belief set \( \beliefssymbol \)} (lr);
    \end{tikzpicture}\caption{Semantic compatibility between marginalization of an epistemic state \( \Psi \) and marginalization of its respective belief set \( \beliefsOf{\Psi} \), \FARBE{cf.\ Proposition~\ref{prop:es_synmag}}.}\label{fig:es_marginalisation_bs_diagram}
\end{figure}

We further assume in this article that epistemic states can be marginalized by restricting \FARBE{their} signature, and that they can be conditionalized by considering only models of a given proposition.
The marginalization of an epistemic state \( \Psi \) to a subsignature \( \Sigma' \subseteq \signatureOf{\Psi} \) will be denoted by \( \margsub{\Psi}{\Sigma^\prime} \).
Hence, \( \margsub{\Psi}{\Sigma^\prime} \) is the unique marginalized epistemic state with \( \signatureOf{\margsub{\Psi}{\Sigma^\prime}}=\Sigma' \) and  with $ \margsub{\Psi}{\Sigma^\prime} \ESmodels (B|A) $ \FARBE{iff} $\Psi \ESmodels (B|A)$ and $(B|A) \in (\mathcal{L}_{\Sigma^\prime}|\mathcal{L}_{\Sigma^\prime})$.
\FARBE{ We denote with \( \margsub{\beliefsOf{\Psi}}{\Sigma'} \) the marginalization of the set of all beliefs accepted by \( \Psi \) (cf.\ Equation~\eqref{eq_margsub_for_deductively_closed_set} in Section \ref{sec:bg_marginlisation}).
Since \( \beliefsOf{\Psi} \) is a deductively closed set of formulas, marginalization} of epistemic states \( \Psi \) and marginalization of their respective belief sets \( \beliefsOf{\Psi} \) are compatible as illustrated in Figure \ref{fig:es_marginalisation_bs_diagram} and as shown in
the following proposition.

	\begin{proposition}\label{prop:es_synmag}
	Let \( \Psi \) be an epistemic state with \( \signatureOf{\Psi}=\Sigma \) and \( \Sigma' \subseteq \Sigma \).
	 Then we have \( 
 \beliefsOf{\margsub{\Psi}{\Sigma'}} = \margsub{\beliefsOf{\Psi}}{\Sigma'} \).
\end{proposition}
\begin{proof}
        Recall that \( \margsub{\beliefsOf{\Psi}}{\Sigma'}  = \synmarg{\Sigma}{\Sigma'}{\beliefsOf{\Psi}} \).
        Using Proposition \ref{prop:bsmargrepresentation_extended},  we obtain \(  \synmarg{\Sigma}{\Sigma'}{\beliefsOf{\Psi}} =  \beliefsOf{\Psi} \cap \mathcal{L}_{\Sigma'} \) for the belief set \( \beliefsOf{\Psi} \). 
	Remember that \( \Psi \models (A|\top) \) if and only if \( A \in \beliefsOf{\Psi} \).
	Thus, employing the definition of marginalization of \( \Psi \) yields \( \beliefsOf{\margsub{\Psi}{\Sigma'}} = \{ \FARBE{A \in \mathcal{L}_{\Sigma'}} \mid A \in \beliefsOf{\Psi} \} \). 
    By standard set theory, we have \( \beliefsOf{\margsub{\Psi}{\Sigma'}} = \beliefsOf{\Psi} \cap \mathcal{L}_{\Sigma'} \FARBE{= \margsub{\beliefsOf{\Psi}}{\Sigma'}} \).
\end{proof}
Another operation available in this framework is conditionalization of epistemic states. For an epistemic state \( \Psi \) and a sentence \( A \), the conditionalization operator $|$ yields $\Psi|A$, the conditionalization of \( \Psi \) by \( A \), which is an epistemic state such
\FARBE{the following  conditions hold:
\begin{align}
&\Psi|A \ESmodels A
\\
&\Psi|A \ESmodels (C|B) \textrm{\ \ iff \  }  \Psi \ESmodels (C|B) \textrm{\ \quad\quad for \ }  B\models A \textrm{ \ and \ } C \models A	
\end{align}
Intuitively,} $\Psi|A$ has the intended meaning that $\Psi$ should be interpreted under the assumption that $A \in \mathcal{L}$ holds.

If $\Psi$ is a prior state, then we denote with $\forgetBy{\Psi}{A}$ the posterior belief state of the agent after forgetting $A$, or after applying the change operation $ \circ $ to $ \Psi $ with input $ A $
\FARBE{which is a formula from $\cL$.}
We focus on the base case of forgetting a single logical \FARBE{item $A$,} but the approaches and results presented in this article can
be extended to deal with forgetting of \FARBE{more general items, e.g., conditionals}.
\FARBE{Furthermore, in order to concentrate on the core concepts and to avoid
\FARBE{lengthy}
case distinctions, we generally assume that $A$ is \FARBE{\emph{contingent},} i.e., $A \not\equiv \top$ and  $A \not\equiv \bot$; all given results can be extended to these borderline cases by taking appropriate case distinctions into account.}

The key idea of this article is that different notions of forgetting can be specified by the inferences an agent can or can no longer draw after forgetting
\cite{BeierleKernIsbernerSauerwaldBockRagni2019KIzeitschrift}.  %
In the following, we focus on three major types of forgetting---contraction, marginalization and \FARBE{\revocation. Additionally, we consider \ignore as  a special kind of contraction that covers a different aspect of forgetting that is of interest for various applications as well. Moreover, we address conditionalization explicitly} as a limit case of \revocation.
\FARBE{For contingent formulas $A$, these}
five forgetting operations are described on \FARBE{an abstract} level as follows:
\begin{align}
                                   \textbf{Contraction}: & \ \quad \forgetBy{\Psi}{A} \notESmodels A    \label{eqGeneralContration}                \\[1.1ex]
                                       \textbf{\Ignore}: & \ \quad \forgetBy{\Psi}{A} \notESmodels A \textrm{ and } \Psi^\circ \notESmodels \notA   \label{eqGeneralIgnoration}  \\[1.1ex] 
                                   \textbf{\Revocation}: & \ \quad \forgetBy{\Psi}{A} \models \ol{A}    \label{eqGeneralRevocation}                \\[1.1ex]
                               \textbf{Marginalization}: & \ \quad \forgetBy{\Psi}{A} = \margsub{\Psi}{\Sigma\setminus \signatureMinOf{A}}   \label{eqGeneralMarginalization}  \\[1.1ex]
                            \textbf{Conditionalization}: & \ \quad \forgetBy{\Psi}{A} = \Psi|\notA \textrm{ with a conditionalization operator } | \textrm{ on } \Psi
 \label{eqGeneralConditionalization}
\end{align}

Contraction refers to the intention to directly give up information $A$, as it is known from the AGM framework~\cite{AlchourronGaerdenforsMakinson85}.
\Ignore is a special contraction that enforces undecidedness between $A$ and $\notA$, thus giving up the judgement on $ A $. 
\Revocation\ is stronger than contraction in that it not only forgets $A$ but also establishes belief in $\ol{A}$.
Marginalization and conditionalization are well-established operators; they are reinterpreted here for defining forgetting operators that take the information $A$ to be forgotten as their argument.
For further discussion on these operators please see~\cite{BeierleKernIsbernerSauerwaldBockRagni2019KIzeitschrift}.
After presenting and discussing postulates for forgetting operators in the following section,
we will instantiate this general framework with ordinal conditional functions and specific forgetting operators based on them. 

\section{General Postulates for Forgetting Operators}
\label{sec:general_props}
In this section, we develop a wide range of postulates for forgetting operators \FARBE{by adopting well-known postulates from the literature to our more general epistemic framework presented in Section \ref{sec:kindsOfForgetting}}. We start with AGM contraction and propose a translation of the AGM contration postulates as postulates for general forgetting operators
(Section~\ref{sec_agm_contraction}).
In Section~\ref{sec_asp_postulates},
we present forgetting postulates inspired by work in answer set programming, and in
Section~\ref{sec_relation_postulates},
we study and elaborate 
interrelationships among the postulates.

\subsection{AGM Contraction}
\label{sec_agm_contraction}

AGM theory \cite{AlchourronGaerdenforsMakinson85} deals with belief change in the context of belief sets, i.e., deductively closed sets of propositions.
\FARBE{Three types of belief change
\FARBE{operations --- expansion, revision, and contraction --- are considered in the AGM theory. Contraction}
corresponds to a form of forgetting: after contracting (a proposition) $A$, it should no longer be possible to infer $A$. Formally, in our framework of epistemic states, this amounts to postulating that $\forgetBy{\Psi}{A} \notESmodels A$ (see (\ref{eqGeneralContration}) in Section~\ref{sec:kindsOfForgetting}). AGM contraction is specified by a set of postulates that we re-interpret here for forgetting operations $\circ$; note that we base this translation on the version of the postulates for epistemic states presented in \cite{KS_CaridroitKoniecznyMarquis2017}.
}

\begin{postulate}
    \begingroup
    \renewcommand{\postulateskip}{10.5ex}
    \setzePostulat{\agm{1}}{Inclusion}%
    {$ \beliefsOf{\forgetBy{\Psi}{A}} \subseteq \beliefsOf{\Psi} $}%
    {Forgetting does not yield new beliefs.}%
    
    \setzePostulat{\agm{2}}{Vacuity}%
    {$ A \notin \beliefsOf{\Psi} \text{, then } \beliefsOf{\Psi} \subseteq \beliefsOf{\forgetBy{\Psi}{A}} $}%
    {Forgetting of \( A \) does not alter the beliefs, if \( A \) is not believed.}%
    
    \setzePostulat{\agm{3}}{Success}%
    {$ \text{If } A \in \beliefsOf{\forgetBy{\Psi}{A}}  \text{, then } A\equiv\top $}%
    {Forgetting \( A \) revokes believing in \( A \), except \( A \) is tautological.}%
    
    \setzePostulat{\agm{4}}{Recovery}%
    {$ \beliefsOf{\Psi} \subseteq \Cn(\beliefsOf{\forgetBy{\Psi}{A}}\cup \{A\}) $}%
    {Forgetting of \( A \) is constituted in a way such that the prior beliefs can be regained by adding \( A \).}%
    
    \setzePostulat{\agm{5}}{Extensionality}%
    {$ \text{If } A\equiv C \text{, then }  \beliefsOf{\forgetBy{\Psi}{A}} = \beliefsOf{\forgetBy{\Psi}{C}} $}%
    {Forgetting of semantically equivalent beliefs results in equivalent beliefs.}%
    
    \setzePostulat{\agm{6}}{Conjunctive Overlap}%
    {$ \beliefsOf{\forgetBy{\Psi}{A}} \cap \beliefsOf{\forgetBy{\Psi}{C}} \subseteq \beliefsOf{\forgetBy{\Psi}{A\land C}} $}%
    {Forgetting of  \( A\land C \) should overlap the  common part of forgetting by \( A \) and forgetting by \( C \).}%
    
    \setzePostulat{\agm{7}}{Conjunctive Inclusion}%
    {$ \text{If }  A \notin \beliefsOf{\forgetBy{\Psi}{A\land C}} \text{, then }    \beliefsOf{\forgetBy{\Psi}{A\land C}} \subseteq \beliefsOf{\forgetBy{\Psi}{A}} $}%
    {If forgetting of \( A\land C \) yields forgetting of \( A \), then forgetting by \( A\land C  \) removes more beliefs than forgetting by \( A \).}%
  \endgroup
  \end{postulate}
  
    \FARBE{We also transfer the following postulate of \emph{conjunctive factoring \propCF} which is known to be implied by %
    the original AGM postulates for contraction in the case of belief sets
    \cite{KS_Hansson1999}:}

     \begin{postulate}
    \begingroup
    \setzePostulat{(\textlabel{CF}{pstl:CF})}{\mbox{Conjunctive Factoring}}%
    {$ \beliefsOf{\forgetBy{\Psi}{A\land C}}{=}\beliefsOf{\forgetBy{\Psi}{A}} \text{ or } \beliefsOf{\forgetBy{\Psi}{A\land C}}{=}\beliefsOf{\forgetBy{\Psi}{C}} \text{ or } \beliefsOf{\forgetBy{\Psi}{A\land C}}{=}\beliefsOf{\forgetBy{\Psi}{A}}\cap \beliefsOf{\forgetBy{\Psi}{C}} $}%
    {Forgetting of a conjunction is reducible to forgetting of each of the conjuncts independently on the level of belief sets.}%
    \endgroup
\end{postulate}

The next \FARBE{subsection deals with postulates for forgetting which are derived} from postulates for forgetting in answer set programming.
 
\subsection{ASP-inspired Postulates}
\label{sec_asp_postulates}

In \cite{KernIsbernerBockBeierleSauerwald2019a},
\FARBE{postulates for forgetting are presented}
which had been inspired by ASP postulates for forgetting \cite{KS_GoncalvesKnorrLeite2016a}. We recall this list of postulates here and extend it by other suitable postulates adapted from \cite{KS_GoncalvesKnorrLeite2016a}. For in-depth discussions on how terms and techniques for epistemic state (may) correspond to the respective counterparts in ASP, please see \cite{KernIsbernerBockBeierleSauerwald2019a}.
\begin{postulate}
    \setzePostulat{\propW}{Weakening}%
    {$\Psi \nmmodels \forgetBy{\Psi}{A}$}%
    {The posterior belief state has at most the same conditional inferences as the prior belief state.}
    
    \setzePostulat{\propWE}{weak Equivalence}%
    {If $ \beliefsOf{\Psi} = \beliefsOf{\Phi} $ then $ \beliefsOf{\forgetBy{\Psi}{A}} \FARBE{=} \beliefsOf{\forgetBy{\Phi}{A}} $}%
 {If the belief sets of two belief states are identical before the forgetting operation, then the posterior belief sets are identical too.}   

    \setzePostulat{\propE}{(Strong) Equivalence}%
    {If $\Psi \cong \Phi$, then $(\forgetBy{\Psi}{A} \cong \forgetBy{\Phi}{A})$}%
    {If two belief states are equivalent before the forgetting operation, then the posterior belief states are equivalent as well.}%
    
    \setzePostulat{\propOI}{Order Independence}%
    {$\forgetBy{(\forgetBy{\Psi}{A})}{B} \cong \forgetBy{(\forgetBy{\Psi}{B})}{A}$ }%
    {The result of an iterated forgetting is independent of the order in which the operators are applied.}%
 \end{postulate}   
    
    \FARBE{In \cite{KernIsbernerBockBeierleSauerwald2019a}, also the postulates \emph{weakened Consequence \propWC, strengthened Consequence \propSC}, and \emph{Consequence Persistence \propCP} were presented. We propose slight variations of these postulates here that ensure that comparisons between belief sets of epistemic states are made with respect to a common signature. For this, we make use of the notation $X \subseteqCNsig{\Sigma} Y$ for relating two sets $X, Y$ of formulas over $\Sigma'$ and $\Sigma''$, respectively, with $\Sigma', \Sigma'' \subseteq \Sigma$.
    \FARBE{Because $\Sigma'$ and $\Sigma''$ may be different, we compare
\(X\) and \(Y\)}
with respect to their consequences in $\Sigma$.
\FARBE{Formally,
\begin{align}\label{eq_subseteqCNsig}
X \subseteqCNsig{\Sigma} Y
\textrm{\ \ \ abbreviates \ \ \ }
\Cn_{\Sigma}(X) \subseteq \Cn_{\Sigma}(Y)
\end{align}
and}
$X \gleichCNsig{\Sigma} Y$
stands for
$\Cn_{\Sigma}(X) = \Cn_{\Sigma}(Y)$.
    }

\begin{postulate}
 \setzePostulat{\hspace{-1em}\propWCSigma}{\FARBE{weakened $\Sigma$-Consequence}}%
    {$ \FARBE{\margsub{\beliefsOf{\Psi}}{\signatureOf{\Psi}\setminus\signatureMinOf{A}} \subseteqCNsig{\signatureOf{\Psi}} \beliefsOf{\forgetBy{\Psi}{A}}} $}%
    {Consequences over $\signatureOf{\Psi}$ of marginalized beliefs of the prior state are preserved; in particular those that do not contain any forgotten atoms}
    
    \setzePostulat{\hspace{-1em}\propSCSigma}{\FARBE{strengthened $\Sigma$-Consequence}}%
    {$ \FARBE{\beliefsOf{\forgetBy{\Psi}{A}} \subseteqCNsig{\signatureOf{\Psi}} \margsub{\beliefsOf{\Psi}}{\signatureOf{\Psi}\setminus\signatureMinOf{A}}} $}%
    {Consequences over $\signatureOf{\Psi}$ of beliefs of the posterior belief state have to be consequences of marginalized beliefs of the  prior belief state.}
    
    \setzePostulat{\hspace{-1em}\propCPSigma}{$\Sigma$-Consequence Persistence}%
    {$ \FARBE{\beliefsOf{\forgetBy{\Psi}{A}} \gleichCNsig{\signatureOf{\Psi}} \margsub{\beliefsOf{\Psi}}{\signatureOf{\Psi}\setminus\signatureMinOf{A}}} $}%
    {Consequences over $\signatureOf{\Psi}$ of beliefs in the posterior belief state are the same as consequences over $\signatureOf{\Psi}$ of marginalized beliefs of the prior belief state; in particular, formulas that do not contain any forgotten atoms are preserved.}
 \end{postulate}

\FARBE{We continue with the \emph{Persistence} postulates that demand for preserving all beliefs that are defined over a sublanguage without any syntactic link to the proposition to be forgotten.}    
    
\begin{postulate}        
    \setzePostulat{\propPP}{Positive Persistence}%
    {If $\Psi \nmmodels \Psi'$ where $\signatureOf{\Psi'} \subseteq \Sigma'$, and $\FARBE{\signatureMinOf{A}} \subseteq \signatureOf{\Psi} \backslash \Sigma'$, then $ \forgetBy{\Psi}{A} \nmmodels \Psi'$}%
    {If $\Psi$ entails an epistemic state $\Psi'$ over a sublanguage, then after forgetting a proposition which is formed over the complementary sublanguage,
    $\Psi'$ shall \FARBE{still be entailed .}}%
    
    \setzePostulat{\propNP}{Negative Persistence}%
    {If $\Psi \not\nmmodels \Psi'$ where $\signatureOf{\Psi'} \subseteq \Sigma'$, and $\FARBE{\signatureMinOf{A}} \subseteq \signatureOf{\Psi} \backslash \Sigma'$, then $ \forgetBy{\Psi}{A} \not\nmmodels \Psi'$}%
    {If $\Psi$ does not entail an epistemic state $\Psi'$ over a sublanguage, then after forgetting a proposition which is formed over the complementary sublanguage,  $\Psi'$ shall not be entailed.
    Note that the contrapositive form of \propNP\ claims that for $ \signatureOf{\Psi'} \subseteq \Sigma'$, and $ \signatureOf{A} \subseteq \signatureOf{\Psi} \backslash \Sigma'$, $ \forgetBy{\Psi}{A} \nmmodels \Psi'$ implies $\Psi \nmmodels \Psi'$, i.e., \propNP\ is the reverse form of \propPP.}
\end{postulate}

\FARBE{Note that \propPP and \propNP can also be  written more concisely in the following form: 

\begin{itemize}        
    \item[(PP')]
    If $\Psi \nmmodels \Psi'$ and $\signatureMinOf{A} \subseteq \signatureOf{\Psi}  \backslash \signatureOf{\Psi'} $, then $ \forgetBy{\Psi}{A} \nmmodels \Psi'$
    \item[(NP')]
    If $\Psi \not\nmmodels \Psi'$ and $\signatureMinOf{A} \subseteq \signatureOf{\Psi}  \backslash \signatureOf{\Psi'} $, then $ \forgetBy{\Psi}{A} \not\nmmodels \Psi'$
\end{itemize}
Nevertheless, we stick to the wordings in \propPP and \propNP which are closer to the wordings in the original postulates, as given in \cite{KS_GoncalvesKnorrLeite2016a}. 
}

\noindent \FARBE{Moreover, since} $\Psi \nmmodels \Psi'$ iff $\Cons(\Psi') \subseteq \Cons(\Psi)$, the Persistence postulates deal with the preservation of conditional beliefs under forgetting. The following postulates make this more explicit:

\begin{postulate}\renewcommand{\postulateskip}{8ex}
    \setzePostulat{\propPPcond}{Conditional Positive Persistence}%
    {If $ \signatureOf{\Psi} = \Sigma_1 \cupdisjoint \Sigma_2$, $B,C \in \cL_{\Sigma_1}$ and $A \in \cL_{\Sigma_2}$, then $\Psi \models (C|B)$ implies $ \forgetBy{\Psi}{A} \models (C|B)$.}%
    {}%
     
    \setzePostulat{\propNPcond}{Conditional Negative Persistence}%
    {If $\signatureOf{\Psi} = \Sigma_1 \cupdisjoint \Sigma_2$, $B,C \in \cL_{\Sigma_1}$ and $A \in \cL_{\Sigma_2}$, then $ \forgetBy{\Psi}{A} \models (C|B)$ implies $\Psi \models (C|B)$.}%
    {}%
\end{postulate}

\noindent Given the symmetry between \propPP\ and \propNP, we propose a novel postulate that fully lifts the idea of persistence to the level of epistemic states: 
\begin{postulate}
    \setzePostulat{\propEP}{Epistemic Persistence}%
    {If $ \signatureOf{\Psi} = \Sigma_1 \cupdisjoint \Sigma_2$ and $A \in \cL_{\Sigma_2}$, then $\margsub{\Psi}{{\Sigma_1}} = \margsub{(\forgetBy{\Psi}A)}{\Sigma_1}$.}%
    {After forgetting a formula of a proper sublanguage, the marginalization of the resulting epistemic state on the complementary subsignature should be the same as before.}%
\end{postulate}
A weaker version of Epistemic Persistence focusses on the beliefs of the involved epistemic states:
\begin{postulate}
    \setzePostulat{\propBP}{Belief Persistence}%
    {If $\signatureOf{\Psi} = \Sigma_1 \cupdisjoint \Sigma_2$ and $A \in \cL_{\Sigma_2}$, then $ \margsub{\beliefsOf{\Psi}}{\Sigma_1}  = \margsub{\beliefsOf{\forgetBy{\Psi}A}}{\Sigma_1}$.}%
    {After forgetting a formula of a proper sublanguage, the marginalization of the resulting beliefs on the complementary subsignature should be the same as before.}%
\end{postulate}
Furthermore, we propose a novel postulate that is in between weak and strong equivalence: 
\begin{postulate}
    \setzePostulat{\propBE}{Belief Equivalence}%
    {If $\Psi \cong \Phi$, then 
        $ \beliefsOf{\forgetBy{\Psi}{A}} = \beliefsOf{\forgetBy{\Phi}{A}} $}%
    {If two epistemic states are equivalent before the forgetting operation, then  their posterior belief sets are
    identical.}%
\end{postulate}
Belief Equivalence postulates that equivalent epistemic states yield (at least) equivalent belief sets after forgetting the same formula. This means that the concept of equivalence between epistemic states is able to capture all information that is relevant for forgetting on the level of belief sets.

Note that the idea behind \propBE can also be used to strengthen \agm{5}:
\begin{postulate}\renewcommand{\postulateskip}{6ex}
    \setzePostulat{\propEBE}{\FARBE{Extensional Belief Equivalence}}%
    {If $\Psi \cong \Phi$ and $A \equiv C$, then $ \beliefsOf{\forgetBy{\Psi}{A}} = \beliefsOf{\forgetBy{\Phi}{C}} $}%
    {If two equivalent formulas are forgotten in  two equivalent epistemic states, then  their posterior belief sets after forgetting are
    identical.}%
\end{postulate}
\begin{figure}[p]
	\centering
		\begin{tabular}[t]{lp{11.0cm}}
			\toprule
			\multicolumn{2}{c}{success postulates} \\\midrule
			\agm{3}  & $\FARBE{ \text{If } A \not\equiv\ \top \text{, then } A \not\in \beliefsOf{\forgetBy{\Psi}{A}} }$ \\[0.25em]
			\midrule\multicolumn{2}{c}{belief related postulates} \\\midrule
			\propW        & $\Psi \nmmodels \forgetBy{\Psi}{A}$                                                                                                                                              \\[0.25em]
			\agm{1}  & $ \beliefsOf{\forgetBy{\Psi}{A}} \subseteq \beliefsOf{\Psi} $ \\[0.25em]
			\agm{2}  & $\text{If } A \notin \beliefsOf{\Psi} \text{, then } \beliefsOf{\Psi} \subseteq \beliefsOf{\forgetBy{\Psi}{A}} $ \\[0.25em]
			\agm{4}  & $ \beliefsOf{\Psi} \subseteq \Cn(\beliefsOf{\forgetBy{\Psi}{A}}\cup \{A\}) $ \\[0.25em]
			\FARBE{\propWCSigma}      & $ \margsub{\beliefsOf{\Psi}}{\signatureOf{\Psi}\setminus\signatureMinOf{A}} \FARBE{\subseteqCNsig{\signatureOf{\Psi}}} \beliefsOf{\forgetBy{\Psi}{A}} $                                                                                 \\[0.25em]
			\FARBE{\propSCSigma}      & $ \beliefsOf{\forgetBy{\Psi}{A}} \FARBE{\subseteqCNsig{\signatureOf{\Psi}}}  \margsub{\beliefsOf{\Psi}}{\signatureOf{\Psi}\setminus\signatureMinOf{A}} $                                                                                 \\[0.25em]
			\FARBE{\propCPSigma}      & $ \beliefsOf{\forgetBy{\Psi}{A}} \FARBE{\gleichCNsig{\signatureOf{\Psi}}} \margsub{\beliefsOf{\Psi}}{\signatureOf{\Psi}\setminus\signatureMinOf{A}} $                                                                                         \\[0.25em]
			\midrule\multicolumn{2}{c}{equivalence related postulates} \\\midrule
			\agm{5}  & $ \text{If } A\equiv C \text{, then }  \beliefsOf{\forgetBy{\Psi}{A}} = \beliefsOf{\forgetBy{\Psi}{C}} $ \\[0.25em]
			\propWE       & If $ \beliefsOf{\Psi} = \beliefsOf{\Phi} $ then $ \beliefsOf{\forgetBy{\Psi}{A}} = \beliefsOf{\forgetBy{\Phi}{A}} $                                                         \\[0.25em]
			\propE        & If $\Psi \cong \Phi$, then $(\forgetBy{\Psi}{A} \cong \forgetBy{\Phi}{A})$                                                                                                       \\[0.25em]
			\propBE       & If $\Psi \cong \Phi$, then $ \beliefsOf{\forgetBy{\Psi}{A}} = \beliefsOf{\forgetBy{\Phi}{A}} $                                                                              \\[0.25em]
			\propDE       & If $\Psi \cong \Phi$ and $A \equiv C$, then   $ \beliefsOf{\forgetBy{\Psi}{A}} = \beliefsOf{\forgetBy{\Phi}{C}} $                                                           \\[0.25em]
			\propOI       & $\forgetBy{(\forgetBy{\Psi}{A})}{B} \cong \forgetBy{(\forgetBy{\Psi}{B})}{A}$                                                                                                    \\[0.25em]
			\midrule\multicolumn{2}{c}{signature related postulates} \\\midrule
			\propPP       & If $\Psi \nmmodels \Psi'$ where $\signatureOf{\Psi'} \subseteq \Sigma'$, and $\signatureMinOf{A} \subseteq \signatureOf{\Psi} \backslash \Sigma'$, then $ \forgetBy{\Psi}{A} \nmmodels \Psi'$               \\[0.25em]
			\propNP       & If $\Psi \not\nmmodels \Psi'$ where $\signatureOf{\Psi'} \subseteq \Sigma'$, and $\signatureMinOf{A} \subseteq \signatureOf{\Psi} \backslash \Sigma'$, then $ \forgetBy{\Psi}{A} \not\nmmodels \Psi'$       \\[0.25em]
			\propPPcond   & If $\signatureOf{\Psi} = \Sigma_1 \cupdisjoint \Sigma_2$, $B,C \in \cL_{\Sigma_1}$ and $A \in \cL_{\Sigma_2}$,\newline then $\Psi \models (C|B)$ implies $ \forgetBy{\Psi}{A} \models (C|B)$     \\[1.25em] 
			\propNPcond   & If $\signatureOf{\Psi} = \Sigma_1 \cupdisjoint \Sigma_2$, $B,C \in \cL_{\Sigma_1}$ and $A \in \cL_{\Sigma_2}$,\newline then $ \forgetBy{\Psi}{A} \models (C|B)$ implies $\Psi \models (C|B)$     \\[1.25em]
			\propEP       & If $\signatureOf{\Psi} = \Sigma_1 \cupdisjoint \Sigma_2$ and $A \in \cL_{\Sigma_2}$, then $\margsub{\Psi}{{\Sigma_1}} = \margsub{(\forgetBy{\Psi}A)}{\Sigma_1}$                         \\[0.25em]
			\propBP       & If $\signatureOf{\Psi} = \Sigma_1 \cupdisjoint \Sigma_2$ and $A \in \cL_{\Sigma_2}$, then $ \margsub{\beliefsOf{\Psi}}{\Sigma_1}  = \margsub{\beliefsOf{\forgetBy{\Psi}A}}{\Sigma_1}$   \\[0.25em] 
			\midrule\multicolumn{2}{c}{structure related postulates} \\\midrule
			\agm{6}  & $ \beliefsOf{\forgetBy{\Psi}{A}} \cap \beliefsOf{\forgetBy{\Psi}{C}} \subseteq \beliefsOf{\forgetBy{\Psi}{A\land C}} $ \\[0.25em] 
			\agm{7}  &  $ \text{If }  A \notin \beliefsOf{\forgetBy{\Psi}{A\land C}} \text{, then }    \beliefsOf{\forgetBy{\Psi}{A\land C}} \subseteq \beliefsOf{\forgetBy{\Psi}{A}} $\\[0.25em] 
			\eqref{pstl:CF}  & $ \beliefsOf{\forgetBy{\Psi}{A\land C}}{=}\beliefsOf{\forgetBy{\Psi}{A}} \text{ or } \beliefsOf{\forgetBy{\Psi}{A\land C}}{=}\beliefsOf{\forgetBy{\Psi}{C}} \text{ or } \beliefsOf{\forgetBy{\Psi}{A\land C}}{=}\beliefsOf{\forgetBy{\Psi}{A}}\cap \beliefsOf{\forgetBy{\Psi}{C}} $ \\[0.25em] 
			\bottomrule
		\end{tabular}
	\caption{Overview of all postulates for forgetting considered in Section \ref{sec:general_props}.} \label{fig:allthepostulates}
\end{figure}

An overview of the postulates considered above is given in Figure~\ref{fig:allthepostulates}.
These postulates will prove useful for evaluating  forgetting operators, and for explaining differences between them. But before we check these properties for the forgetting operators from Section \ref{sec:kindsOfForgetting} in the OCF framework to be presented in Section \ref{sec:instantiating}, we work out relationships between the postulates to make evaluation easier.
 
\subsection{Relationships Among Postulates}
\label{sec_relation_postulates}

The postulates presented in the previous subsections are not independent, but
various relationships hold among them. We show some of the most important relationships in this subsection, and start with relationships among ASP-inspired postulates.

\begin{proposition}
\label{prop_relations_postulates_asp}
	The following relations among ASP-inspired postulates hold:
	\begin{enumerate}[(1)]
\FARBE{
	\item \propSCSigma and \propWCSigma together are equivalent to \propCPSigma.
	\item \propCPSigma implies \propBP.
\item \propDE implies \propBE.
\FARBE{\item \propE implies \propBE.}
}
\end{enumerate}
\end{proposition}
\begin{proof}
	We show the relationships among postulates point by point.
\begin{itemize}

	\item[]\hspace{-2em}\emph{Statement (1)  \enquote{\propSCSigma \& \propWCSigma \( \Leftrightarrow \) \propCPSigma\!\!}.}
	This statement can be easily obtained via basic set theory.

	\item[]\hspace{-2em}\emph{Statement (2)  \enquote{\propCPSigma \( \Rightarrow \) \propBP\!\!}.} 
	Note that $\beliefsOf{\forgetBy{\Psi}{A}} \gleichCNsig{\Sigma} \margsub{\beliefsOf{\Psi}}{\Sigma\setminus\signatureMinOf{A}}$ holds if and only if $\Mod_{\Sigma}(\beliefsOf{\forgetBy{\Psi}{A}}) = \Mod_{\Sigma}(\margsub{\beliefsOf{\Psi}}{\Sigma\setminus\signatureMinOf{A}})$. Equal models assign the same truth values to all atoms and are, therefore, still equal when restricted to subsignatures. Therefore, $\margsub{\beliefsOf{\forgetBy{\Psi}{A}}}{\Sigma\setminus\signatureMinOf{A}} 
	\FARBE{\gleichCNsig{\Sigma\setminus\signatureMinOf{A}}}
	\margsub{\beliefsOf{\Psi}}{\Sigma\setminus\signatureMinOf{A}}$. Now, if $\signatureOf{\Psi} = \Sigma_1 \cupdisjoint \Sigma_2$ and $A \in \cL_{\Sigma_2}$, then $\Sigma\setminus\signatureMinOf{A} \subseteq \Sigma_1$. Hence, we can restrict the models even further to obtain $\margsub{\beliefsOf{\forgetBy{\Psi}{A}}}{\Sigma_1} 
	\FARBE{\gleichCNsig{\Sigma_1}} 
	\margsub{\beliefsOf{\Psi}}{\Sigma_1}$. As the marginalized belief sets are deductively closed over the subsignature $\Sigma_1$ and have the same models over $\Sigma_1$, we can conclude $\margsub{\beliefsOf{\forgetBy{\Psi}{A}}}{\Sigma_1} = \margsub{\beliefsOf{\Psi}}{\Sigma_1}$.
		
    \item[]\hspace{-2em} \emph{Statement (3)  \enquote{\propDE \( \Rightarrow \) \propBE\!\!}.} 
	By setting \( C=A \) in \propDE, we obtain \propBE directly.
    
        \FARBE{\item[]\hspace{-2em} \emph{Statement (4)  \enquote{\propE \( \Rightarrow \) \propBE\!\!}.} 
    Observe that \( \beliefsOf{\forgetBy{\Psi}{A}} = \beliefsOf{\forgetBy{\Phi}{A}} \) \FARBE{holds if for all \( C \in \mathcal{L} \) we have:}
    \begin{equation*}
        (C|\top) \in \Cons(\forgetBy{\Psi}{A}) \text{ if and only if } 
        (C|\top) \in \Cons(\forgetBy{\Phi}{A}) 
    \end{equation*}
    According to \propE, we have that \( \Psi \cong \Phi \) implies \( \forgetBy{\Psi}{A} \cong \forgetBy{\Phi}{A}  \).
    The latter is equivalent to stating \( \Cons(\forgetBy{\Psi}{A}) = \Cons(\forgetBy{\Phi}{A})  \).
    Thus, when \propE is given, \( \Psi \cong \Phi \) implies \( \beliefsOf{\forgetBy{\Psi}{A}} = \beliefsOf{\forgetBy{\Phi}{A}} \).
    Consequently, we have that \propE implies \propBE.
    }
\qedhere
\end{itemize}
\end{proof}

We recall well-known results from belief revision theory \cite{KS_Hansson1999} which carry over immediately to our framework.

\begin{proposition}
\label{prop_relations_postulates_agm}
\propCF implies \agm{6}.
\FARBE{If \agm{1} -- \agm{5} are given},
then \eqref{pstl:CF} is equivalent to the conjunction of \agm{6} and \agm{7}. 
\end{proposition}
In the next proposition, we consider relationships among ASP-inspired postulates.The following Proposition

\begin{proposition}
\label{prop_relations_postulates_mixed}
	The following relations among ASP-inspired postulates on the one hand and AGM-inspired postulates on the other hand hold:
\FARBE{
	\begin{enumerate}[(1)]
\item %
\propW implies \agm{1}.
\item \propSCSigma implies \agm{1}.
\item \FARBE{If $A \not\in \beliefsOf{\Psi}$, then}
  \agm{2} implies \propWCSigma.
\item 
If $A$ is contingent then
\propSCSigma implies \agm{3}.
\item \propDE implies \agm{5}.
\end{enumerate}}
\end{proposition}

\begin{proof}
	We show the relationships among postulates point by point.
	\begin{itemize}
		\item[]\hspace{-2em}\emph{Statement (1) \enquote{\propW \( \Rightarrow \) \agm{1}}.}
    We obtain (1) by expanding the definition of the entailment relation: $\Psi \nmmodels \forgetBy{\Psi}{A}$ means that $\Cons(\forgetBy{\Psi}{A}) \subseteq \Cons(\Psi)$.
    It follows that $\forgetBy{\Psi}{A} \models (A|\top)$ implies $\Psi \models (A|\top)$, which is equivalent to $\beliefsOf{\forgetBy{\Psi}{A}} \subseteq \beliefsOf{\Psi}$.  

	\item[]\hspace{-2em}\emph{Statement (2) \enquote{\propSCSigma \( \Rightarrow \) \agm{1}}.}
	For (2), first note that $\margsub{\beliefsOf{\Psi}}{\Sigma\setminus\signatureMinOf{A}} \subseteq \beliefsOf{\Psi}$ and, therefore, $\margsub{\beliefsOf{\Psi}}{\Sigma\setminus\signatureMinOf{A}} \subseteqCNsig{\Sigma} \beliefsOf{\Psi}$. With \propSCSigma we now obtain $\beliefsOf{\forgetBy{\Psi}{A}} \subseteqCNsig{\Sigma} \beliefsOf{\Psi}$ due to the transitivity of $\subseteqCNsig{\Sigma}$. Because $\beliefsOf{\Psi}$ is deductively closed over $\Sigma$, this is equivalent to $\beliefsOf{\forgetBy{\Psi}{A}} \subseteq \beliefsOf{\Psi}$.

    \item[]\hspace{-2em}\emph{Statement (3) \enquote{If \( A\notin\beliefsOf{\Psi} \), then \agm{2} \( \Rightarrow \) \propWCSigma}.}
    \FARBE{Note that \linebreak $\margsub{\beliefsOf{\Psi}}{\Sigma\setminus\signatureMinOf{A}} 
    \subseteqCNsig{\Sigma}
    \beliefsOf{\Psi}$, \FARBE{due to Proposition \ref{prop:bsmargrepresentation_extended}}. If $A \notin \beliefsOf{\Psi}$, then $\beliefsOf{\Psi} \subseteq \beliefsOf{\forgetBy{\Psi}{A}}$ according to \agm{2}.  Therefore, $\margsub{\beliefsOf{\Psi}}{\Sigma\setminus\signatureMinOf{A}} \subseteq \beliefsOf{\forgetBy{\Psi}{A}}$,
    hence also 
    $\margsub{\beliefsOf{\Psi}}{\Sigma\setminus\signatureMinOf{A}} 
    \subseteqCNsig{\Sigma}    
    \beliefsOf{\forgetBy{\Psi}{A}}$,
    i.e., \propWCSigma holds.}
    
    \item[]\hspace{-2em}\emph{Statement (4) \enquote{If \( A \) is contingent, then \propSCSigma \( \Rightarrow \) \agm{3}}.}
    \FARBE{For (4), \propSCSigma yields $\beliefsOf{\forgetBy{\Psi}{A}} 
     \subseteqCNsig{\Sigma}     
    \margsub{\beliefsOf{\Psi}}{\Sigma\setminus\signatureMinOf{A}}$, and due to Proposition \ref{prop:bsmargrepresentation_extended}, we have 
    \( A\notin \margsub{\beliefsOf{\Psi}}{\Sigma\setminus\signatureMinOf{A}}\); consequently, \agm{3} is satisfied.}
    \item[]\hspace{-2em}\emph{Statement (5) \enquote{\propDE \( \Rightarrow \) \agm{5}}.}
    Claim (5) is immediately given by setting $\Psi = \Phi$ in \propDE.  
\qedhere
\end{itemize}
\end{proof}

Now, we look more closely into relations among the Persistence postulates,
proving that  the conditional versions of the Persistence postulates are equivalent to the respective original postulates, and that  the \propEP postulate has a central position among the Persistence postulates. 

\begin{proposition}
 \label{prop_condppnp}
 \label{prop_epppnp}
 \label{prop_epbp}
 The following relationships hold among the Persistence properties: 
\FARBE{
\begin{enumerate}
  \item \FARBE{\propPPcond is equivalent to \propPP, and \propNPcond is equivalent to \propNP.}
  \item 
\propEP implies both \propPP and \propNP. 
 \item  \propEP implies \propBP.
 \end{enumerate}
 }
\end{proposition}

\begin{proof}
We start with proving the first statement (1). 
 Let $\signatureOf{\Psi}  = \Sigma_1 \cupdisjoint \Sigma_2$, $\signatureOf{\Psi'}  = \Sigma_1$ (w.l.o.g.), $B,C \in \cL_{\Sigma_1}$, and $A \in \cL_{\Sigma_2}$. 
 
 We first prove that \propPPcond\ implies \propPP. Let $\Psi \nmmodels \Psi'$, i.e., $\Cons(\Psi') \subseteq \Cons(\Psi)$. We have to show that $ \forgetBy{\Psi}{A} \nmmodels \Psi'$, i.e., $\Cons(\Psi') \subseteq \Cons(\forgetBy{\Psi}{A})$. Let $(C|B) \in \Cons(\Psi')$, then also $(C|B) \in \Cons(\Psi)$ which means $\Psi \models (C|B)$. By \propPPcond, also $\forgetBy{\Psi}{A}   \models (C|B)$, and therefore $(C|B) \in \Cons(\forgetBy{\Psi}{A})$. Conversely, we presuppose \propPP\ and show that \propPPcond\ holds. So, let $\Psi \models (C|B)$, then by definition also $\margsub{\Psi}{{\Sigma_1}} \models (C|B)$. Furthermore, clearly,  $\Psi \nmmodels \margsub{\Psi}{{\Sigma_1}}$, so the preconditions of \propPP\ are satisfied, and hence
 $\forgetBy{\Psi}{A} \nmmodels \margsub{\Psi}{{\Sigma_1}}$ and $\margsub{\Psi}{{\Sigma_1}} \models (C|B)$,
 which yields 
 $\forgetBy{\Psi}{A} \models (C|B)$. 
 
 The proof regarding \propNP is very similar, so we abbreviate it a bit; please note that we make use of the contrapositive form of \propNP. %
 \FARBE{First, we show that \propNPcond\ implies \propNP. }
  $ \forgetBy{\Psi}A \nmmodels \Psi'$ and $\Psi' \models (C|B)$ implies $ \forgetBy{\Psi}A  \models (C|B)$, hence $ \Psi  \models (C|B)$. Because this holds for arbitrary $B,C \in \cL_{\Sigma_1}$, we obtain $\Psi \nmmodels \Psi'$. 
  \FARBE{Conversely, we prove that \propNP\ implies \propNPcond.} Again via marginalization, $ \forgetBy{\Psi}A  \models (C|B)$ implies $\margsub{{(\forgetBy{\Psi}A)}}{{\Sigma_1}} \models (C|B)$, and because  $ \forgetBy{\Psi}A \nmmodels \margsub{{(\forgetBy{\Psi}A)}}{{\Sigma_1}}$, we obtain $ \Psi \nmmodels \margsub{{(\forgetBy{\Psi}A)}}{{\Sigma_1}}$, and therefore $\Psi \models (C|B)$. 
 \medskip

Now we focus on \propEP, proving statement (2) first. 
 Let $\signatureOf{\Psi}  = \Sigma_1 \cupdisjoint \Sigma_2$, $\signatureOf{\Psi'}  = \Sigma_1$ (w.l.o.g.), 
 and $A \in \cL_{\Sigma_2}$. We presuppose that \propEP\ holds. 
  
In order to prove that then, also \propPP\ and \propNP\ hold, we make use of the following equivalences: $\Psi \nmmodels \Psi'$ iff $\margsub{\Psi}{{\Sigma_1}}\nmmodels \Psi'$ iff, via \propEP,  $\margsub{{(\forgetBy{\Psi}A)}}{{\Sigma_1}}\nmmodels \Psi'$ iff $\forgetBy{\Psi}A \nmmodels \Psi'$. 
\medskip

Finally, we show the relationship between \propEP and \propBP, proving statement (3). 
	Suppose \( \signatureOf{\Psi}=\Sigma_1 \cupdisjoint \Sigma_2 \) and \( A\in \cL_{\Sigma_2} \). By \propEP we have \( \margsub{\Psi}{{\Sigma_1}} = \margsub{(\forgetBy{\Psi}A)}{\Sigma_1} \).
		Note that by Proposition \ref{prop:es_synmag} we have  \( \beliefsOf{\margsub{\Psi}{{\Sigma_1}}}=\margsub{\beliefsOf{\Psi}}{{\Sigma_1}} \) and \( \beliefsOf{\margsub{(\forgetBy{\Psi}A)}{{\Sigma_1}}}=\margsub{\beliefsOf{(\forgetBy{\Psi}A)}}{{\Sigma_1}} \).
		All observations together imply the desired result.
\end{proof}

\FARBE{Propositions~\ref{prop_relations_postulates_asp} and \ref{prop_relations_postulates_mixed} reveal links between AGM-inspired and ASP-inspired postulates and show that some of the relations that hold for the ASP versions of the postulates also hold for their epistemic interpretations (e.g., quite trivially, the conjunction of \propSCSigma and \propWCSigma is equivalent to \propCPSigma). However, in general, the epistemic versions of the ASP postulates allow for a deeper analysis of forgetting. Therefore,  differences between the postulates become apparent that are not visible on the level of answer sets.}
For example, in the \FARBE{world of ASP forgetting}, \propW is equivalent to \propNP \FARBE{\cite{KS_GoncalvesKnorrLeite2016a}},
but this no longer holds for the corresponding \FARBE{general} epistemic versions. More precisely, \propW is strictly stronger than \propNP here, as the next proposition shows:

\begin{proposition}
 \label{prop_npw}
 \propW implies \propNP, but not the other way round. 
\end{proposition}
\begin{proof}
  We assume that \propW\ holds, i.e., $\Psi \nmmodels \forgetBy{\Psi}A$. Let $\signatureOf{\Psi'}  \subseteq \Sigma'$,
  \FARBE{$\signatureOf{A} \subseteq \Sigma \backslash \Sigma'$,}
  and $ \forgetBy{\Psi}A \nmmodels \Psi'$. By transitive chaining of $\nmmodels$ resp.\ of the  subset relations of the respective sets of conditionals, this  implies $\Psi \nmmodels \Psi'$. Therefore, \propNP\ holds. 
 
 The converse is not true, \propNP\ cannot imply \propW\ in general because \propNP\ makes use of specific prerequisites which are not present in \propW. For a counterexample in the context of OCFs,
 please see \FARBE{Example~\ref{ex_np_not_implies_w} in the \hyperref[app_examples_c-contractions]{Appendix}}.
\end{proof}

\FARBE{After having presented and discussed the postulates for general epistemic states, in the next section, we are now going to instantiate forgetting operations within the framework of ranking functions that take the role of epistemic states. Ranking functions provide a particularly popular type of epistemic states that is widely used in nonmonotonic reasoning and belief revision, with natural definitions of conditionalization and marginalization, similar to probability theory \cite{spohn12}. We show that all kinds of forgetting presented in Section \ref{sec:kindsOfForgetting} can be realized with the help of ranking functions. This allows for evaluating all types of forgetting operators in a common semantic framework with respect to the postulates presented in this section, providing detailed and expressive comparisons among the different forgetting operators. }

\section{Instantiating the Framework by Ranking Functions}
\label{sec:instantiating}
Ordinal conditional functions,
or ranking functions $\kappa$ over $ \Sigma $ provide all technical features that we expect from epistemic states. 
In the following subsections, we recall and extend the instantiations of the basic forgetting operators described above 
within the framework of OCFs from \cite{BeierleKernIsbernerSauerwaldBockRagni2019KIzeitschrift}.
The object $A$ which is to be forgotten can be a formula from $\mathcal{L}$, and thus in particular also a variable from $\Sigma$. 

\FARBE{We start with elaborating on some technical results regarding the equivalence of ranking functions which are useful in the context of this article.}

\subsection{Equivalence of OCFs}
\label{sec_ocf_equivalence}

Two ranking functions $\kappa$ and $\kappa^\prime$ are equivalent as epistemic states, i.e., $\kappa \cong \kappa'$, iff
$\Cons(\kappa) = \Cons(\kappa')$ (cf.\ Section~\ref{subsec_kindsofforgetting}), i.e., iff 
for all conditionals $(B|A)$ it holds that $\kappa \models (B|A)$ iff $\kappa^\prime \models (B|A)$. Note that this
coincides with 
the notion of \emph{inferential equivalence}, as given in \cite[Def.~4]{BeierleKutsch2019AppliedIntelligence}, from where we can derive immediately a handy characterization of the entailment $\kappa \nmmodels \kappa'$ and the equivalence  $\kappa \cong \kappa'$ between ranking functions; the proof of the following proposition is a direct consequence of \cite[Prop.~1]{BeierleKutsch2019AppliedIntelligence}.

\begin{proposition}%
\label{prop:inf_equiv}
For two ranking functions $\kappa,\kappa^\prime$, we have $\kappa \nmmodels \kappa^\prime$ iff for all $\omega_1,\omega_2 \in \Omega$ it holds that $\kappa(\omega_1) \leq \kappa(\omega_2)$ implies $\kappa^\prime(\omega_1) \leq \kappa^\prime(\omega_2)$. Moreover, two ranking functions $\kappa,\kappa^\prime$ are  equivalent, $\kappa \cong \kappa^\prime$, iff for all $\omega_1,\omega_2 \in \Omega$ it holds that $\kappa(\omega_1) \leq \kappa(\omega_2)$ iff $\kappa^\prime(\omega_1) \leq \kappa^\prime(\omega_2)$.
\end{proposition}

Equivalent OCFs may assign different ranks to possible worlds resp.\ formulas, but they preserve the property of a world being a minimal model for a formula.

\FARBE{
\begin{proposition}
 \label{prop_ocf_equiv_minmodel}
Let $\kappa_1  \cong \kappa_2$ be two equivalent OCFs, let $A \in \cL_\Sigma$. 
Then a model $\omega$ of $A$ is a minimal model of $A$ with respect to $\kappa_1$ iff it is a minimal model of $A$ with respect to $\kappa_2$. 
In particular,  \FARBE{$\modBeliefsOf{\kappa_1} = \modBeliefsOf{\kappa_2}$,} and for any model $\omega$ of $A$,  $\kappa_1(\omega) = \kappa_1(A)$ iff $\kappa_2(\omega) = \kappa_2(A)$.
\end{proposition}

\begin{proof}
 Let $\omega \models A$ be a minimal model of $A$ with respect to $\kappa_1$, i.e., $\kappa_1(\omega) \leq \kappa_1(\omega')$ for all models $\omega'$ of $A$. Due to Proposition \ref{prop:inf_equiv}, then also $\kappa_2(\omega) \leq \kappa_2(\omega')$ for all models $\omega'$ of $A$ holds, and therefore $\omega$ is also a minimal model of $A$ with respect to $\kappa_2$. 
 
 Since $\modBeliefsOf{\kappa_1}, \modBeliefsOf{\kappa_2}$ are the respective minimal models of a tautology $\top$, $\modBeliefsOf{\kappa_1} = \modBeliefsOf{\kappa_2}$ follows as a special case. 
 
 Finally, observing that $\kappa_i(\omega) = \kappa_i(A)$ holds if and only if $\omega$ is a minimal model of $A$ with respect to $\kappa_i$, $i \in \{1,2\}$, this yields $\kappa_1(\omega) = \kappa_1(A)$ iff $\kappa_2(\omega) = \kappa_2(A)$.
\end{proof}
}

\FARBE{As an immediate consequence of} \FARBE{Proposition~\ref{prop_ocf_equiv_minmodel},}
\FARBE{we obtain that $\beliefsOf{\kappa}$ is identical for all inferentially equivalent $\kappa$.}

\begin{proposition}
 \label{prop_ocf_equiv_implies_bel_equal}
\FARBE{If $\kappa_1 \cong \kappa_2$ for OCFs $\kappa_1, \kappa_2$, then $\beliefsOf{\kappa_1} = \beliefsOf{\kappa_2}$.}
\end{proposition}

An especially interesting form of equivalence can be obtained by considering the multiples of an OCF $\kappa$, i.e., the set of OCFs $q \cdot \kappa$ for any positive rational number $q$. Note that we allow $q$ to be rational, but $q \cdot \kappa$ must be an OCF in the end, yielding only natural numbers for ranks. It is obvious, that $\kappa$ and any of its multiples $q \cdot \kappa$ are equivalent.

\begin{definition}[\FARBE{\cite{AH_KernIsberner-etal_2024-KR}}]
 \label{def_ocf_xequiv}
 Two OCFs $\kappa_1, \kappa_2$ \FARBE{over $\Omega_{\Sigma}$} are \emph{\FARBE{linearly} equivalent}, in symbols $\kappa_1 \xequiv \kappa_2$, if there is a positive rational number $q$ such that $\kappa_2 = q \cdot \kappa_1$, \FARBE{i.e., $\kappa_2(\omega) = q \cdot \kappa_1(\omega)$ holds for all $\omega \in \Omega$}.
\end{definition}

It is obvious that $\xequiv$ is also an equivalence relation on OCFs. Suitable representatives of the equivalence classes are, e.g.,  those $\kappa$ where the greatest common divisor of all $\kappa(\omega)$ is $1$. Note that $\xequiv$ is a subrelation of $\cong$, so all statements of Proposition \ref{prop_ocf_equiv_minmodel} also hold for \FARBE{linearly} equivalent OCFs. 

\FARBE{
Proposition \ref{prop_ocf_equiv_minmodel} states that for any equivalent OCFs $\kappa_1, \kappa_2$ and for any model $\omega$ of $A$,  $\kappa_1(\omega) = \kappa_1(A)$ iff $\kappa_2(\omega) = \kappa_2(A)$. From this, however, we cannot derive a relationship between $\kappa_1(A)$ and $\kappa_2(A)$, mainly because of empty layers being possibly present in the OCFs. For the special case of \FARBE{linearly} equivalent OCFs, a useful result can be shown here. 

\begin{lemma}[\FARBE{\cite{AH_KernIsberner-etal_2024-KR}}]
 \label{prop_lin_equiv_ocfs}
If $\kappa_2 = q \cdot \kappa_1$, then $\kappa_2(A) = q \cdot \kappa_1(A)$ for any formula $A$.
\end{lemma}

\begin{proof}
 Let $\omega$ be a minimal model of $A$ with respect to $\kappa_2$, i.e., $\kappa_2(A) = \kappa_2(\omega) = q \cdot \kappa_1(\omega)$. Due to Proposition \ref{prop_ocf_equiv_minmodel}, $\kappa_1(\omega) = \kappa_1(A)$, and the statement of the lemma follows immediately. 
\end{proof}
}

\subsection{Marginalization and Conditionalization}
\label{sec:instantiating_marginalisation_conditionalisation}

OCFs can be seen as qualitative abstractions of probabilities \cite{GoldszmidtPearl96}. Therefore, very similar to probabilities, they also allow for marginalization and conditionalization while observing their specific arithmetic characteristics (cf.\ \cite{Spohn88}).

\begin{definition}[marginalization of $\kappa$ to $\Sigma^\prime$ \cite{BeierleKernIsbernerSauerwaldBockRagni2019KIzeitschrift}]\label{def:marginalization}
	Let $ \kappa $ be an OCF over $ \Sigma $ and $ \Sigma'\subseteq\Sigma $. 
	The {marginalization of $ \kappa $ to $ \Sigma' $}, denoted by $ \margsub{\kappa}{\Sigma'}:\Omega_{\Sigma'}\to\naturals $, is given by
	\begin{equation}\label{eq_marginalization_kappa}
	\margsub{\kappa}{\Sigma'}(\omega') =\min\{ \kappa(\omega) \mid \omega\in\Omega_\Sigma \text{ and } \omega\models\omega' \}.
	\end{equation}
\end{definition}
\noindent Applying this to implement marginalization as a forgetting operation, we obtain the operation of \emph{forgetting by OCF-marginalization}: 
\FARBE{%
\begin{equation}\label{eq_forgetting_by_OCF_marginalization}
\kappa^{\circ}_A = \margsub{\kappa}{\Sigma \setminus \FARBE{\signatureMinOf{A}}}
\end{equation}
}

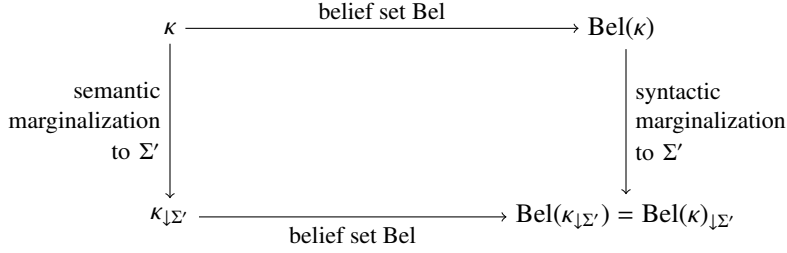
\begin{figure}\centering
    \begin{tikzpicture}
        \node (start)  at (0,0){$\kappa$};
        \node [text width=1cm] (ur) at (6,0) {\( \beliefsOf{\kappa} \)};
        \node (ll)  at (0,-2.5) {$\margsub{\kappa}{\Sigma^\prime}$};
        \node (lr) at (6,-2.5) {$\beliefsOf{\margsub{\kappa}{\Sigma^\prime}} = \margsub{\beliefsOf{\kappa}}{\Sigma^\prime}$};
        
        \draw (start) edge[->] node [above] {\small belief set \( \beliefssymbol \)} (ur);
        \draw (start) edge[->] node [left,text width=2cm,align=right] {\small \FARBE{semantic}\\ marginalization\\ to \( \Sigma' \)} (ll);
        \draw (ur) edge[->] node [right,text width=2cm] {\small syntactic\\ marginalization\\ to \( \Sigma' \)} (lr);
        \draw (ll) edge[->] node [below] {\small belief set \( \beliefssymbol \)} (lr);
    \end{tikzpicture}\caption{Semantic compatibility between marginalization of a ranking function \( \kappa \) and marginalization of its respective belief set \( \beliefsOf{\kappa} \), \FARBE{cf.\ Proposition~\ref{prop:kappa_synmag}}.}\label{fig:kappa_marginalisation_bs_diagram}
\end{figure}

Recall that Proposition \ref{prop:es_synmag} shows that the marginalization of epistemic states is compatible with the marginalization of their respective beliefs sets. 
Because ranking functions fit so well into the framework of epistemic states, Proposition \ref{prop:es_synmag} carries over to ranking functions.
Figure \ref{fig:kappa_marginalisation_bs_diagram} illustrates the connection between marginalization of a ranking function \( \kappa \) and its respective beliefs set  \( \beliefsOf{\kappa} \) stated in the following proposition \FARBE{by Sauerwald et al. \cite{SauerwaldKernIsbernerBeckerBeierle2022SUM}}.

\begin{proposition}[\cite{SauerwaldKernIsbernerBeckerBeierle2022SUM}]\label{prop:kappa_synmag}
    Let \( \kappa \) \FARBE{be an} OCF with \( \signatureOf{\kappa}=\Sigma \) and \( \Sigma' \subseteq \Sigma \).
    Then we have \( 
    \beliefsOf{\margsub{\kappa}{\Sigma'}} = \margsub{\beliefsOf{\kappa}}{\Sigma'} \).
\end{proposition}
\FARBE{\begin{proof}
	This is a direct consequence of Proposition~\ref{prop:es_synmag}.
\end{proof}
}
\FARBE{\noindent Proposition~\ref{prop:kappa_synmag} gives rise to the following property.
\begin{proposition}\label{prop:kappa_synmag_extended}
	For each OCF \( \kappa \)  with \( \signatureOf{\kappa}=\Sigma \) and \( A\in\mathcal{L}_{\Sigma} \), we have
	\begin{equation*}
		\beliefsOf{\margsub{\kappa}{\Sigma\setminus\signatureMinOf{A}}} = \beliefsOf{\kappa} \cap \mathcal{L}_{\Sigma \setminus \signatureMinOf{A}} \ .
	\end{equation*}
\end{proposition}
\begin{proof}
	At first, recall that \( \kappa^{\circ}_A = \margsub{\kappa}{\Sigma \setminus \signatureMinOf{A}} \) holds (see Equation~\ref{eq_forgetting_by_OCF_marginalization}).
	From Proposition~\ref{prop:kappa_synmag}, we obtain \( \beliefsOf{\forgetBy{\kappa}{A}} = \margsub{\beliefsOf{\kappa}}{\Sigma \setminus \signatureMinOf{A}}  \), and due to Proposition~\ref{prop:bsmargrepresentation_extended}, we have \( \margsub{\beliefsOf{\kappa}}{\Sigma \setminus \signatureMinOf{A}} = \beliefsOf{\kappa} \cap \mathcal{L}_{\Sigma \setminus \signatureMinOf{A}} \).
	By combining the last two observations, we obtain the statement.
\end{proof}}

\FARBE{Furthermore, inferential equivalence of ranking functions is preserved under marginalization.

\begin{proposition}
 \label{prop_ocf_equiv_preserved_under_marginalization}
 Let $\kappa_1, \kappa_2$ be OCFs with  \( \signatureOf{\kappa_1}=  \signatureOf{\kappa_2}=\Sigma \), and let $\Sigma' \subseteq \Sigma$.
 Then 
$\kappa_1 \cong \kappa_2$
implies
$\margsub{\kappa_1}{\Sigma'} \cong \margsub{\kappa_2}{\Sigma'}$.
\end{proposition}}

\FARBE{
\begin{proof}
The proof is obtained by a straightforward application of Proposition~\ref{prop_ocf_equiv_minmodel}.
\end{proof}
}

\begin{definition}[conditionalization of $\kappa$ by $A$ \cite{Spohn88}]\label{def:conditionalization}
	Let $ \kappa $ be a ranking function and $ A $ a proposition, then the conditionalization of $ \kappa $ by $ A $ is the ranking function \FARBE{$ \kappa|A: \Mod(A)\to \naturals $}, defined on the models of $ A $ as follows:
	\begin{equation}\label{eq_def_condionalization}
	\kappa|A (\omega) = \kappa(\omega) - \kappa(A)
	\end{equation}
\end{definition}

\noindent This yields the operation of \emph{forgetting by OCF-conditionalization}:
\FARBE{%
\begin{equation}\label{eq_forgetting_by_OCF_condionalization}
  \kappa^{\circ}_A = \kappa|\overline{A}
\end{equation}
}

The following proposition collects useful basic statements on marginalization and conditionalization:

\begin{proposition}
 \label{prop_basics_marginalization_conditionalization}
 Let $\Sigma' \subseteq \Sigma$ be signatures, let $\kappa$ be an OCF over $\Sigma$. 
  \begin{enumerate}
  \item \FARBE{For any $A \in \cL_{\Sigma}$, and} for any $C \models A$, we have $ \kappa|A(C) = \kappa(C) - \kappa(A)$.
  \item For any $A \in \cL_{\Sigma'}$, we have $ \margsub{\kappa}{\Sigma'}(A) = \kappa(A)$. 
 \item \FARBE{Marginalized OCFs are entailed by the original OCF, i.e., 
  $\kappa \nmmodels \margsub{\kappa}{\Sigma'}$.} 
 \end{enumerate}
\end{proposition}
\FARBE{
\begin{proof}
 We show (1). Let $A \in \cL_{\Sigma}$ and  $C \models A$. By definition, $\kappa|A (C) = \min\{\kappa|A (\omega) \mid \omega \models C\} 
 = \min\{\kappa(\omega) - \kappa(A) \mid \omega \models C\}
 = \min\{\kappa(\omega) \mid \omega \models C\} - \kappa(A) = \kappa(C) - \kappa(A)$.

 For (2), consider $A \in \cL_{\Sigma'}$. We have by definition $\margsub{\kappa}{\Sigma'}(A) = \min \{\margsub{\kappa}{\Sigma'}(\omega') \mid \omega'\in\Omega_{\Sigma'}, \omega' \models A\} =\min\{ \kappa(\omega) \mid \omega\in\Omega_\Sigma, \omega'\in\Omega_{\Sigma'}, \omega\models\omega' \models A \} = \min\{ \kappa(\omega) \mid \omega\in\Omega_\Sigma, \omega \models A \} = \kappa(A)$. 
 
 (3) is now an easy consequence of  (2).
 We have to show that $\Cons(\margsub{\kappa}{\Sigma'})=\{(B'|A')  \in (\cL_{\Sigma'} | \cL_{\Sigma'}) \mid \margsub{\kappa}{\Sigma'} \ESmodels (B'|A')\}\subseteq \Cons(\kappa)$.
 So, let $\margsub{\kappa}{\Sigma'} \ESmodels (B'|A'), A',B' \in \cL_{\Sigma'}$, i.e., $\margsub{\kappa}{\Sigma'}(A'B') < \margsub{\kappa}{\Sigma'}(A'\ol{B'})$. Then $A', B'$ are also in $\cL_{\Sigma}$,
 \FARBE{and by (2),}
 we obtain $\kappa(A'B') = \margsub{\kappa}{\Sigma'}(A'B') < \margsub{\kappa}{\Sigma'}(A'\ol{B'}) = \kappa(A'\ol{B'})$, so $\kappa \models (B'|A')$, and hence $(B'|A') \in \Cons(\kappa)$. 
 \end{proof}
 
Note that for conditionalization,
\FARBE{the property corresponding to
  statement (3) for marginalization}
  of Proposition \ref{prop_basics_marginalization_conditionalization} does not hold, i.e., OCFs do not entail their conditionalizations in general. This is  because conditionalizations only reflect relationships on some part of the models that may not hold for the whole set of the models. Counterexamples can be found easily. 
}

\subsection{Lifted OCF-Marginalization}
\label{sec:lifting}

For $\Sigma' \subseteq \Sigma$ and $\omega \in \Omega_{\Sigma}$,
the marginalization 
of $\omega$ to $\Omega'$ is the $\Sigma'$-part
of $\omega$,
i.e., $\margsub{\omega}{\Sigma'}$ is the projection of $\omega$ to 
the propositional
variables in $\Omega'$ and hence $\margsub{\omega}{\Sigma'} \in \Omega_{\Sigma'}$.
On the other hand, we can lift an OCF $\kappa$ over the worlds of a subsignature
to any larger signature.

\begin{definition}[$\embedsup{\kappa}{\Sigma}$, lifting of $\kappa$ to $\Sigma$]\label{def:liftung}
	Let $ \kappa $ be an OCF over $ \Sigma' $ and $ \Sigma'\subseteq\Sigma $. 
	The {lifting of $ \kappa $ to $ \Sigma $}, denoted by
        $ \embedsup{\kappa}{\Sigma}:\Omega_{\Sigma}\to\naturals $,
        is given by, for any $\omega \in \Omega_{\Sigma}$,
	\begin{equation}
	\label{def_marginalization_lifted}
	 \embedsup{\kappa}{\Sigma}(\omega) = \kappa(\omega^{\Sigma'}).
	\end{equation}
\end{definition}

\FARBE{Analogously to marginalization to a smaller signature, inferential equivalence of ranking functions is preserved under lifting to a \FARBE{larger} signature.

\begin{proposition}\label{prop:isoToLifting}
 \label{prop_ocf_equiv_preserved_under_lifting}
 Let $\kappa_1, \kappa_2$ be OCFs with  \( \signatureOf{\kappa_1}=  \signatureOf{\kappa_2}=\Sigma' \), and let $\Sigma' \subseteq \Sigma$.
 Then 
$\kappa_1 \cong \kappa_2$
implies
$\embedsup{\kappa_1}{\Sigma} \cong \embedsup{\kappa_2}{\Sigma}$.
\end{proposition}

\FARBE{
\begin{proof}
The proof is obtained by a straightforward application of Proposition~\ref{prop_ocf_equiv_minmodel}.
\end{proof}
}

Furthermore,}
for any OCF $ \kappa: \Omega_{\Sigma} \to\naturals $ over $ \Sigma $, we can consider the OCF 
\begin{align}
 \margANDembed{\kappa}{\Sigma'}{\Sigma}:  \Omega_{\Sigma} \to\naturals 
\end{align}
obtained by first marginalizing to $\Sigma'$ and then embedding into $\Sigma$.
\FARBE{For sake of a readable notation, we omit the parentheses around $ \margsub{\kappa}{\Sigma'}$ before lifting it to $\Sigma$ here; this should cause no confusion since both marginalization and lifting can only be applied to ranking functions \FARBE{here.}}
Note the differences among $\kappa, \margsub{\kappa}{\Sigma'}$ and
$\margANDembed{\kappa}{\Sigma'}{\Sigma}$.
Compared to $\kappa$, all propositional variables $a \in \Sigma \setminus \Sigma'$ are forgotten in  $\margsub{\kappa}{\Sigma'}$. On the other hand,
compared to  $\kappa$, no propositional variables  are forgotten in
$\margANDembed{\kappa}{\Sigma'}{\Sigma}$, but for all $\omega \in \Sigma$, the value of
$\margANDembed{\kappa}{\Sigma'}{\Sigma}(\omega)$ is independent
of all variables $a \in \Sigma \setminus \Sigma'$ occurring in $\omega$.

\FARBE{The operation of \emph{forgetting by lifted OCF-marginalization} is then defined via:}
\FARBE{%
\begin{equation}\label{eq_forgetting_by_lifted_OCF_marginalization}
\FARBE{\kappa^{\circ}_A =  \margANDembed{\kappa}{(\Sigma \setminus \signatureMinOf{A})}{\Sigma}}
\end{equation}
}

\begin{proposition}
	Let $ \kappa' $ be an OCF over $ \Sigma' $ and $ \Sigma'\subseteq\Sigma $. Then we have:
\begin{equation*}        
\margsub{\left(\Mod(\Bel(\embedsup{{\kappa^\prime}}{\Sigma}))\right)}{\Sigma'} = \Mod(\beliefsOf{\kappa'})
\end{equation*}       
\end{proposition}

The properties for marginalization given in Proposition \ref{prop_basics_marginalization_conditionalization} also hold for lifted marginalization.
\begin{proposition}\label{prop_basics_lifted_marginalization}
        Let $\Sigma' \subseteq \Sigma$ be signatures, let $\kappa$ be an OCF over $\Sigma$,
\FARBE{and let $A \in \cL_{\Sigma'}$.}  
        Then the following statements hold:
        \begin{enumerate}
            \item $ \margsub{\kappa}{\Sigma'}(A) = \margANDembed{\kappa}{\Sigma'}{\Sigma}(A)$, and
            \item $\margANDembed{\kappa}{\Sigma'}{\Sigma} \nmmodels \margsub{\kappa}{\Sigma'}  $. 
        \end{enumerate}
\end{proposition}

\begin{proof}
\FARBE{The first statement (1) holds according to the following derivation:}
\FARBE{%
\begin{align*}
&\margsub{\kappa}{\Sigma'}(A)
\\
=\ &\min\{\margsub{\kappa}{\Sigma'}(\omega') \mid \omega' \in \Omega_{\Sigma'} \textrm{ and } \omega' \models A\}
\\
=\ &\min\{\margsub{\kappa}{\Sigma'}(\omega^{\Sigma^\prime}) \mid \omega \in \Omega_{\Sigma} \textrm{ and } \omega \models A\}
\\
=\ &\min\{\margANDembed{\kappa}{\Sigma'}{\Sigma}(\omega) \mid \omega \in \Omega_{\Sigma} \textrm{ and } \omega \models A\}
\\
=\ &\margANDembed{\kappa}{\Sigma'}{\Sigma}(A)
\end{align*}

For the second statement (2), we have to show that
$\Cons(\margsub{\kappa}{\Sigma'})
\subseteq \Cons(\margANDembed{\kappa}{\Sigma'}{\Sigma})$.
Let $(B'|A') \in \Cons(\margsub{\kappa}{\Sigma'})$; we will show that
$(B'|A') \in \Cons(\margANDembed{\kappa}{\Sigma'}{\Sigma})$.
Because $(B'|A') \in \Cons(\margsub{\kappa}{\Sigma'})$ we have
$(B'|A')  \in (\cL_{\Sigma'} | \cL_{\Sigma'})$ and
$\margsub{\kappa}{\Sigma'} \ESmodels (B'|A')$ and therefore
$\margsub{\kappa}{\Sigma'}(A'B') < \margsub{\kappa}{\Sigma'}(A'\ol{B'})$.
Since $\Sigma' \subseteq \Sigma$, the formulas $A', B'$ are also in $\cL_{\Sigma}$, and by (1),
we get
 $\margANDembed{\kappa}{\Sigma'}{\Sigma}(A'B') = \margsub{\kappa}{\Sigma'}(A'B') < \margsub{\kappa}{\Sigma'}(A'\ol{B'}) = \margANDembed{\kappa}{\Sigma'}{\Sigma}(A'\ol{B'})$, so $\margANDembed{\kappa}{\Sigma'}{\Sigma} \ESmodels (B'|A')$,}
 \FARBE{and hence $(B'|A') \in \Cons(\margANDembed{\kappa}{\Sigma'}{\Sigma})$.} 
\end{proof}

\FARBE{
	The following proposition shows several properties for marginalization, lifting, and the relationship among them.
	\begin{proposition}\label{prop:margeorderproposition}
		\FARBE{Let $ \Sigma, \Sigma', \Sigma'' $ be signatures and let $\kappa$ be an OCF over $\Sigma$.  }
		The following statements hold:
		\begin{enumerate}[(1)]
			\item \FARBE{If \( \Sigma' \subseteq \Sigma \) and \( \Sigma'' \subseteq \Sigma \), then \( \margANDmarg{\kappa}{\Sigma'}{\Sigma''} = \margsub{\kappa}{(\Sigma' \cap \Sigma'')} \).}
			\item \FARBE{If \( \Sigma \subseteq \Sigma' \) and \( \Sigma \subseteq \Sigma'' \), then \( \embedANDembed{\kappa}{\Sigma'}{\Sigma''} = \embedsup{\kappa}{(\Sigma' \cup \Sigma'')} \).}
			\item \FARBE{If \( \Sigma' \subseteq \Sigma \) and \( \Sigma'' \subseteq \Sigma \), then \( \margANDembedANDmarg{\kappa}{\Sigma'}{\Sigma}{\Sigma''} = \margANDembed{\kappa}{(\Sigma'\cap\Sigma'')}{\Sigma''}  \).}
		\end{enumerate}
	\end{proposition}
	\begin{proof}		 
	We show the satisfaction of the statements (1)--(3).
	\begin{description}
\item[{\normalfont\normalfont (1)}]
		At first, observe that \( \margANDmarg{\kappa}{\Sigma'}{\Sigma''} \) and \( \margsub{\kappa}{(\Sigma' \cap \Sigma'')} \) are functions of the same type, i.e., we have \( \margANDmarg{\kappa}{\Sigma'}{\Sigma''}: \Omega_{(\Sigma' \cap \Sigma'')} \to \naturals \) and \( \margsub{\kappa}{(\Sigma' \cap \Sigma'')}: \Omega_{(\Sigma' \cap \Sigma'')} \to \naturals \). Next, by considering Definition~\ref{def:marginalization}, we obtain the following:
		\begin{align*}
			\margsub{\kappa}{\Sigma'}(\omega')                       & = \min\{ \kappa(\omega) \mid \omega \in\Omega_\Sigma \text{ and } \omega' = \omega^{\Sigma'} \}       \\
			\margsub{\kappa}{(\Sigma' \cap \Sigma'')}(\omega')         & = \min\{ \kappa(\omega) \mid \omega \in\Omega_\Sigma \text{ and } \omega' = \omega^{(\Sigma' \cap \Sigma'')} \} \\
			\margANDmarg{\kappa}{\Sigma'}{\Sigma''}(\omega') & = \min\{ \margsub{\kappa}{\Sigma'}(\omega) \mid \omega' \in\Omega_{\Sigma'} \text{ and } \omega' = \omega^{\Sigma''} \}
		\end{align*}
		The following line of equivalences gives proof to \( \margANDmarg{\kappa}{\Sigma'}{\Sigma''}  = \margsub{\kappa}{(\Sigma' \cap \Sigma'')} \):        
		\begin{align*}
			\margANDmarg{\kappa}{\Sigma'}{\Sigma''}(\omega'') & = \min\{ \margsub{\kappa}{\Sigma'}(\omega') \mid \omega' \in \Omega_{\Sigma'} \text{ and } \omega'' = (\omega')^{\Sigma''} \} \\
			& = \min\{ \min\{ \kappa(\omega) \mid \omega \in \Omega_\Sigma \text{ and } \omega' = \omega^{\Sigma'} \} \mid \omega' \in \Omega_{\Sigma'} \text{ and } \omega'' = (\omega')^{\Sigma''} \} \\
			& = \min\{ \kappa(\omega) \mid \omega \in \Omega_\Sigma  \text{ and } \omega' \in \Omega_{\Sigma'} \text{ and } \omega' = \omega^{\Sigma'} \text{ and } \omega'' = (\omega')^{\Sigma''} \} \\
			& = \min\{ \kappa(\omega) \mid \omega \in \Omega_\Sigma \text{ and } \omega'' = \omega^{(\Sigma' \cap \Sigma'')} \}           \\
			& = \margsub{\kappa}{(\Sigma' \cap \Sigma'')}(\omega'')
		\end{align*}
        \item[{\normalfont (2)}]		
		Recall that the following holds:
		\begin{align*}
		    \embedsup{\kappa}{\Sigma'}(\omega)                 & = \kappa(\omega^{\Sigma})                      & \embedsup{\kappa}{\Sigma'}                & : \Omega_{\Sigma'} \to \naturals                 \\
		    \embedANDembed{\kappa}{\Sigma'}{\Sigma''}(\omega)          & = \embedsup{\kappa}{\Sigma'}(\omega^{\Sigma'}) & \embedANDembed{\kappa}{\Sigma'}{\Sigma''} & : \Omega_{(\Sigma' \cup \Sigma'')} \to \naturals \\
		    \embedsup{\kappa}{(\Sigma' \cup \Sigma'')}(\omega) & = \kappa(\omega^{\Sigma})                      & \embedsup{\kappa}{\Sigma'\cup\Sigma''}    & : \Omega_{(\Sigma' \cup \Sigma'')} \to \naturals
		\end{align*}
		Because \( \Sigma \subseteq \Sigma' \) holds, we obtain the following line of equivalences:
		\begin{align*}
			\embedANDembed{\kappa}{\Sigma'}{\Sigma''} (\omega) &  =  \embedsup{\kappa}{\Sigma'}(\omega^{\Sigma'}) =  \kappa((\omega^{\Sigma'})^{\Sigma}) = \kappa(\omega^{\Sigma}) = \embedsup{\kappa}{(\Sigma' \cup \Sigma'')}(\omega)
		\end{align*}
\item[{\normalfont (3)}]		
			First, recall that we have the following:
			\begin{align*}
				\margsub{\kappa}{\Sigma'}(\omega')             & = \min\{ \kappa(\omega) \mid \omega \in\Omega_\Sigma \text{ and } \omega' = \omega^{\Sigma'} \} & \margsub{\kappa}{\Sigma'} &: \Omega_{\Sigma'} \to \naturals   \\
				\margANDembed{\kappa}{\Sigma'}{\Sigma}(\omega) & = \margsub{\kappa}{\Sigma'}(\omega^{\Sigma'})  & \margANDembed{\kappa}{\Sigma'}{\Sigma} &: \Omega_{\Sigma} \to \naturals
			\end{align*}
			Hence, we obtain:
			\begin{align*}
				\margANDembed{\kappa}{\Sigma'}{\Sigma}(\omega) & = \margsub{\kappa}{\Sigma'}(\omega^{\Sigma'})                                                                                        \\
				                                               & = \min\{ \kappa(\omega_{\ast}) \mid \omega_{\ast} \in\Omega_\Sigma \text{ and } \omega_{\ast}^{\Sigma'} = \omega^{\Sigma'} \} 
			\end{align*}
			We also have:      
			\begin{align*}
				\margANDembedANDmarg{\kappa}{\Sigma'}{\Sigma}{\Sigma''}(\omega'') & = \min\{ \margANDembed{\kappa}{\Sigma'}{\Sigma}(\omega) \mid \omega \in\Omega_\Sigma \text{ and } \omega'' = \omega^{\Sigma''} \}       & \margANDembedANDmarg{\kappa}{\Sigma'}{\Sigma}{\Sigma''} : \Omega_{\Sigma''} \to \naturals
			\end{align*}
			By combining the equations above, we obtain:     
			\begin{align*}
				\margANDembedANDmarg{\kappa}{\Sigma'}{\Sigma}{\Sigma''}(\omega'')  & = \min\{ \margANDembed{\kappa}{\Sigma'}{\Sigma}(\omega) \mid \omega \in\Omega_\Sigma \text{ and } \omega'' = \omega^{\Sigma''} \}       \\
				& = \min\{ \min\{ \kappa(\omega_{\ast}) \mid \omega_{\ast} \in\Omega_\Sigma \text{ and } \omega_{\ast}^{\Sigma'} = \omega^{\Sigma'} \} \mid \omega \in \Omega_\Sigma \text{ and } \omega'' = \omega^{\Sigma''} \}\\
				& = \min\{  \kappa(\omega_{\ast}) \mid \omega_{\ast} \in \Omega_\Sigma \text{ and } \omega \in \Omega_\Sigma  \text{ and } \omega_{\ast}^{\Sigma'} = \omega^{\Sigma'} \text{ and } \omega'' = \omega^{\Sigma''} \}\\
				& = \min\{  \kappa(\omega) \mid \omega \in \Omega_\Sigma \text{ and }  (\omega'')^{\Sigma'\cap\Sigma''} = \omega^{\Sigma'\cap\Sigma''} \}
			\end{align*}        
			Now, observe that we have:
			\begin{align*}
				\margsub{\kappa}{(\Sigma' \cap \Sigma'')}(\omega')         & = \min\{ \kappa(\omega) \mid \omega \in\Omega_\Sigma \text{ and } \omega' = \omega^{(\Sigma' \cap \Sigma'')} \} & \margsub{\kappa}{(\Sigma' \cap \Sigma'')} &: \Omega_{\Sigma' \cap \Sigma''} \to \naturals \\
				\margANDembed{\kappa}{(\Sigma'\cap\Sigma'')}{\Sigma''}(\omega'') & = \margsub{\kappa}{\Sigma'\cap \Sigma''}((\omega'')^{(\Sigma' \cap \Sigma'')}) & \margANDembed{\kappa}{(\Sigma'\cap\Sigma'')}{\Sigma''} &: \Omega_{\Sigma''} \to \naturals
			\end{align*}
			Hence, we obtain:
			\begin{align*}
				\margANDembed{\kappa}{(\Sigma'\cap\Sigma'')}{\Sigma''}(\omega'') & = \margsub{\kappa}{\Sigma'\cap \Sigma''}((\omega'')^{(\Sigma' \cap \Sigma'')}) \\
				& = \min\{ \kappa(\omega) \mid \omega \in\Omega_\Sigma \text{ and } (\omega'')^{(\Sigma' \cap \Sigma'')} = \omega^{(\Sigma' \cap \Sigma'')} \}
			\end{align*}
	\end{description}
    \FARBE{This completes the proof of statements (1)--(3).\qedhere}
	\end{proof}
}
\FARBE{
Note that statement (1) and (2) of Proposition~\ref{prop:margeorderproposition} imply that both marginalization to subsignatures and lifting to supersignatures are commutative, i.e., the following statements hold:
\begin{itemize}
    \item \FARBE{If \( \Sigma' \subseteq \Sigma \) and \( \Sigma'' \subseteq \Sigma \), then} \, \( \margANDmarg{\kappa}{\Sigma'}{\Sigma''} = \margANDmarg{\kappa}{\Sigma''}{\Sigma'} \).
    \item \FARBE{If \( \Sigma \subseteq \Sigma' \) and \( \Sigma \subseteq \Sigma'' \), then} \, \(
   \FARBE{\embedANDembed{\kappa}{\Sigma'}{\Sigma''}}
= \embedANDembed{\kappa}{\Sigma''}{\Sigma'} \).
\end{itemize}
}

The lifting of a ranking function $\kappa$ to a larger signature corresponds to a model expansion used by Delgrande \cite{Delgrande17}. Liftings and
model expansions are defined along the inclusion of a subsignature to a larger
signature. Both can be extended to general morphisms between signatures in the
framework of institutions \cite{GoguenBurstall92JACM} that
have also been used for modelling many different conditional logics and transformations among them \cite{BeierleKernIsberner2009IJAR,BeierleKernIsberner2012AMAI}.

\subsection{Contractions}
\label{sec:instantiating_c_change}

In \cite{KernIsbernerBockSauerwaldBeierle2017}, a general framework for
contractions of (conditional) beliefs in epistemic states represented by OCFs
is presented.
The general case of a contraction of a ranking function $\kappa$ by
a proposition $A$ is given by a posteriori ranking function $ \kappa^\circ $
that does not accept $A$.
\begin{definition}[contraction by a single proposition]\label{def:contraction}
An operator $-$ taking an OCF $\kappa$ and a propositional formula $A$ and yielding an OCF $\contract{\kappa}{A}$ is a \emph{contraction} if $A \equiv \top$ implies
$\contract{\kappa}{A} = \kappa$, and if $A \not\equiv \top$ implies
$\contract{\kappa}{A} \not\models A$, i.e., \textbf{Contraction} as given
by Equation~\eqref{eqGeneralContration}
in
Section~\ref{sec:kindsOfForgetting} is satisfied.
\end{definition}
Within the class of contraction operators for OCFs and propositional formulas,
c-contrations constitute a particular subclass  governed by the principle of
conditional preservation
\cite{KernIsberner2004AMAI,KernIsbernerBockSauerwaldBeierle2017}.
The following definition is an adopted version of c-contractions
by a single proposition given in \cite{BeierleKernIsbernerSauerwaldBockRagni2019KIzeitschrift}.

\begin{definition}[c-contraction by a single proposition]\label{def:c_contraction}
Let $\kappa$ be an OCF and $A$ be a propositional formula. An OCF $ \contract{\kappa}{A} $ is a \emph{c-contraction of $ \kappa $ with $A$} if
in case $A \equiv \top$, $\contract{\kappa}{A} = \kappa$, 
and if in case $A \not\equiv \top$,   
there exist integers %
$\gamma^+,\gamma^- $
satisfying
		\begin{align}\label{eq:short_propositional_c_contraction_cond1}
			\gamma^- - \gamma^+ \leq \kappa(A)-\kappa(\ol{A}) 
		\end{align}
such that 
$\contract{\kappa}{A}$ has the form 
		\begin{align}\label{eq:short_propositional_c_contraction_cond2}
			\contract{\kappa}{A}(\omega) = - \gamma^- - \kappa(\ol{A}) + \kappa(\omega) + \begin{cases}
				\gamma^+ & \text{if } \omega\models A\\
				\gamma^- & \text{if } \omega\models \ol{A}
			\end{cases}
		\end{align}
\end{definition}

The integers $\gamma^+,\gamma^- $ are shifting factors the role of which is to make models of $A$ resp.\ $\ol{A}$ more or less plausible in a uniform way.
The values for $\gamma^+$ and $\gamma^- $ must be chosen such
that after the c-contraction $\kappa^{\circ}_A \not\models A$ holds; 
this is ensured by inequality (\ref{eq:short_propositional_c_contraction_cond1}). 
The term $- \gamma^- - \kappa(\ol{A}) $ normalizes 
$\contract{\kappa}{A}$, i.e., it guarantees that $\min_{\omega \in \Omega} \contract{\kappa}{A}(\omega) = 0$. 
Thus, we get the the following soundness result for c-contractions.

\begin{proposition}\label{prop_soundness_c_contraction}
C-contractions satisfy \textbf{Contraction}
given by Equation~\eqref{eqGeneralContration}
in Section~\ref{sec:kindsOfForgetting}.
\end{proposition}

In \cite{KernIsbernerBockSauerwaldBeierle2017}, 
type $ \alpha $ c-contractions and type $ \beta $ c-contractions were introduced for
contractions of conditionals.
The next definition carries over these types
to c-contractions of propositional formulas.

\begin{definition}
	Let $ \contract{\kappa}{A} $ be a c-contraction of $ \kappa $ with a proposition $ A $ of the form (\ref{eq:short_propositional_c_contraction_cond2}) with integers $\gamma^+,\gamma^- $ such that (\ref{eq:short_propositional_c_contraction_cond1}) holds.
	Then $ \contract{\kappa}{A} $ is called
	\begin{itemize}
		\item a \emph{type $ \alpha $ c-contraction} if $ \gamma^+ = 0 $, and
		\item a \emph{type $ \beta $ c-contraction} if $ \gamma^- = 0 $.
	\end{itemize}
\end{definition}

Type $ \alpha $ c-contractions impose no (additional) effect on models of $A$, while type $\beta$ c-contractions affect only the models of $A$ explicitly. From Definition \ref{def:c_contraction}, we derive immediately characterizations of type $\alpha$ resp.\ type $\beta$ c-contractions: 
\begin{proposition}
\label{prop_alpha_beta_characterizations}
	Let $ A $ a proposition such that $ A\not\equiv \top $.
	
	An OCF $ \contract{\kappa}{A} $ is a type $ \alpha $ c-contraction of $ \kappa $ with $ A $ if and only if there is an integer $ \gamma^- $ such that
				$\gamma^-  \leq \kappa(A)-\kappa(\ol{A}) $ and 
	\begin{equation}
		\label{eq:c_contract_a}
		\contract{\kappa}{A}(\omega) = - \kappa(\notA)+\kappa(\omega)+
		\begin{cases}
			- \gamma^- & \textit{if~}\omega \models A \\
			0 & \textit{if~}\omega \models \ol{A}
		\end{cases}
	\end{equation}
	An OCF $ \contract{\kappa}{A} $ is a type $ \beta $ c-contraction of $ \kappa $ with $ A $ if and only if there is an integer $ \gamma^+ $ such that $ - \gamma^+ \leq \kappa(A)-\kappa(\ol{A}) $ and 
	\begin{equation}
	\label{eq:c_contract_b}
\contract{\kappa}{A}(\omega)
	 = - \kappa(\notA)+\kappa(\omega)+
	\begin{cases}
		\gamma^+ & \textit{if~}\omega \models A \\
		0 & \textit{if~}\omega \models \ol{A}
	\end{cases}
\end{equation}
\end{proposition}

From this proposition, we can see immediately that the classes of type  $ \alpha $ and type $ \beta $  c-contractions coincide in the propositional case.

\begin{proposition}
\label{prop_alpha_eq_beta}
An OCF $ \contract{\kappa}{A} $ is a type  $ \alpha $ c-contraction of $ \kappa $ with $ A $ if and only if $ \contract{\kappa}{A} $ is a type $ \beta $ c-contraction of $ \kappa $ with $ A $.
\end{proposition}
\begin{proof}
Observe that substitution of $ \kappaminus $ with $ -\kappaplus $ in Equation \eqref{eq:c_contract_a} yields Equation \eqref{eq:c_contract_b}, and vice versa, and also the appertaining inequalities correspond to each other.
\end{proof}

Now, from Definition \ref{def:c_contraction} and Proposition \ref{prop_alpha_eq_beta}, we can easily derive 
that every c-contraction with a proposition is a type $ \alpha $ c-contraction (respectively a type $ \beta $ c-contraction).
\begin{proposition}
\label{prop_alpha_beta_all}
	Every propositional c-contraction with a proposition $ A $ is a type $ \alpha $ (resp.\ type $\beta$) c-contraction.
\end{proposition}
\begin{proof}
	Let $ \contract{\kappa}{A} $ be the result of a c-contraction of $ \kappa $ by a proposition $ A $. 
	Then by %
	Definition \ref{def:c_contraction}, 
	there are $ \gamma^+ $ and $ \gamma^- $ with $ \gamma^- -\gamma^+ \leq \kappa(A) - \kappa(\ol{A}) $, such that Equation \eqref{eq:short_propositional_c_contraction_cond2} is satisfied, i.e.,
	\begin{equation*}
	\contract{\kappa}{A}(\omega)=-\gamma^- -\kappa(\overline{A}) + \kappa(\omega)+\begin{cases}
	\gamma^+ & \text{if } \omega\models A \\
	\gamma^- & \text{if } \omega\models \overline{A}\\
	\end{cases}
	\end{equation*}
	A simple algebraic transformation of this equation
	yields:
	\begin{equation}
	\label{eq_all_alpha}
	\contract{\kappa}{A}(\omega)= -\kappa(\overline{A}) + \kappa(\omega)+
	\begin{cases}
	\gamma^+ -\gamma^- & \text{if } \omega\models A \\
	0 & \text{if } \omega\models \overline{A}\\
	\end{cases}
	\end{equation}
	Clearly, this is a type $ \alpha $ c-contraction.
\end{proof}

So, in the case of propositional c-contractions, type $\alpha$ and type $\beta$ c-contractions both coincide with the class of all c-contractions. This is due to the fact that here, we only have two cases (a world can satisfy a proposition or not), and the normalization factor balances differences. But note that in the conditional case that is dealt with in \cite{KernIsbernerBockSauerwaldBeierle2017}, type $\alpha$ and type $\beta$ c-contractions form distinct and proper subclasses of all c-contractions.

A consequence of Proposition \ref{prop_alpha_beta_all} is that we need not distinguish between $\gamma^+$ and $\gamma^-$ anymore and hence obtain from (\ref{eq_all_alpha}) an even simpler characterization of c-contractions with the following proposition.

\begin{proposition}
 \label{prop_c-contr_gamma}
 Let $\kappa$ be an OCF and $A \not\equiv \top$ be a propositional formula. An OCF $ \contract{\kappa}{A} $ is a c-contraction of $ \kappa $ with $A$ iff
there exists an integer $\gamma$
satisfying
		\begin{align}\label{eq:short_propositional_c_contraction_cond1_gamma}
			\gamma \geq \kappa(\ol{A}) - \kappa(A)
		\end{align}
such that 
$\contract{\kappa}{A}$ has the form 
		\begin{align}\label{eq:short_propositional_c_contraction_cond2_gamma}
			\contract{\kappa}{A}(\omega) =  - \kappa(\ol{A}) + \kappa(\omega) + \begin{cases}
				\gamma & \text{if } \omega\models A\\
				0 & \text{if } \omega\models \ol{A}
			\end{cases}
		\end{align}
\end{proposition}

c-Contractions form a large family of diverse contraction operations which deliberately go beyond AGM \cite{KernIsbernerBockSauerwaldBeierle2017}. Therefore, we choose special instances of c-contractions to elaborate on the potential of c-contractions for forgetting operations in more depth.
In the following
subsection,
we
\FARBE{employ selection strategies \cite{BeierleKernIsberner2021FLAIRS} for conveniently capturing}
the choosing of c-contractions for different purposes and under different points of view.

\subsection{Selection Strategies for C-Contractions}
\label{sec:selection_strategies}
\newcommand{\contractOpSelect}{\ensuremath{-_{\selectionStrategyContractProp}}}
\newcommand{\ignoreOpSelect}{\ensuremath{\ominus_{\selectionStrategyContractProp}}}

As shown above,
c-contractions by a single proposition $A$ are parametrized by
$\gamma$
satisfying
\eqref{eq:short_propositional_c_contraction_cond1_gamma}.
The next defintion provides a representation of the solution space
for the value of $\gamma$
in a c-contration
by a constraint satisfaction problem (CSP).

\begin{definition}[\cspContractProp]
\label{def_csp_c_contration}
  Let $\kappa$ be an OCF
  and $A \not\equiv \top$ be a proposition.
The \emph{constraint satisfaction problem for c-contractions of $\kappa$ by $A$},
denoted by \cspContractProp, is given by the constraint
	\begin{align}\label{eq_impacts_contration}
    	\gamma \geq \kappa(\overline{A}) - \kappa(A)
	\end{align}
where $\gamma$ is a constraint variable taking values in $\mathbb{Z}$.
\end{definition}

For a constraint satisfaction problem \(\mathit{CSP}\), the set of solutions is denoted by \(\solutionsOf{\mathit{CSP}}\).
Thus, with \solutionsContractProp\ we denote the set of all solutions of \cspContractProp.
For any
\(
   \gamma \in \solutionsContractProp
\),
the function $\kappaCirc = \contract{\kappa}{A}$ as defined by
Equation~\eqref{eq:short_propositional_c_contraction_cond2_gamma}
 is denoted by \(\induzierteContractProp{\gamma}\).
From Proposition~\ref{prop_c-contr_gamma}, we immediately get that
\cspContractProp\ is sound an complete.

\begin{proposition}[Soundness and completeness of \cspContractProp]
  \label{prop_csp_ContractProp_sound_complete}
  Let $\kappa$ be an OCF and $A \not\equiv \top$ be a proposition.
\begin{itemize}
\item
If 
\(\gamma \in  \solutionsContractProp\)
then
\(\induzierteContractProp{\gamma}\)
is a c-contraction of $\kappa$ by $A$.

\item
  If $\kappaCirc$ is a c-contraction of $\kappa$ by $A$ then there is
\(\gamma \in  \solutionsContractProp\)
such that \(\kappaCirc = \induzierteContractProp{\gamma}\).
\end{itemize}
\end{proposition}

Given the
general schema
for c-contraction
operators for propositional formulas, many c-contractions
are possible.
Therefore, it might 
be useful to impose further constraints on the
constraint variable $\gamma$. 
For instance, one might want 
to take minimal values satisfying (\ref{eq_impacts_contration}).
For being able to systematically impose further restrictions on the impact
$\gamma$ determining a c-contraction
of a proposition, we will employ the notion of 
a selection strategy \cite{BeierleKernIsberner2021FLAIRS} for choosing the impact value.

  \begin{definition}[selection strategy \selectionStrategyContractProp, \FARBE{(strategic)} c-contraction \contractOpSelect]
  \label{def_selection_strategy_contract_ignore}
  A \emph{selection strategy (for c-contractions of OCFs by single propositions)}
  is a function
 \[
  \selectionStrategyContractProp:  (\kappa, A) \mapsto \gamma %
  \]
  assigning to each pair of an OCF $\kappa$ and a propositional formula $A$
  with $A \not\equiv \top$
  an impact %
  $\gamma \in \solutionsContractProp$.
If $\selectionStrategyContractProp(\kappa, A) = \gamma$, the
\emph{c-contraction}
of $\kappa$ by $A$ determined by \selectionStrategyContractProp\ is
 \(\induzierteContractProp{\gamma}\),
 denoted by $\kappa \contractOpSelect A$
 and implicitly extended
 by $\kappa \contractOpSelect A = \kappa$ for  $A \equiv \top$.
 Furthermore, \contractOpSelect\ is called a \emph{(strategic) c-contraction (operator)}.
\end{definition}

Note that a strategic c-contraction operator \contractOpSelect\
selects a single c-contraction for each $\kappa$ and each
propositional formula $A$.
 Thus, 
 focusing on strategic c-contractions
 does
 not exclude 
 any possible options
 because
 for every c-contraction $-$,
 there
 is a selection strategy \selectionStrategyContractProp\ such that
 ${-} = {\contractOpSelect}$.
 On the other hand, employing the concept of selection strategies,
 we can conveniently specify conditions for \selectionStrategyContractProp\ that ensure properties we expect from  \contractOpSelect.
 First, we present a basic postulate ensuring independence of the syntactic representation of the formula to be contracted.

\begin{description}
\item[\postSyntaxIndependent]
  \selectionStrategyContractProp\
  is \emph{syntax independent} if
  $A \equiv A'$ implies
  $\selectionStrategyContractProp(\kappa, A) = \selectionStrategyContractProp(\kappa, A')$.
\end{description}

\begin{proposition}
If \selectionStrategyContractProp\ is a selection strategy satisfying \postSyntaxIndependent,
$\kappa$ is an OCF and $A, A'$ are
propositional formulas, 
then  $A \equiv A'$ implies
 $\kappa \contractOpSelect A = \kappa \contractOpSelect A'$.
\end{proposition}

In this article, we will usually presuppose that selection strategies satisfy \postSyntaxIndependent. Beyond that, other
postulates partition c-contractions into distinct relevant subclasses.
Postulates for such conditions will be elaborated in the next subsection.

\subsection{Subclasses of Contractions}
\label{sec:instantiating_subclasses}

In the general case, if a ranking function 
$\kappa - A$
is a contraction of $\kappa$  by a propositional formula $A$, we have 
$\kappa - A \not\models A$.
 Now, two possibilities arise: it might be that also 
 $ \kappa - A \not\models \neg A$, 
  in which case 
  $\kappa - A$
  is completely \emph{ignorant} about $A$, or it might be that 
 $ \kappa - A \models \neg A$, 
  in which case we have a \emph{revocation} of $A$, cf.\ Section~\ref{sec:kindsOfForgetting}.
 Embedding these general concepts of ignorance and revocation into the
 framework of c-contractions can be done conveniently by means of
 corresponding postulates for selection strategies \selectionStrategyContractProp\ for c-contractions.

\begin{description}
\item[\postIgnore]
  \selectionStrategyContractProp\
  is \emph{ignorant} if
  $\selectionStrategyContractProp(\kappa, A) = \gamma$
  implies $\gamma = \kappa(\ol{A}) - \kappa(A)$.

\item[\postRevoke]
  \selectionStrategyContractProp\
  is \emph{revoking} if
  $\selectionStrategyContractProp(\kappa, A) = \gamma$
  implies $\gamma > \kappa(\ol{A}) - \kappa(A)$.

\end{description}

These postulates immediately lead to two subclasses of c-contractions.

\begin{definition}[c-ignoration and c-revocation by a single proposition]
\label{def_cignoration_crevocation}
Let \selectionStrategyContractProp\ be a selection \FARBE{strategy for} c-contractionbs by a single proposition.
Then \contractOpSelect\ is a \emph{c-ignoration} if \selectionStrategyContractProp\ satisfies \postIgnore, and
\contractOpSelect\ is a \emph{c-revocation} if \selectionStrategyContractProp\ satisfies 
\postSyntaxIndependent\ and 
\postRevoke.
\end{definition}

Analogously to Proposition~\ref{prop_soundness_c_contraction},  we get the the following soundness results for c-ignorations and c-revocations.

\begin{proposition}\label{prop_soundness_c_ignoration_c_revocation}
C-ignorations satisfy \textbf{Ignoration} and
c-revocations satisfy \textbf{Revocation}
(cf.\ Section~\ref{sec:kindsOfForgetting}).
\end{proposition}

In addition to c-ignorations and c-revocations,
the next two \FARBE{postulates induce} 
natural subclasses of c-contractions.

\begin{description}
\item[\postMinimal]
  \selectionStrategyContractProp\
  is \emph{minimal} if
  $\selectionStrategyContractProp(\kappa, A) = \gamma$
  implies %
  $\gamma = \kappa(\ol{A})$.

\item[\postNonMinimal]
  \selectionStrategyContractProp\
  is \emph{non-minimal} if
  $\selectionStrategyContractProp(\kappa, A) = \gamma$
  implies 
  $\gamma \not= \kappa(\ol{A})$. 

\end{description}
\FARBE{Both postulates give rise to spefic classes of c-contractions, which we define in the following.}

\begin{definition}[minimal and non-minimal c-contraction]
\label{def_minimal_nonminimal_ccontraction}
Let \selectionStrategyContractProp\ be a selection \FARBE{strategy for} c-contractions by a single proposition.
Then \contractOpSelect\ is a \emph{minimal c-contraction} if \selectionStrategyContractProp\ satisfies \postMinimal, and
\contractOpSelect\ is a \emph{non-minimal c-contraction} if \selectionStrategyContractProp\ satisfies 
\postSyntaxIndependent\ and 
\postNonMinimal.
\end{definition}

Intuitively, minimal c-contractions, that are also defined in 
\cite{KernIsbernerBockSauerwaldBeierle2019}, but without employing
selection strategies, 
are those c-contractions where minimal changes on the ranking functions are made to achieve the success condition 
$ \kappa - A \not\models A $.
More specifically, in minimal c-contractions, the ranks of $A$-models are not shifted at all, and the ranks of $\ol{A}$-models are shifted in a minimal way.

\FARBE{

\FARBE{
	We define selection strategies \( \strategyMinimal \) and \( \strategyIgnore \)  that implement {\postMinimal} and {\postIgnore} by
	\begin{align*}
		\strategyMinimal(\kappa,A)   & = \kappa(\ol{A})         \ \ \text{ and}      \\  
		\strategyIgnore(\kappa,A)   & = \kappa(\ol{A}) - \kappa(A) \ .                                           
	\end{align*}
	Each of {\strategyMinimal} and {\strategyIgnore} \FARBE{gives} rise to exactly one (strategic) c-contraction (operator):
	\begin{align*}
		{\selectOpMinimal} & = {-_{\strategyMinimal}} \ \ \ \ \text{ and } & {\selectOpIgnore} & = {-_{\strategyIgnore}}
	\end{align*}
Clearly, these approaches are unqiue realizations of {\postMinimal} and {\postIgnore}, which is stated in the following proposition.
\begin{proposition}\label{prop:unqiue_min_ignore}
The following statements hold:
\begin{itemize}
	\item The one and only selection strategy that satisfies {\postMinimal}, respectively {\postIgnore}, is \( \strategyMinimal \), respectively \( \strategyIgnore \).
	\item {\selectOpMinimal} is the unique  (strategic) c-contraction (operator) that is a minimal c-contraction.
	\item {\selectOpIgnore} is the unique (strategic) c-contraction (operator) that is a c-ignoration.
\end{itemize}

\end{proposition}
}

\FARBE{Note that we need not presuppose \postSyntaxIndependent\ for c-ignorations and minimal c-contractions because 
both are insensitive to the syntactic structure of the new information $A$ by definition, as the following proposition shows:}
\begin{proposition}
    \label{prop_ignore_min_si}
    The \FARBE{selection strategies} \FARBE{\strategyMinimal\ and   \strategyIgnore\ satisfy} \postSyntaxIndependent. 
\end{proposition}
\begin{proof}
Expanding the definition of \( \strategyMinimal \) yields the following observation for all \( A,B\in\mathcal{L} \) with \( A\equiv B \):
\begin{align*}
    \strategyMinimal(\kappa,A) & = \kappa(\ol{A}) = \min_{ \omega\in\modelsOf{\ol{A}}  }\{ \kappa(\omega) \} = \min_{ \omega\in\modelsOf{\ol{B}}  }\{ \kappa(\omega) \} = \kappa(\ol{B}) = 
    \strategyMinimal(\kappa,B)
\end{align*}
Analogously, for \( \strategyIgnore \) we obtain
\begin{align*}
    \strategyIgnore(\kappa,A) & = \kappa(\ol{A})  - \kappa(A)  = \min_{ \omega\in\modelsOf{\ol{A}}}\{ \kappa(\omega) \} - \min_{ \omega\in\modelsOf{A}}\{ \kappa(\omega) \} \\
            & = \min_{ \omega\in\modelsOf{\ol{B}}  }\{ \kappa(\omega) \} - \min_{ \omega\in\modelsOf{B}}\{ \kappa(\omega) \} = \kappa(\ol{B}) - \kappa(B) = 
    \strategyIgnore(\kappa,B)
\end{align*} 
for all \( A,B\in\mathcal{L} \) with \( A\equiv B \).
\end{proof}
}

\FARBE{Note that
	\FARBE{both c-revocations and c-ignorations on the one hand and also minimal  and
		non-minimal c-contractions on the other hand}
	are exclusive but not exhaustive classes of c-contractions because strategies can be mixed, e.g., yielding a c-revocation for some $\kappa, A$, and a c-ignoration for others. 
	\FARBE{Moreover, in certain cases the strategies coincide. This is especially true for revocation and non-minimal c-contractions, and for c-ignorations and minimal c-contractions, as the following proposition \FARBE{shows.}}
	\begin{proposition}\label{prop:sep_c_contraction_coincide}
		The following statements hold:
			\begin{enumerate}[(a)]
				\item For each OCF \( \kappa \) and for each \( A\in\cL  \) with \( \kappa(A)=0 \), it holds that \( \kappa \selectOpIgnore A = \kappa \selectOpMinimal A \), i.e., c-ignorations and minimal c-contractions coincide if $\kappa(A) = 0$.
				\item If $\kappa(A) > 0$, then every minimal c-contraction is also a c-revocation.
				\item If $\kappa(A) = 0$, then every c-revocation is also a non-minimal c-contraction, and vice versa.
			\end{enumerate}
	\end{proposition}
	\begin{proof}
		The different statements are given by the following observations:		
		\begin{itemize}[(a)]
				\item[]\textit{Statement (a).} Considering Equation \eqref{eq:short_propositional_c_contraction_cond1_gamma} reveals immediately that \postIgnore\ and \postMinimal\ \FARBE{coincide} if and only if  \( \kappa(A)=0 \) holds.
				
				\item[]\textit{Statement (b).} Let $\kappa(A) > 0$ and let the change from \( \kappa \) to \( \kappa \selectOpMinimal A \) be the minial c-contraction.
				From $\kappa(A) > 0$, we obtain that \(  \kappa(\ol{A})=0 \) holds. Consequently, according to {\postMinimal}, we obtain that \( \kappa =  \kappa \selectOpMinimal A\) holds and thus, also that \( \kappa \selectOpMinimal A(\ol{A})=0 \) and \( \kappa \selectOpMinimal A(A) > 0 \) holds. From this last observation we obtain that \( \kappa \selectOpMinimal A \models \neg A \) holds. \FARBE{Consequently, the} change from \( \kappa \) to \( \kappa \selectOpMinimal A \) is a c-revocation.
				
				\item[]\textit{Statement (c).} For \( \kappa(A)=0 \) we obtain that \( \gamma > \kappa(\ol{A}) \) holds if and only if \( \gamma \neq \kappa(\ol{A}) \) holds. Consequently, the postulates \postRevoke\ and \postNonMinimal\ coincide for \( \gamma > \kappa(\ol{A}) \).\qedhere
		\end{itemize}
\end{proof}}

By considering the postulates \postIgnore, \postRevoke, \postMinimal, {\postNonMinimal} its seems natural
to think that the classes of selection strategies that fulfill one of these postulates are pairwise disjoint.
The following proposition attests that this
impression
    is true indeed.
\begin{proposition}
\label{prop:sep_c_contraction_types}
For each \(\mathbf{X},\mathbf{Y}\in \{ \postIgnore,\postRevoke,\postMinimal,\postNonMinimal  \}\) with \( \mathbf{X} \neq \mathbf{Y} \) there exists a selection strategy \( \selectionStrategyContractProp \) that satisfies {\postSyntaxIndependent} and \( \mathbf{X} \), and violates \( \mathbf{Y} \).
\end{proposition}
\begin{proof}
        We use the selection strategies \( \strategyIgnore \) and \( \strategyMinimal \) and the following selection strategies:
    \begin{align*}
        \strategyRevocation(\kappa,A)   & = 1+ \kappa(\ol{A}) - \kappa(A)                                                            \\
        \strategyNonMinimal(\kappa,A) & = \min\{ \gamma \in \mathbb{Z} \mid \gamma \geq \kappa(\ol{A}) - \kappa(A), \gamma \neq \kappa(\ol{A}) \}
    \end{align*}
One can verify that \( \strategyRevocation \) satisfies {\postRevoke} and {\postSyntaxIndependent}, and that \(\strategyNonMinimal  \) satisfies {\postNonMinimal} and {\postSyntaxIndependent}.
Moreover,  the selection strategy \( \strategyRevocation \) is chosen minimally, which is possible due to Equation \eqref{eq:short_propositional_c_contraction_cond1_gamma} in Proposition \ref{prop_c-contr_gamma}. 
Hence, there is no other selection strategy \( \selectionStrategyContractProp \) that satisfies \postRevoke\ such that there are an OCF \( \kappa \) and a formula \( A\in\cL \) such that \( \selectionStrategyContractProp(\kappa,A) < \strategyRevocation(\kappa,A) \).
Likewise, the selection strategy \( \strategyNonMinimal \) is chosen minimally, i.e. there is no other selection strategy \( \selectionStrategyContractProp \) that satisfies \postNonMinimal\ such that there are an OCF \( \kappa \) and a formula \( A\in\cL \) such that \( \selectionStrategyContractProp(\kappa,A) < \strategyNonMinimal(\kappa,A) \).
Recall that according to Proposition~\ref{prop:unqiue_min_ignore} the only selection strategy that satisfies {\postIgnore} is \( \strategyIgnore \) and the only selection strategy that satisfies {\postMinimal} is  \( \strategyMinimal \).

\begin{figure}[t]
    \begin{center}
        \newcolumntype{P}[1]{>{\centering\arraybackslash}b{#1}}
        \begin{tabular}{r||P{1.5cm}|P{1.5cm}|P{1.5cm}|P{1.5cm}}
            \diagbox{\( X \)}{\( \mathbf{Y} \)} & \postIgnore    & \postRevoke    & \postMinimal   & \postNonMinimal \\ \hline\hline
            \strategyIgnore &                & \( \kappa_1 \) & \( \kappa_1 \) & \( \kappa_2 \)  \\[.1em] \hline
            \strategyRevocation & \( \kappa_1 \) &                & \( \kappa_2 \) & \( \kappa_1 \)  \\[.1em] \hline
            \strategyMinimal & \( \kappa_1 \) & \( \kappa_2 \) &                & \( \kappa_1 \)  \\[.1em] \hline
            \strategyNonMinimal & \( \kappa_2 \) & \( \kappa_1 \) & \( \kappa_2 \) &
        \end{tabular}
    \end{center}
    \caption{OCFs that witness that the strategies \( X\in \{ \) {\strategyIgnore}, {\strategyRevocation}, {\strategyMinimal}, {\strategyNonMinimal} \( \} \) from the proof of Proposition~\ref{prop:sep_c_contraction_types} violate postulates of \( \mathbf{Y}\in\{\postIgnore,\postRevoke,\postMinimal,\postNonMinimal\} \).}
    \label{fig:sep_c_contraction_types_emp}
\end{figure}
In the following, we give the proof for \( \Sigma=\{a\} \); the result carries over to arbitrary signatures canonically.
We will make use of the following OCFs \( \kappa_1 \) and \( \kappa_2 \) given by:
\begin{align*}
    \kappa_1(a) & = 1 & \kappa_1(\ol{a}) & = 0  &
    \kappa_2(a) & = 0 & \kappa_1(\ol{a}) & = 1 
    \end{align*}
Next, we list results of the above-mentioned strategies for \( \kappa_1 \) and \( \kappa_2 \), and the formula  \( a \):
\begin{align*}
    \strategyIgnore(\kappa_1,a) & = -1 & \strategyRevocation(\kappa_1,a) & = 0 & \strategyMinimal(\kappa_1,a) & = 0 & \strategyNonMinimal(\kappa_1,a) & = -1 \\
    \strategyIgnore(\kappa_2,a) & = 1 & \strategyRevocation(\kappa_2,a) & = 2 & \strategyMinimal(\kappa_2,a) & = 1 & \strategyNonMinimal(\kappa_2,a) & = 2 
\end{align*}
Let \( X=\textit{ign} \) and \( \mathbf{Y}=\postRevoke \). As given above, we have that \( \selectionStrategyContractProp_X(\kappa_1,a)<\strategyRevocation(\kappa_1,a) \) holds. By using this and because \( \strategyRevocation \) is the minimal selection strategy that satisfies \( \mathbf{Y} \), we obtain that \( \selectionStrategyContractProp_X \) violates \( \mathbf{Y} \).
Let \( X=\textit{rev} \) and \( \mathbf{Y}=\postIgnore \); and recall that \( \strategyIgnore \) is the only selection strategy that satisfies {\postIgnore}. Consequently, as \( \strategyIgnore(\kappa_1,a)\neq\selectionStrategyContractProp_X(\kappa_1,a) \) holds, we obtain that \( \selectionStrategyContractProp_X \) violates {\postIgnore}.
By using an analogous argumentation, one can show for each of \(X\in \{ \textit{ign},\textit{rev},\textit{min},\textit{non-min}\}\) and \( \mathbf{Y}\in \{ \postIgnore,\allowbreak\postRevoke,\allowbreak\postMinimal,\allowbreak\postNonMinimal  \} \)  that \( \sigma_{\!X}\) violates \( \mathbf{Y} \) with the following exceptions: \( X=\textit{ign} \) and \( \mathbf{Y}=\postIgnore \); \( X=\textit{rev} \) and \( \mathbf{Y}=\postRevoke \); \( X=\textit{min} \) and \( \mathbf{Y}=\postMinimal \); \( X=\textit{non-min} \) and \( \mathbf{Y}=\postNonMinimal \); \( X=\textit{rev} \) and \( \mathbf{Y}=\postNonMinimal \). Figure~\ref{fig:sep_c_contraction_types_emp} summarzies which OCF acts as witnesses for these claims.

We show that \( \strategyRevocation \) violates \( \postNonMinimal \). Note that according to {\postNonMinimal} every selection strategy \( \selectionStrategyContractProp \) that satisfies {\postNonMinimal} is a strategy that satisfies \( \selectionStrategyContractProp(\kappa, A) \neq \kappa(\ol{A})\). Consequently, \( \strategyRevocation \) violates \postNonMinimal, as \( \strategyRevocation(\kappa_1,a)=\kappa_1(\ol{a}) \) holds.
\end{proof}

\begin{figure}[t]
    \centering
    \newcolumntype{P}[1]{>{\centering\arraybackslash}b{#1}}
    \begin{tabular}{ccc|P{2cm}P{2cm}P{2cm}P{2cm}}
        \toprule
        & & & c-ignoration & c-revocation & minimal c-contraction & non-minimal c-contraction \\\midrule\midrule
        & \multicolumn{2}{c|}{\( \kappa_1 \)} & \(  \kappa_1 \selectOpIgnore a \) & \(  \kappa_1 \selectOpRevocation a \) & \(  \kappa_1 \selectOpMinimal a \) &\(  \kappa_1 \selectOpNonMinimal a \) \\[0.25em]\cmidrule(lr{1em}){2-7} %
        & 1 & \( a \)      & & \( a \)& \( a \)& \\ %
        \multirow{2}{*}[1em]{\begin{sideways}{\normalsize\ \ \ \ rank\ }\end{sideways}}& 0 & \( \ol{a} \) & \( \ol{a},a \)  & \( \ol{a} \)& \( \ol{a} \) & \( \ol{a},a \) \\
        \midrule\midrule
        & \multicolumn{2}{c|}{\( \kappa_2 \)} & \(  \kappa_2 \selectOpIgnore a \) & \(  \kappa_2 \selectOpRevocation a \) & \(  \kappa_2 \selectOpMinimal a \) &\(  \kappa_2 \selectOpNonMinimal a \)  \\[0.25em]\cmidrule(lr{1em}){2-7} %
        & 1 &  \( \ol{a} \) & & \( a \)& & \( a \) \\ %
        \multirow{2}{*}[1em]{\begin{sideways}{\normalsize\ \ \ \ rank\ }\end{sideways}}& 0 & \( a \) & \( \ol{a},a \)  & \( \ol{a} \)& \( \ol{a},a \) & \( \ol{a} \) \\\midrule
        \bottomrule
    \end{tabular}
    \caption{The c-ignoration operator \( \selectOpIgnore \), the c-revocation opereator \( \selectOpRevocation = {-_{\strategyRevocation}} \), the minimal c-contraction operator \( \selectOpMinimal \) and the non-minimal c-contraction operator \( \selectOpNonMinimal = {-_{\strategyNonMinimal}} \) induced by the strategies used in the proof of Proposition~\ref{prop:sep_c_contraction_types} and their behaviour on \( \kappa_1 \), \( \kappa_2 \) and the formula \( a \) from Proposition~\ref{prop:sep_c_contraction_types}. The table is a witness for Collorary~\ref{col:sep_c_contraction_types_operators}.}\label{fig:sep_c_contraction_types}
\end{figure}
    The following corollary presents the results of Proposition~\ref{prop:sep_c_contraction_types} from the perspective of  strategic c-contraction operators, i.e., it describes that the classes of c-ignoration operators, c-revocation operators, minimal c-contraction operators and non-minimal c-contraction operators are pairwise different. 
\begin{corollary}
    \label{col:sep_c_contraction_types_operators}
    For each \( X,Y \in \{ \)\text{c-ignoration}, \text{c-revocation}, \text{minimal c-contraction}, \text{non-minimal c-revocation}\( \} \) with \( X\neq Y \) there exists a syntax-independet strategic c-contraction operator \( {-} \) such that \( {-} \) is a \( X \) but not a \( Y \).
\end{corollary}
Clearly, the selection strategies \( \strategyIgnore, \strategyRevocation, \strategyMinimal \) and \( \strategyNonMinimal \) give rise to \FARBE{corresponding} strategic c-contraction operators, which act as witness for Corollary~\ref{col:sep_c_contraction_types_operators}. Figure~\ref{fig:sep_c_contraction_types} presents the behaviour of these strategic c-contraction operators for the OCFs \( \kappa_1 \) and \( \kappa_2 \) given in the proof of Proposition~\ref{prop:sep_c_contraction_types}.

\FARBE{In this section, we showed how all kinds of forgetting presented in the general setting used in Section~\ref{sec:kindsOfForgetting} can be realized in the context of ranking functions.
In the following, we will evaluate all forgetting operators obtained here with respect to the general postulates for  forgetting operators developed in Section~\ref{sec:general_props}.}

\section{\FARBE{Specifics of OCFs for Postulates}}
\label{sec_eval_ocf_specifics}

\FARBE{Before evaluating the instantiations of the forgetting operations in the OCF framework which we  presented in Section \ref{sec:instantiating} with respect to the postulates from Section \ref{sec:general_props} in the next section, we elaborate on particularly concise characterization of some postulates for OCFs.
Sections \ref{sec_specifics_agm} and \ref{sec_specifics_asp} focus on postulates from the AGM resp.\ ASP domain, and
Propositions~\ref{prop:agm-postulates-ocf-characterization} and~\ref{prop:asp-postulates-ocf-characterization} summarize these equivalent reformulations which are possible due to the specific features of OCFs. Moreover, in Section \ref{sec_specifics_lin_eq}, we consider linear equivalence, a specific form of equivalence for OCFs (\FARBE{cf.}\ Definition~\ref{def_ocf_xequiv}), and propose a postulate  \propxE that is tailor-made to handle the arithmetics of OCFs more accurately.

For this section, please recall that a forgetting operation on OCFs is a function \( {}^\circ \) that maps an OCF \( \kappa \) and a formula \( A \) to an OCF $\forgetBy{\kappa}{A}$.
}
\FARBE{
\subsection{\FARBE{AGM-inspired Forgetting Postulates for OCFs}}
\label{sec_specifics_agm}
We start with considering AGM-inspired postulates. \FARBE{In order to avoid too technical reformulations, we limit the following proposition to forgetting operations which do not change the \FARBE{domain}\FARBE{, which we call domain-preserving forgetting operators:
    \begin{center}
        (\emph{domain-preservation}) \begin{minipage}[t]{0.7\textwidth}
          For all $\kappa$ and \( A \), the result of forgetting $\forgetBy{\kappa}{A}$ is defined over the same set of possible worlds $\Omega_{\Sigma}$ as $\kappa$.
        \end{minipage}
    \end{center}%
}
    This is to avoid ambiguity in the interpretation of the AGM postulates, since these postulates were designed without signature changes (e.g.\ forgetting by marginalization) in mind.}
}

\begin{proposition}[Characterizations of AGM-Inspired Postulates for OCF-Forgetting]
	\label{prop:agm-postulates-ocf-characterization}
    Let $\kappa$ be an OCF over $\Sigma$, and let $\omega \in \Omega_{\Sigma}$ and $A,C \in \cL_{\Sigma}$. The following statements hold for every \FARBE{domain-preserving} forgetting operation  \( {}^\circ \) on OCFs:
	\begin{enumerate}
		\renewcommand*\labelenumi{(\theenumi)}
		\item \agm{1} is satisfied by \( {}^\circ \) iff the following is satisfied:
		\begin{description}
			\item[] If $\kappa(\omega) = 0 \text{, then }  \forgetBy{\kappa}{A}(\omega) = 0 \ .$
		\end{description} 
		      
		\item \agm{2} is satisfied by \( {}^\circ \) iff the following is satisfied:
		\begin{description}
		\item[] If $\kappa(\ol{A})  = 0$ holds, then $ \forgetBy{\kappa}{A}(\omega) = 0$ implies  $\kappa(\omega) = 0$.  
		\end{description}      
		       
		\item \agm{3} is satisfied by \( {}^\circ \) iff the following is satisfied:
		\begin{description}
		\item[] If $A$ is not a tautology, then $ \forgetBy{\kappa}{A}(\ol{A}) = 0$. 
		\end{description}   
		
		\item \agm{4} is satisfied by \( {}^\circ \) iff the following is satisfied:
		\begin{description}
		\item[] If $\omega \models A$, then $\kappa(\omega) > 0$ implies $ \forgetBy{\kappa}{A}(\omega) > 0$. 
		\end{description}
		
		\item \agm{5} is satisfied by \( {}^\circ \) iff the following is satisfied:
		\begin{description}
		\item[] If $A \equiv C$, then $ \forgetBy{\kappa}{A}(\omega) = 0 \Leftrightarrow \forgetBy{\kappa}{C}(\omega) = 0$.
		\end{description}
		
		\item \agm{6} is satisfied by \( {}^\circ \) iff the following is satisfied:
		\begin{description}
		\item[] If $ \forgetBy{\kappa}{AC}(\omega) = 0$, then
		$ \forgetBy{\kappa}{A}(\omega) = 0$ or $ \forgetBy{\kappa}{C}(\omega) = 0$. 
		\end{description}
		
		\item \agm{7} is satisfied by \( {}^\circ \) iff the following is satisfied:
		\begin{description}
		\item[] If $ \forgetBy{\kappa}{AC}(\ol{A}) = 0$ holds, then $ \forgetBy{\kappa}{A}(\omega) = 0$ implies 
		$ \forgetBy{\kappa}{AC}(\omega) = 0$.
		\end{description}
		
		\item \propCF\ is satisfied by \( {}^\circ \) iff the following is satisfied:
		\begin{description}
		\item[] $ {\forgetBy{\kappa}{AC}}^{-1}(0) = \begin{cases}
		{\forgetBy{\kappa}{A}}^{-1}(0) & \text{ or}\\
	    {\forgetBy{\kappa}{C}}^{-1}(0) & \text{ or}\\
	    {\forgetBy{\kappa}{A}}^{-1}(0) \cup {\forgetBy{\kappa}{C}}^{-1}(0)
		\end{cases}$
		\end{description}
	\end{enumerate}
\end{proposition}

\FARBE{Note that due to the prerequisite that the forgetting operation must not change the \FARBE{domain}, this proposition cannot be applied to marginalization \FARBE{and conditionalization, but to} lifted marginalization.}
\begin{proof}
	We consider each statement independently.
\begin{description}
	\item[{[}\normalfont Statement (1) on \agm{1}.{]}]
	At first, we consider a sequence of equivalences. 
	\begin{align*}
	 \text{${}^\circ$ satisfies \agm{1} } & \Leftrightarrow \text{ $\beliefsOf{\forgetBy{\kappa}{A}} \subseteq \beliefsOf{\kappa}$} \\
	 & \Leftrightarrow \text{ $Th(\{\omega \mid \forgetBy{\kappa}{A} (\omega) = 0\}) \subseteq Th(\{\omega \mid \kappa(\omega) = 0\})$} \\
	 & \Leftrightarrow \text{ $\{\omega \mid \kappa (\omega) = 0\} \subseteq \{\omega \mid \forgetBy{\kappa}{A}(\omega) = 0\}$} 
	\end{align*}
	From Lemma \ref{Lemma_Th} and the latter equivalence,  we easily obtain that $\forgetBy{\kappa}{A}$ satisfies \agm{1} if and only if
	$\kappa(\omega ) = 0$ implies $\forgetBy{\kappa}{A} (\omega) = 0$.
	
	\item[{[}\normalfont Statement (2) on \agm{2}.{]}]
	At first, we provide several equivalences.
	\begin{align*}
		\text{$\forgetBy{\kappa}{A}$ satisfies \agm{2} } & \Leftrightarrow \text{ If \( A \notin \beliefsOf{\kappa} \) implies $\beliefsOf{\kappa} \subseteq \beliefsOf{\forgetBy{\kappa}{A}}$} \tag{\( \star 0 \)}\label{proofS2AGM2_star0} \\ 
	\text{$\beliefsOf{\kappa} \subseteq \beliefsOf{\forgetBy{\kappa}{A}}$ }	& \Leftrightarrow \text{ $Th(\{\omega \mid \kappa(\omega) = 0\}) \subseteq Th(\{\omega \mid \forgetBy{\kappa}{A} (\omega) = 0\})$} \\
		& \Leftrightarrow \text{ $\{\omega \mid \forgetBy{\kappa}{A}(\omega) = 0\} \subseteq \{\omega \mid \kappa(\omega) = 0\}$} \tag{\( \star 1 \)}\label{proofS2AGM2_star1}\\
	\text{\( A \notin \beliefsOf{\kappa} \) }	& \Leftrightarrow \text{ $\kappa(\ol{A}) = 0 $ } \tag{\( \star 2 \)}\label{proofS2AGM2_star2}
	\end{align*}
	Hence, by using \eqref{proofS2AGM2_star1} and \eqref{proofS2AGM2_star2}, we obtain that \( A \notin \beliefsOf{\kappa} \) implies $\beliefsOf{\kappa} \subseteq \beliefsOf{\forgetBy{\kappa}{A}}$ if and only if $\forgetBy{\kappa}{A} (\omega ) = 0$ must imply $\kappa (\omega) = 0$ in the case $\kappa(\ol{A}) = 0 $. Combining the latter with \eqref{proofS2AGM2_star0} is the statement to show.
	
	\item[{[}\normalfont Statement (3) on \agm{3}.{]}]
	The operator \( {}^\circ \) satisfies \agm{3} if  and only if $A \in \beliefsOf{\forgetBy{\kappa}{A}}$ implies $A \equiv \top$.
	The contraposition of the latter is, if $A \not\equiv \top$, then $A \not\in \beliefsOf{\forgetBy{\kappa}{A}}$.
	We obtain Statement (3), as $A \not\in \beliefsOf{\forgetBy{\kappa}{A}}$ holds if and only $\forgetBy{\kappa}{A}(\ol{A}) = 0$ holds.
	
	\item[{[}\normalfont Statement (4) on \agm{4}.{]}]
	\FARBE{In terms of ranking functions, \agm{4} requires that $\beliefsOf{\kappa} \subseteq \Cn(\beliefsOf{\forgetBy{\kappa}{A}}\cup \{A\})$.  We transfer this to the level of possible worlds. Let $\varphi = \bigvee_{\omega: \kappa(\omega) = 0} \omega$ and $\varphi^\circ = \bigvee_{\omega: \forgetBy{\kappa}{A}(\omega) = 0} \omega$.
    Due to Lemma \ref{Lemma_Bel}, the statement to be proven is equivalent to $\Cn(\varphi) \subseteq \Cn(\varphi^\circ, A)$. It can easily be checked that in turn, $\Cn(\varphi) \subseteq \Cn(\varphi^\circ, A)$ is equivalent to $A \Rightarrow \varphi \in \Cn(\varphi^\circ)$.   
    Indeed, if $A \Rightarrow \varphi \in \Cn(\varphi^\circ)$, then \FARBE{$\varphi^\circ \land A \models \varphi$} and hence $\varphi \in \Cn(\varphi^\circ, A)$, therefore also $\Cn(\varphi) \subseteq \Cn(\varphi^\circ, A)$. 
    Conversely, $\Cn(\varphi) \subseteq \Cn(\varphi^\circ, A)$ implies in particular $\varphi \in \Cn(\varphi^\circ, A)$, i.e., 
	\FARBE{${\varphi^\circ \land A} \models \varphi$}, and thus $\varphi^\circ \models A \Rightarrow \varphi$.
	By Lemma \ref{Lemma_Bel} \FARBE{and by \(  \Cn(\varphi^\circ) =\beliefsOf{\forgetBy{\kappa}{A}} \), we have that $A \Rightarrow \varphi \in \beliefsOf{\forgetBy{\kappa}{A}}$ holds if and only if 
		\begin{equation*}
            \FARBE{0 < }
            \forgetBy{\kappa}{A}(\neg(A \Rightarrow \varphi)) = \forgetBy{\kappa}{A}(A \wedge \ol{\varphi})  = \forgetBy{\kappa}{A}(A \wedge \!\!\bigvee_{\omega: \kappa(\omega) > 0} \omega)
		\end{equation*}
        \FARBE{holds. This inequality}
		} is satisfied if and only if $\forgetBy{\kappa}{A}(\omega) >0$ for all $\omega \models A$ with $\kappa(\omega) > 0$ holds.
		Thus, we obtain the desired result.
	}
	
	\item[{[}\normalfont Statement (5) on \agm{5}.{]}]
	\FARBE{The operator ${}^\circ$} satisfies \agm{5} if and only if
	\begin{equation*}
		A \equiv C
		\text{ implies } \beliefsOf{\forgetBy{\kappa}{A}} = \beliefsOf{\forgetBy{\kappa}{C}}
	\end{equation*}
	holds.
	By employing the theory operation \( \Th \) and Lemma \ref{Lemma_Th} (similar to the proofs above), on sees easily that \agm{5}\ is equivalent to $\forgetBy{\kappa}{A} (\omega) = 0$ if and only $\forgetBy{\kappa}{C} (\omega)$ whenever $A \equiv C$ holds.
	
	\item[{[}\normalfont Statement (6) on \agm{6}.{]}]
    \FARBE{The operator ${}^\circ$ satisfies \agm{6} if and only if
    \begin{equation*}
        \beliefsOf{\forgetBy{\kappa}{A}} \cap \beliefsOf{\forgetBy{\kappa}{C}} \subseteq \beliefsOf{\forgetBy{\kappa}{AC}}
    \end{equation*}
    holds.}
	\FARBE{By employing Lemma~\ref{Lemma_Th}, we immediately obtain $\beliefsOf{\forgetBy{\kappa}{A}} \cap \beliefsOf{\forgetBy{\kappa}{C}} = \Th(\{\omega \in \Omega \mid \forgetBy{\kappa}{A}(\omega) = 0 \mbox{ or } \forgetBy{\kappa}{C}(\omega) = 0\})$ \FARBE{and $\beliefsOf{\forgetBy{\kappa}{AC}} = \Th(\{\omega \in \Omega \mid \forgetBy{\kappa}{AC}(\omega) = 0\})$}. The statement then follows via \FARBE{Lemma~\ref{Lemma_Th}}.
    \FARBE{Note that, even though the domain of $\forgetBy{\kappa}{A}$ (or $\forgetBy{\kappa}{C}$, respectively) may not contain $\omega$, the rank of $\omega$ in $\forgetBy{\kappa}{A}$ is well-defined since $\omega$ can be read as a formula in this case.}
	}
	
	\item[{[}\normalfont Statement (7) on \agm{7}.{]}]
	\FARBE{The operator ${}^\circ$} satisfies \agm{7} if and only if 
	\begin{equation*}
		A \not\in \beliefsOf{\forgetBy{\kappa}{AC}} \text{ implies } \beliefsOf{\forgetBy{\kappa}{AC}} \subseteq 	\beliefsOf{\forgetBy{\kappa}{A}}
	\end{equation*}
	As in the proofs of the statements above, by employing the theory operation \( \Th \) and Lemma \ref{Lemma_Th}, one obtains that \agm{7}\ is equivalent to stating that $\forgetBy{\kappa}{AC}(\ol{A}) = 0$ and $ \forgetBy{\kappa}{A}(\omega) = 0$ together imply 
	$ \forgetBy{\kappa}{A C}(\omega) = 0$.
	
	\item[{[}\normalfont Statement (8) on \propCF.{]}] 
    The operator ${}^\circ$ satisfies \propCF if and only if \begin{equation*}
            \beliefsOf{\forgetBy{\kappa}{AC}} = \begin{cases}
                \beliefsOf{\forgetBy{\kappa}{A}} & \text{ or}\\
                \beliefsOf{\forgetBy{\kappa}{C}} & \text{ or}\\
                \beliefsOf{\forgetBy{\kappa}{A}} \cap \beliefsOf{\forgetBy{\kappa}{C}}
            \end{cases} \qquad .
    \end{equation*}
    Recall that $\beliefssymbol$ is defined by applying the $\Th$ operator to the minimal worlds, i.e.\ $\beliefsOf{\kappa} = \Th(\kappa^{-1}(0))$. From this definition we can see immediately that two ranking functions have equal belief sets if and only if they have the same worlds at rank $0$. This means that $\beliefsOf{\forgetBy{\kappa}{AC}} = \beliefsOf{\forgetBy{\kappa}{A}}$ is equivalent to ${\forgetBy{\kappa}{AC}}^{-1}(0) = {\forgetBy{\kappa}{A}}^{-1}(0)$, and that $\beliefsOf{\forgetBy{\kappa}{AC}} = \beliefsOf{\forgetBy{\kappa}{C}}$ is equivalent to ${\forgetBy{\kappa}{AC}}^{-1}(0) = {\forgetBy{\kappa}{C}}^{-1}(0)$. This covers the first two cases above. For the last case, 
    \FARBE{using Lemma~\ref{Lemma_Th}}, we obtain that $\beliefsOf{\forgetBy{\kappa}{AC}} = \beliefsOf{\forgetBy{\kappa}{A}} \cap \beliefsOf{\forgetBy{\kappa}{C}}$ is equivalent to
    \begin{align*}
        \Th\left({\forgetBy{\kappa}{AC}}^{-1}(0)\right) 
        &= \Th\left({\forgetBy{\kappa}{A}}^{-1}(0) \cup {\forgetBy{\kappa}{C}}^{-1}(0)\right)
    \end{align*}
    and therefore equivalent to ${\forgetBy{\kappa}{AC}}^{-1}(0) = {\forgetBy{\kappa}{A}}^{-1}(0) \cup {\forgetBy{\kappa}{C}}^{-1}(0)$. Altogether, we have shown that \propCF\ is equivalent to:
    \begin{equation*}
     \hspace*{2cm}   
        {\forgetBy{\kappa}{AC}}^{-1}(0) = \begin{cases}
            {\forgetBy{\kappa}{A}}^{-1}(0) & \text{ or}\\
            {\forgetBy{\kappa}{C}}^{-1}(0) & \text{ or}\\
            {\forgetBy{\kappa}{A}}^{-1}(0) \cup {\forgetBy{\kappa}{C}}^{-1}(0)     & \hspace{3cm} \qedhere
        \end{cases}
    \end{equation*}
\end{description}
\end{proof}

\FARBE{
    Note that many of the forgetting operators considered here are domain-preserving.
\begin{proposition}
    The forgetting operations lifted marginalization, c-contractions and strategic c-contractions are domain-preserving.
\end{proposition}
}

\FARBE{
\subsection{\FARBE{ASP-inspired Forgetting Postulates for OCFs}}
\label{sec_specifics_asp}

Now we look into the ASP-inspired postulates. Please note that for \propOI, \propE and the Persistence postulates, the interpretation via OCFs is immediate.} \FARBE{Observe that, in contrast to Proposition~\ref{prop:agm-postulates-ocf-characterization}, we allow \FARBE{domain} changes here.} 
\FARBE{However, we assume that the signature is not expanded. Such forgetting operators are called signature-preserving, respectively, signature-reducing:
       \begin{center}
        \begin{tabular}{@{}rp{0.65\textwidth}@{}}
            (\emph{signature-preservation}) & For all $\kappa$ and \( A \), the result of forgetting $\forgetBy{\kappa}{A}$ is defined over the same set signature as $\kappa$, i.e., $\signatureOf{\forgetBy{\kappa}{A}} =  \signatureOf{\kappa}$. \\[1em]
               (\emph{signature-reduction}) & For all $\kappa$ and \( A \), the result of forgetting $\forgetBy{\kappa}{A}$ is defined over a subsignature of $\kappa$, i.e., $\signatureOf{\forgetBy{\kappa}{A}} \subseteq  \signatureOf{\kappa}$.
        \end{tabular}
    \end{center}%
We will see later, that these conditions are natural for many forgetting operators.
}

\begin{proposition}[Characterizations of ASP-Inspired Postulates for OCF-Forgetting]
    \label{prop:asp-postulates-ocf-characterization}
    Let $\kappa, \kappa_1, \kappa_2$ be OCFs over $\Sigma$, and let $\omega \in \Omega_{\Sigma}$ and $A \in \cL_{\Sigma}$. The following statements hold for every \FARBE{signature-preserving or signature-reducing} forgetting operation $\circ$ on the given OCFs.
    \begin{enumerate}
        \renewcommand{\iff}{\ \mbox{iff}\ }
        \renewcommand*\labelenumi{(\theenumi)}%
        \item \propW is satisfied by $^\circ$ iff the following is satisfied:
        \smallskip\\\-\enspace
        If $ \forgetBy{\kappa}{A} \models (C|B)$ holds, then $\kappa \models (C|B)$.
        
        \item \propWCSigma is satisfied by $^\circ$ iff the following is satisfied:
        \smallskip\\\-\enspace
        If $\forgetBy{\kappa}{A}(\omega^{\FARBE{\signatureOf{\forgetBy{\kappa}{A}}}}) = 0$ holds, then $\kappa\big(\omega^{\FARBE{\Sigma}\setminus\signatureMinOf{A}}\big) = 0$.
        
        \item \propSCSigma is satisfied by $^\circ$ iff the following is satisfied:
        \smallskip\\\-\enspace
        If $\kappa\big(\omega^{\FARBE{\Sigma}\setminus\signatureMinOf{A}}\big) = 0$ holds, then $\forgetBy{\kappa}{A}(\omega^{\FARBE{\signatureOf{\forgetBy{\kappa}{A}}}}) = 0$.
        
        \item \propWE is satisfied by $^\circ$ iff the following is satisfied:
        \smallskip\\\-\enspace
        If $\big(\kappa_1(\omega) = 0 \iff \kappa_2(\omega) = 0\big)$ holds, then $\big(\forgetBy{(\kappa_1)}{A}(\omega^{\FARBE{\Sigma'}}) = 0 \iff \forgetBy{(\kappa_2)}{A}(\omega^{\FARBE{\Sigma'}}) = 0\big)$
        \\\-\enspace
        \FARBE{with $\Sigma' = \signatureOf{\forgetBy{(\kappa_1)}{A}} = \signatureOf{\forgetBy{(\kappa_2)}{A}}$}.
        
        \item \propBE is satisfied by $^\circ$ iff the following is satisfied:
        \smallskip\\\-\enspace
        If $\kappa_1 \cong \kappa_2$ holds, then $\big(\forgetBy{(\kappa_1)}{A}(\omega^{\FARBE{\Sigma'}}) = 0 \iff  \forgetBy{(\kappa_2)}{A}(\omega^{\FARBE{\Sigma'}}) = 0\big)$
        \\\-\enspace
        \FARBE{with $\Sigma' = \signatureOf{\forgetBy{(\kappa_1)}{A}} = \signatureOf{\forgetBy{(\kappa_2)}{A}}$}.
        \item \propDE is satisfied by $^\circ$ iff the following is satisfied:
        \smallskip\\\-\enspace
        If $\kappa_1 \cong \kappa_2$ and $A \equiv C$ hold, then $\big(\forgetBy{(\kappa_1)}{A}(\omega^{\FARBE{\Sigma'}}) = 0 \iff \forgetBy{(\kappa_2)}{C}(\omega^{\FARBE{\Sigma'}}) = 0\big)$
        \\\-\enspace
        \FARBE{with $\Sigma' = \signatureOf{\forgetBy{(\kappa_1)}{A}} = \signatureOf{\forgetBy{(\kappa_2)}{A}}$}.    
    \end{enumerate}
\end{proposition}
\begin{proof}
    We consider each statement independently.
    
    \begin{description}
        \item[{[}\normalfont Statement (1) on \propW.{]}] 
        The proof for \propW is immediate from the definition.

        \item[{[}\normalfont Statement (2) on \propWCSigma.{]}]
        \FARBE{The operator $^\circ$ satisfies \propWCSigma if and only if 
        \[
            \margsub{\beliefsOf{\kappa}}{\Sigma\setminus\signatureMinOf{A}}
            \subseteqCNsig{\Sigma} 
            \Bel(\forgetBy{\kappa}{A})
        \]
        holds. 
        According to Lemma~\ref{Lemma_Bel}, Proposition~\ref{prop:variableelemention_disjunction} and Proposition~\ref{prop:bsmargrepresentation_extended}, this is equivalent to 
        \[
            \Cn_{\Sigma} \left( \bigvee\nolimits_{\omega \in \kappa^{-1}(0)} \omega^{\Sigma\setminus\signatureMinOf{A}} \right)
            \subseteqCNsig{\Sigma}
            \Cn_{\signatureOf{\forgetBy{\kappa}{A}}} \left( \bigvee\nolimits_{\omega \in {\forgetBy{\kappa}{A}}^{-1}(0)}\omega \right) \, .
        \]
        Note that the $\subseteqCNsig{\Sigma}$-relation is a $\subseteq$-relation with $\Cn_{\Sigma}$ applied on both sides, and $\Cn_{\Sigma} \circ \Cn_{\Sigma'} = \Cn_{\Sigma}$ for any $\Sigma' \subseteq \Sigma$ \FARBE{due to Lemma~\ref{lem:cn-subsignature}}. Therefore, the equation above is equivalent to
        \[
            \Cn_{\Sigma} \left( \bigvee\nolimits_{\omega \in \kappa^{-1}(0)} \omega^{\Sigma\setminus\signatureMinOf{A}} \right)
            \subseteq
            \Cn_{\Sigma} \left( \bigvee\nolimits_{\omega \in {\forgetBy{\kappa}{A}}^{-1}(0)}\omega \right) \, .
        \]
        Since $\Cn_{\Sigma}(X) \subseteq \Cn_{\Sigma}(Y)$ holds if and only if $\Mod_{\Sigma}(X) \supseteq \Mod_{\Sigma}(Y)$, the above is equivalent to
        \[
            \{ \omega \in \Omega_{\Sigma} \mid \kappa(\omega^{\Sigma\setminus\signatureMinOf{A}}) = 0 \}
            \supseteq
            \{ \omega \in \Omega_{\Sigma} \mid \forgetBy{\kappa}{A}(\omega^{\signatureOf{\forgetBy{\kappa}{A}}}) = 0 \} \, .
        \]
        Therefore, \propWCSigma holds if and only if $\forgetBy{\kappa}{A}(\omega^{\signatureOf{\forgetBy{\kappa}{A}}}) = 0$ implies $\kappa(\omega^{\Sigma\setminus\signatureMinOf{A}}) = 0$ for all $\omega \in \Omega_{\Sigma}$.
        }        
        \item[{[}\normalfont Statement (3) on \propSCSigma.{]}]
        \FARBE{The operator $^\circ$ satisfies \propSCSigma if and only if 
        \[
            \Bel(\forgetBy{\kappa}{A})
            \subseteqCNsig{\Sigma} 
            \margsub{\beliefsOf{\kappa}}{\Sigma\setminus\signatureMinOf{A}}
        \]
        holds. Therefore, this proof is analogous to the proof for \propWCSigma above.
        Using \FARBE{Lemma~\ref{lem:cn-subsignature},} Lemma~\ref{Lemma_Bel}, Proposition~\ref{prop:variableelemention_disjunction} and Proposition~\ref{prop:bsmargrepresentation_extended}, we can rewrite the equation above to
        \[
            \Cn_{\Sigma} \left( \bigvee\nolimits_{\omega \in {\forgetBy{\kappa}{A}}^{-1}(0)}\omega \right)
            \subseteq
            \Cn_{\Sigma} \left( \bigvee\nolimits_{\omega \in \kappa^{-1}(0)} \omega^{\Sigma\setminus\signatureMinOf{A}} \right) \, ,
        \]
        which is equivalent to
        \[
            \{ \omega \in \Omega_{\Sigma} \mid \forgetBy{\kappa}{A}(\omega^{\signatureOf{\forgetBy{\kappa}{A}}}) = 0 \} 
            \supseteq
            \{ \omega \in \Omega_{\Sigma} \mid \kappa(\omega^{\Sigma\setminus\signatureMinOf{A}}) = 0 \}\, .
        \]
        Therefore, \propSCSigma holds if and only if $\kappa(\omega^{\Sigma\setminus\signatureMinOf{A}}) = 0$ implies  $\forgetBy{\kappa}{A}(\omega^{\signatureOf{\forgetBy{\kappa}{A}}}) = 0$ for all $\omega \in \Omega_{\Sigma}$.
        }
        \item[{[}\normalfont Remaining statements.{]}]
        The rest of the statements are easy reformulations of the postulates for ranking functions while Lemma \ref{Lemma_Bel} is taken into regard. \qedhere
    \end{description}
\end{proof}

\FARBE{
Note that all forgetting operators considered here are domain signature-reducing, and many of them also signature-preserving.
    \begin{proposition}
        The forgetting operations lifted marginalization, conditionalization, c-contractions and strategic c-contractions are signature-preserving.
        Marginalization is signature-reducing.
    \end{proposition}
}

\FARBE{
\subsection{\FARBE{Linear Equivalence}\ for OCFs}
\label{sec_specifics_lin_eq}

Furthermore, in the context of OCFs,  it makes sense to consider the behavior of forgetting operations
under linear equivalence (cf.\ Definition \ref{def_ocf_xequiv}). So, we restrict \FARBE{(but also strengthen)} the Equivalence postulate accordingly.}
\smallskip
\FARBE{%
\begin{description}
    \item[\propxE] If $\kappa_1 \xequiv \kappa_2$, then $\forgetBy{(\kappa_1)}{A} \xequiv \forgetBy{(\kappa_2)}{A}$
    \hfill \textbf{(Linear Equivalence)}
\end{description}
}%
We will also consider this postulate in the following, equivalent reformulations that are helpful for technical proofs can be obtained immediately via Proposition \ref{prop:inf_equiv}.

\FARBE{For strategic c-contractions, whereas
Equivalence \propE  cannot be expected in general because the empty layers of OCFs can lead to quite different numerical changes under contraction,  focusing on linear equivalence, however, allows for aligning c-contractions with equivalence by making use of strategies.
A property of strategies that implements the \propxE property %
is the following \FARBE{\cite{AH_KernIsberner-etal_2024-KR}}:
\begin{description}
\item[\postxE]
  $\selectionStrategyContractProp(q \cdot \kappa, A) = q \cdot \selectionStrategyContractProp(\kappa, A)$.
\end{description}

We are now able to show that selection strategies obeying \postxE\ yield c-contractions that satisfy \propxE in general:

\begin{proposition}
 \label{prop_xequiv}
 Let $\selectionStrategyContractProp$ be a selection strategy that fulfills \postxE, and $-_{\sigma}$ its induced c-contraction operator. Then  $-_{\sigma}$ satisfies \propxE.
\end{proposition}
}

\FARBE{
\begin{proof}
Let $\kappa_1, \kappa_2$ be \FARBE{linearly} equivalent OCFs, let $A$ be a proposition in $\cL$, and let $\kappa_i \contractOpSelect A$, $i \in \{1,2\}$, be c-contractions induced by a selection strategy $\selectionStrategyContractProp$ that fulfills \postxE. According to
Proposition \ref{prop_c-contr_gamma},
each $\kappa_i \contractOpSelect A$
 has the form
  		\begin{align*}
  		\kappa_i \contractOpSelect A(\omega) =  - \kappa_i(\ol{A}) + \kappa_i(\omega) + \begin{cases}
				\gamma_i & \text{if } \omega\models A\\
				0 & \text{if } \omega\models \ol{A}
			\end{cases}
		\end{align*}
with $\gamma_i \geq \kappa_i(\ol{A}) - \kappa_i(A)$, $i \in \{1,2\}$. Since $\kappa_1 \xequiv \kappa_2$, we also have  $\kappa_2 = q \cdot \kappa_1$ for some positive rational number $q$. For \propxE, we have to show that
$\kappa_1 \contractOpSelect A  \xequiv \kappa_2 \contractOpSelect A$.

But this is quite obvious: Since $\selectionStrategyContractProp$  fulfills \postxE, we have $\gamma_2 = \selectionStrategyContractProp(\kappa_2, A) =
\selectionStrategyContractProp(q \cdot \kappa_1, A) = q \cdot \selectionStrategyContractProp(\kappa_1, A) = q \gamma_1$, and by Lemma \ref{prop_lin_equiv_ocfs}, we further have $\kappa_2(\ol{A}) = q \cdot \kappa_1(\ol{A})$. Therefore,
 		\begin{align*}
  		\kappa_2 \contractOpSelect A(\omega) =  - q \cdot \kappa_1(\ol{A}) + q \cdot \kappa_1(\omega) + \begin{cases}
				q \cdot \gamma_1 & \text{if } \omega\models A\\
				0 & \text{if } \omega\models \ol{A},
			\end{cases}
		\end{align*}
hence $\kappa_2 \contractOpSelect A = q \cdot (\kappa_1 \contractOpSelect A)$, which shows the linear equivalence of $\kappa_1 \contractOpSelect A$ and  $\kappa_2 \contractOpSelect A$.
\end{proof}
}

\section{Evaluation: Overview}
\label{sec_evaluation_ocf_overview} 
\label{sec:evaluation_agm}
\FARBE{%

In the previous sections, we characterized various types of forgetting operations by high-level properties in Section~\ref{sec:kindsOfForgetting}, elaborated on more detailed properties of forgetting that have been inspired by AGM contraction and ASP forgetting in Section~\ref{sec:general_props}, and presented instantiations of all forgetting operations in the OCF framework in Section~\ref{sec:instantiating}. Here and in the following sections, we evaluate all OCF instantiations of forgetting according to the postulates from Section~\ref{sec:general_props} while
grouping the investigations regarding the more classical operations marginalization and conditionalization in Section \ref{sec_evaluation_ocf_marg_cond}, and considering all forgetting operations based on strategic c-contractions in Section \ref{sec_evaluation_ocf_c-contractions}.

Before going into details, we give an overview on our results which are summarized in Figure~\ref{fig_evaluation_overview}, \FARBE{starting with the postulates inspired by AGM contraction (i.e., the AGM postulates plus \propCF). The ASP-inspired postulates are grouped topically according to whether they affect consequences, deal with equivalence, or address persistence issues.
Please note that
the postulates} \propWCSigma, \propSCSigma, and \propCPSigma are modifications of corresponding properties from ASP forgetting that take explicit account of different signatures a priori vs.\ a posteriori to forgetting.
The postulates \propEP, \propBP, \propBE, \propDE, and \propxE are novel postulates that are specific to our epistemic framework in which we consider forgetting here.

\FARBE{First, let us} focus on \FARBE{answering} the general question \FARBE{of} which of the operations presented in Section~\ref{sec:instantiating} is the ``best'' forgetting \FARBE{operation} according to AGM vs.\ ASP properties. \FARBE{We} clearly see that only minimal c-contractions comply with all AGM axioms, while marginalization performs best with respect to ASP forgetting axioms, with lifted marginalization coming close, violating only \propW. So it appears that (semantic) marginalization implements best the basic ideas of ASP (syntactic) forgetting.
Forgetting by conditionalization and c-revocations satisfy nearly all AGM postulates except for \agm{1}. This proves their generally good compatibility with the AGM ideas. Their violation of \agm{1}, however, is to be expected because forgetting of $A$ by conditionalization  and c-revocations actually implement revisions by $\neg A$, and so the inclusion postulate \agm{1} of AGM contraction is doomed to fail.

Looking now from the perspective of postulates, we notice that there are four resp.\ five postulates that are satisfied by all forgetting operations: \agm{3}, \agm{5}, \propBE, \propDE, and under further prerequisites also \propxE. \agm{3} is the basic success condition that every forgetting operator has to satisfy.  \agm{5}, \propBE, and \propDE deal with semantic equivalences; their satisfaction by all forgetting operations is mostly due to the fact that our operations are defined on the basis of models. However, note that the strong equivalence-related properties \propWE and \propE do not hold for most operations. Here, the ranking-specific postulate \propxE is able to make forgetting compatible with the basic idea of \propE. On the other hand, the failure of \propWE should rather be considered as an intended feature of epistemic change operations---equivalences on the propositional level are not sufficient to preserve equivalences under change; please see also the justification of this behaviour in the context of iterated revision in \cite{DarwichePearl97}.

We start our detailed evaluations with considering forgetting by the basic OCF techniques marginalization and conditionalization in Section~\ref{sec_evaluation_ocf_marg_cond}, and then turn to forgetting operations which are realized by strategic c-contractions in Section~\ref{sec_evaluation_ocf_c-contractions}.
}

\begin{minipage}[t]{\linewidth}
	\newcommand{\opCell}[2]{%
		\begin{tabular}[t]{@{}c@{}}%
			#1\\[-0.2em]%
			\scriptsize{(Def.~#2)}\\
		\end{tabular}%
	}%
	\newcommand{\satCell}[1]{%
		\begin{tabular}[t]{@{}c@{}}%
			\checkmark\\[-0.4em]%
			\scriptsize{(Pro.~#1)}\vspace{0.5em}\\
		\end{tabular}%
	}%
	\newcommand{\conCell}[1]{%
		\begin{tabular}[t]{@{}c@{}}%
			(\checkmark)\makebox[0pt]{\kern.5em*}\\[-0.4em]%
			\scriptsize{(Pro.~#1)}\vspace{0.5em}\\
		\end{tabular}%
	}%
	\newcommand{\vioCell}[1]{%
		\begin{tabular}[t]{@{}c@{}}%
			\xmark\\[-0.4em]%
			\scriptsize{(Pro.~#1)}\vspace{0.5em}%
		\end{tabular}%
	}%
	\newcommand{\gapCell}{\textcolor{magenta}{??}}%
	\centering%
	\begin{tabular}[t]{llccccccc}%
		\toprule
        &
		& \opCell{Marg.}{\ref{def:marginalization}} %
		& \opCell{Lift. M.}{\ref{def:liftung}} %
		& \opCell{Cond.}{\ref{def:conditionalization}} %
		& \opCell{c-Ign.}{\ref{def_cignoration_crevocation}} %
		& \opCell{c-Rev.}{\ref{def_cignoration_crevocation}} %
		& \opCell{\FARBE{min.c-C.}}{\ref{def_minimal_nonminimal_ccontraction}} %
		& \opCell{\FARBE{non-min.c-C.}}{\ref{def_minimal_nonminimal_ccontraction}} \\
		\midrule
        \multirow{16}{*}{\rotatebox[origin=c]{90}{AGM}} &
		\agm{1} & 
			\satCell{\ref{prop:ocfmarg-agm-postulates}} & %
			\satCell{\ref{prop:liftedmarg_agmpostualtes}} & %
			\vioCell{\ref{prop:conditionalization-agm-postulates}} & %
			\satCell{\ref{prop:c-ignoration}} & %
			\vioCell{\ref{prop_results_crevocations}} & %
			\satCell{\ref{prop:min-c-contraction}} & %
			\vioCell{\ref{prop:nonmin-c-contraction-postulates}} \\ %
		& \agm{2} & 
			\vioCell{\ref{prop:ocfmarg-agm-postulates}}
			& %
			\vioCell{\ref{prop:liftedmarg_agmpostualtes}} & %
			\satCell{\ref{prop:conditionalization-agm-postulates}} & %
			\vioCell{\ref{prop:c-ignoration}} & %
			\satCell{\ref{prop_results_crevocations}} & %
			\satCell{\ref{prop:min-c-contraction}} & %
			\vioCell{\ref{prop:nonmin-c-contraction-postulates}} \\ %
		& \agm{3} & 
			\satCell{\ref{prop:ocfmarg-agm-postulates}} & %
			\satCell{\ref{prop:liftedmarg_agmpostualtes}} & %
			\satCell{\ref{prop:conditionalization-agm-postulates}} & %
			\satCell{\ref{prop_ccontr_agmes3}} & %
			\satCell{\ref{prop_ccontr_agmes3}} & %
			\satCell{\ref{prop_ccontr_agmes3}} & %
			\satCell{\ref{prop_ccontr_agmes3}} \\ %
		& \agm{4} & 
			\vioCell{\ref{prop:ocfmarg-agm-postulates}}
			& %
			\vioCell{\ref{prop:liftedmarg_agmpostualtes}} & %
			\satCell{\ref{prop:conditionalization-agm-postulates}} & %
			\vioCell{\ref{prop:c-ignoration}} & %
			\satCell{\ref{prop_results_crevocations}} & %
			\satCell{\ref{prop:min-c-contraction}} & %
			\vioCell{\ref{prop:nonmin-c-contraction-postulates}} \\ %
		& \agm{5} & 
			\satCell{\ref{prop:ocfmarg-agm-postulates}} & %
			\satCell{\ref{prop:liftedmarg_agmpostualtes}} & %
			\satCell{\ref{prop:conditionalization-agm-postulates}} & %
			\satCell{\ref{prop_c-contractions_DE_AGM5}} & %
			\satCell{\ref{prop_c-contractions_DE_AGM5}} & %
			\satCell{\ref{prop_c-contractions_DE_AGM5}} & %
			\satCell{\ref{prop_c-contractions_DE_AGM5}} \\ %
		& \agm{6} & 
			\FARBE{\satCell{\ref{prop:ocfmarg-agm-postulates}}}
			& %
			\vioCell{\ref{prop:liftedmarg_agmpostualtes}} & %
			\satCell{\ref{prop:conditionalization-agm-postulates}} & %
			\vioCell{\ref{prop:c-ignoration}} & %
			\satCell{\ref{prop_results_crevocations}} & %
			\satCell{\ref{prop:min-c-contraction}} & %
			\vioCell{\ref{prop:nonmin-c-contraction-postulates}} \\ %
		& \agm{7} & 
			\FARBE{\vioCell{\ref{prop:ocfmarg-agm-postulates}}} & %
			\vioCell{\ref{prop:liftedmarg_agmpostualtes}} & %
			\satCell{\ref{prop:conditionalization-agm-postulates}} & %
			\vioCell{\ref{prop:c-ignoration}} & %
			\satCell{\ref{prop_results_crevocations}} & %
			\satCell{\ref{prop:min-c-contraction}} & %
			\vioCell{\ref{prop:nonmin-c-contraction-postulates}} \\ %
		& \propCF & 
			\FARBE{\vioCell{\ref{prop:ocfmarg-agm-postulates}}} & %
			\vioCell{\ref{prop:liftedmarg_agmpostualtes}} & %
			\FARBE{\vioCell{\ref{prop:conditionalization-agm-postulates}}} & %
			\vioCell{\ref{prop:c-ignoration}} & %
			\satCell{\ref{prop_results_crevocations}} & %
			\satCell{\ref{prop:min-c-contraction}}
			& %
			\vioCell{\ref{prop:nonmin-c-contraction-postulates}} \\ %
        \midrule
        \multirow{8}{*}{\rotatebox[origin=c]{90}{Consequences}} &
		\propW & 
			\satCell{\ref{prop:ocfmarg-asp-postulates}} & %
			\vioCell{\ref{prop:liftedmarg-asp-postulates}} & %
			\vioCell{\ref{prop:conditionalization_ASPPostulates}} & %
			\vioCell{\ref{prop:c-ignoration}} & %
			\vioCell{\ref{prop_results_crevocations}} & %
			\vioCell{\ref{prop:min-c-contraction}} & %
			\vioCell{\ref{prop:nonmin-c-contraction-postulates}} \\ %
		& \FARBE{\propWCSigma} & 
			\satCell{\ref{prop:ocfmarg-asp-postulates}} & %
			\satCell{\ref{prop:liftedmarg-asp-postulates}} & %
			\vioCell{\ref{prop:conditionalization_ASPPostulates}} & %
			\vioCell{\ref{prop:c-ignoration}} & %
			\vioCell{\ref{prop_results_crevocations}} & %
			\vioCell{\ref{prop:min-c-contraction}} & %
			\vioCell{\ref{prop:nonmin-c-contraction-postulates}} \\ %
		& \FARBE{\propSCSigma} & 
			\satCell{\ref{prop:ocfmarg-asp-postulates}} & %
			\satCell{\ref{prop:liftedmarg-asp-postulates}} & %
			\vioCell{\ref{prop:conditionalization_ASPPostulates}} & %
			\vioCell{\ref{prop:c-ignoration}} & %
			\vioCell{\ref{prop_results_crevocations}} & %
			\vioCell{\ref{prop:min-c-contraction}} & %
			\vioCell{\ref{prop:nonmin-c-contraction-postulates}} \\ %
		& \FARBE{\propCPSigma} & 
			\satCell{\ref{prop:ocfmarg-asp-postulates}} & %
			\satCell{\ref{prop:liftedmarg-asp-postulates}} & %
			\vioCell{\ref{prop:conditionalization_ASPPostulates}} & %
			\vioCell{\ref{prop:c-ignoration}} & %
			\vioCell{\ref{prop_results_crevocations}} & %
			\vioCell{\ref{prop:min-c-contraction}} & %
			\vioCell{\ref{prop:nonmin-c-contraction-postulates}} \\ %
        \midrule
        \multirow{12}{*}{\rotatebox[origin=c]{90}{Equivalence}} &
		\propWE & 
			\satCell{\ref{prop:ocfmarg-asp-postulates}} & %
			\satCell{\ref{prop:liftedmarg-asp-postulates}} & %
			\vioCell{\ref{prop:conditionalization_ASPPostulates}} & %
			\vioCell{\ref{prop:c-ignoration}} & %
			\vioCell{\ref{prop_results_crevocations}} & %
			\vioCell{\ref{prop:min-c-contraction}} & %
			\vioCell{\ref{prop:nonmin-c-contraction-postulates}} \\ %
		& \propE & 
			\satCell{\ref{prop:ocfmarg-asp-postulates}} & %
			\satCell{\ref{prop:liftedmarg-asp-postulates}} & %
			\satCell{\ref{prop:conditionalization_ASPPostulates}} & %
			\vioCell{\ref{prop:c-ignoration}} & %
			\vioCell{\ref{prop_results_crevocations}} & %
			\vioCell{\ref{prop:min-c-contraction}} & %
			\vioCell{\ref{prop:nonmin-c-contraction-postulates}} \\ %
        & \propBE & 
            \FARBE{\satCell{\ref{prop_marg_eval_BE_und_DE}}} & %
            \FARBE{\satCell{\ref{prop:liftedmarg-asp-postulates}}} & %
            \FARBE{\satCell{\ref{prop:conditionalization_ASPPostulates}}} & %
            \satCell{\ref{prop_c-contractions_BE}} & %
            \satCell{\ref{prop_c-contractions_BE}} & %
            \satCell{\ref{prop_c-contractions_BE}} & %
            \satCell{\ref{prop_c-contractions_BE}} \\ %
        & \propDE & 
            \FARBE{\satCell{\ref{prop_marg_eval_BE_und_DE}}} & %
            \FARBE{\satCell{\ref{prop:liftedmarg-asp-postulates}}} & %
            \FARBE{\satCell{\ref{prop:conditionalization_ASPPostulates}}} & %
            \satCell{\ref{prop_c-contractions_DE_AGM5}} & %
            \satCell{\ref{prop_c-contractions_DE_AGM5}} & %
            \satCell{\ref{prop_c-contractions_DE_AGM5}} & %
            \satCell{\ref{prop_c-contractions_DE_AGM5}} \\ %
		& \propxE & 
			\satCell{\ref{prop_marg_eval_lineq}} & %
			\satCell{\ref{prop_lifted_marg_eval_lineq}} & %
			\satCell{\ref{prop_cond_eval_lineq}} & %
			\satCell{\ref{prop_xequiv_ccontr}} & %
			\FARBE{\conCell{\ref{prop_xequiv_ccontr}}} & %
			\satCell{\ref{prop_xequiv_ccontr}} & %
			\FARBE{\conCell{\ref{prop_xequiv_ccontr}}} \\ %
		& \propOI & 
			\satCell{\ref{prop:ocfmarg-asp-postulates}} & %
			\satCell{\ref{prop:liftedmarg-asp-postulates}} & %
			\satCell{\ref{prop:conditionalization_ASPPostulates}} & %
			\vioCell{\ref{prop:c-ignoration}} & %
			\vioCell{\ref{prop_results_crevocations}} & %
			\vioCell{\ref{prop:min-c-contraction}} & %
			\vioCell{\ref{prop:nonmin-c-contraction-postulates}} \\ %
        \midrule
        \multirow{8}{*}{\rotatebox[origin=c]{90}{Persistence}} &
		\propPP & 
		    \satCell{\ref{prop_marg_eval_persistence}}
			& %
	        \FARBE{\satCell{\ref{prop_lm_persistence}}} & %
			\vioCell{\ref{prop_cond_eval_pp}} & %
			\vioCell{\ref{prop:c-ignoration}} & %
			\vioCell{\ref{prop_results_crevocations}} & %
			\vioCell{\ref{prop:min-c-contraction}} & %
			\vioCell{\ref{prop:nonmin-c-contraction-postulates}} \\ %
		& \propNP & 
			\satCell{\ref{prop_marg_eval_persistence}} &
			\FARBE{\satCell{\ref{prop_lm_persistence}}} & %
			\vioCell{\ref{prop_cond_eval_pp}} & %
			\vioCell{\ref{prop:c-ignoration}} & %
			\vioCell{\ref{prop_results_crevocations}} & %
			\vioCell{\ref{prop:min-c-contraction}} & %
			\vioCell{\ref{prop:nonmin-c-contraction-postulates}} \\ %
		& \propEP & 
			\satCell{\ref{prop_marg_eval_persistence}} & %
			\FARBE{\satCell{\ref{prop_lm_persistence}}} & %
			\vioCell{\ref{prop_cond_eval_pp}}
			& %
			\vioCell{\ref{prop:c-ignoration}} & %
			\vioCell{\ref{prop_results_crevocations}} & %
			\vioCell{\ref{prop:min-c-contraction}} & %
			\vioCell{\ref{prop:nonmin-c-contraction-postulates}} \\ %
		& \propBP & 
			\FARBE{\satCell{\ref{prop_marg_eval_persistence}}} & %
			\FARBE{\satCell{\ref{prop_lm_persistence}}} & %
			\FARBE{\vioCell{\ref{prop:conditionalization_ASPPostulates}}} & %
			\vioCell{\ref{prop:c-ignoration}} & %
			\vioCell{\ref{prop_results_crevocations}} & %
			\vioCell{\ref{prop:min-c-contraction}} & %
			\vioCell{\ref{prop:nonmin-c-contraction-postulates}} \\ %
		\bottomrule
	\end{tabular}%
	\captionof{figure}{Fulfilled (\checkmark) and violated (\xmark) postulates for forgetting operations on OCFs as evaluated in Sections~\ref{sec_evaluation_ocf_marg_cond} and~\ref{sec_evaluation_ocf_c-contractions}. 
    \FARBE{The asterisk (*) marks \FARBE{the entries for \propxE} that are only fulfilled if \postxE\ from \FARBE{Section}~\ref{sec_specifics_lin_eq} is satisfied.}
    \label{fig_evaluation_overview}
    }
\end{minipage}

\section{Evaluation: Forgetting by Marginalization and Conditionalization}
\label{sec_evaluation_ocf_marg_cond} 

After having presented an overview of our evaluation results for the various types of forgetting operators in the previous section, in this section we address generell forgetting methods obtained from marginalization and conditionalization of OCFs. In the following three subsections, we elaborate on and prove the evaluation results summarized in the first three columns of Figure~\ref{fig_evaluation_overview}.

\subsection{Forgetting by OCF-Marginalization}
\label{subsec:marginalization}

\FARBE{Forgetting  a formula \( A \) in \( \kappa \) by \ocfMarg removes the signature elements \( \signatureMinOf{A} \) and only take the remaining signature elements into account, i.e.,
\FARBE{$\kappa^{\circ}_A = \margsub{\kappa}{\Sigma \setminus \FARBE{\signatureMinOf{A}}}$
	from Equation~\eqref{eq_forgetting_by_OCF_marginalization}}, which is the marginalization of \( \kappa \) to \( \Sigma\setminus\signatureMinOf{A} \).

}

\begin{proposition}
\label{prop:ocfmarg-agm-postulates}
\FARBE{The following statements hold for forgetting by \ocfMarg:
\begin{itemize}
	\item \agm{1}, \agm{3}, \agm{5}, and \agm{6} are satisfied.
	\item \agm{2}, \agm{4}, \agm{7} and \propCF\ are violated.
\end{itemize}}
\end{proposition}

\begin{proof} 
\FARBE{
We will use the following property, guaranteed by Proposition~\ref{prop:kappa_synmag_extended}:
\begin{equation}
    \beliefsOf{\forgetBy{\kappa}{A}} = \beliefsOf{\kappa} \cap \mathcal{L}_{\Sigma \setminus \signatureMinOf{A}}
     \label{eq:hash:proof:prop:ocfmarg-agm-postulates}\ .
\end{equation}
Next, we show that forgetting by \ocfMarg satisfies \agm{5}.
\begin{description}

\item[{[}\normalfont \agm{5} is satisfied.{]}]
We have to show that \( \beliefsOf{\forgetBy{\kappa}{A}}  = \beliefsOf{\forgetBy{\kappa}{B}} \) holds for equivalent \( A \) and \( B \).
Let \( A,B \) two equivalent formulas, i.e., \( A \equiv B \) and \( \kappa : \Sigma\to\mathbb{N} \) be a ranking function.
The forgetting of \( A \), is the ranking function \( \forgetBy{\kappa}{A} = \margsub{\kappa}{\Sigma \setminus \signatureMinOf{A}}  \) (see Equation~\ref{eq_forgetting_by_OCF_marginalization}).
Analogously, the forgetting of \( B \), is the ranking function \( \forgetBy{\kappa}{A} = \margsub{\kappa}{\Sigma \setminus \signatureMinOf{B}}  \).
Now observe that \( A\equiv B \) implies \( \signatureMinOf{A}=\signatureMinOf{B} \).
Hence, one can see easily, that \( \forgetBy{\kappa}{A} = \forgetBy{\kappa}{B} \) holds and consequently, that \( \beliefsOf{\forgetBy{\kappa}{A}}  = \beliefsOf{\forgetBy{\kappa}{B}} \) holds.

\end{description}
The remaining parts of this proofs rely on satisfaction of  \agm{5}, and we will do not explicitly mention usage of \agm{5}.
Next, we show that {~\ocfMarg~} satisfies \agm{1}, \agm{3}, and \agm{6}.
\begin{description}

\item[{[}\normalfont \agm{1} is satisfied.{]}]
We have to show that \( \beliefsOf{\forgetBy{\kappa}{A}} \subseteq \beliefsOf{\kappa} \) holds. 
As stated by \eqref{eq:hash:proof:prop:ocfmarg-agm-postulates}, \( \beliefsOf{\forgetBy{\kappa}{A}} = \beliefsOf{\kappa} \cap \mathcal{L}_{\Sigma \setminus \signatureMinOf{A}} \) holds.
From that, we obtain \( \beliefsOf{\forgetBy{\kappa}{A}} \subseteq \beliefsOf{\kappa} \) by employing basic set theory.

\item[{[}\normalfont \agm{3} is satisfied.{]}]
We have to show that \( A \in \beliefsOf{\forgetBy{\Psi}{A}} \) implies \( A\equiv\top \).
Assume that \( A \in \beliefsOf{\forgetBy{\Psi}{A}} \) holds and let \( \Sigma = \signatureOf{\kappa} \).
Note that \( \signatureMinOf{A} \subseteq \signatureOf{A} \) and thus, when \( A \in \beliefsOf{\forgetBy{\kappa}{A}} \), then \( \signatureMinOf{A} \subseteq \signatureOf{\forgetBy{\kappa}{A}} \).
Now, recall that according to \eqref{eq:hash:proof:prop:ocfmarg-agm-postulates}, we have that \( \beliefsOf{\forgetBy{\kappa}{A}} = \beliefsOf{\kappa} \cap \mathcal{L}_{\Sigma \setminus \signatureMinOf{A}} \) holds. 
Hence, we have \( \signatureOf{\forgetBy{\kappa}{A}} = \Sigma \setminus \signatureMinOf{A} \).
Consequently, \( A \in \beliefsOf{\forgetBy{\Psi}{A}} \) implies that \( \signatureMinOf{A}=\emptyset \) holds.
As explained in Section~\ref{sec:prelim}, \( \signatureMinOf{A}=\emptyset \) is  only the case if \( A \equiv \top \) or \( A \equiv \bot \) holds.
Because \agm{1} holds, we have that \( \beliefsOf{\forgetBy{\kappa}{A}} \subseteq \beliefsOf{\kappa} \) holds and thus, \( A \in \beliefsOf{\kappa} \).
As \( \kappa \) is an OCF, \( \beliefsOf{\kappa} \) is always consistent and consequently \( \bot \notin \beliefsOf{\kappa} \) holds. 
Hence, \( A \equiv \top \) is the only possible case.

\item[{[}\normalfont \agm{6} is satisfied.{]}]
We have to show that \( \beliefsOf{\forgetBy{\kappa}{A}} \cap \beliefsOf{\forgetBy{\kappa}{C}} \subseteq \beliefsOf{\forgetBy{\kappa}{A\land C}} \) hold.
From \eqref{eq:hash:proof:prop:ocfmarg-agm-postulates}, we obtain:
\begin{align*}
    \beliefsOf{\forgetBy{\kappa}{A}}  & = \beliefsOf{\kappa} \cap \mathcal{L}_{\Sigma \setminus \signatureMinOf{A}}   \\
    \beliefsOf{\forgetBy{\kappa}{C}}  & = \beliefsOf{\kappa} \cap \mathcal{L}_{\Sigma \setminus \signatureMinOf{C}}   \\
    \beliefsOf{\forgetBy{\kappa}{A\land C}}  & = \beliefsOf{\kappa} \cap \mathcal{L}_{\Sigma \setminus \signatureMinOf{A\land C}} 
\end{align*}
Clearly, we have \( \beliefsOf{\forgetBy{\kappa}{A}} \cap \beliefsOf{\forgetBy{\kappa}{C}} = \beliefsOf{\kappa} \cap \mathcal{L}_{\Sigma \setminus \signatureMinOf{A}} \cap \mathcal{L}_{\Sigma \setminus \signatureMinOf{C}} \).
The latter is equivalent to \(\beliefsOf{\forgetBy{\kappa}{A}} \cap \beliefsOf{\forgetBy{\kappa}{C}} = \beliefsOf{\kappa} \cap \mathcal{L}_{\Sigma \setminus (\signatureMinOf{A}\cup  \signatureMinOf{C})} \).
Because we have \( \signatureMinOf{A\land C} \subseteq \signatureMinOf{A} \cup \signatureMinOf{C}  \) (see
Lemma~\ref{lem:sigmin_conjunction}), we have:
\begin{equation*}
   \mathcal{L}_{\Sigma \setminus (\signatureMinOf{A} \cup \signatureMinOf{C})} \subseteq \mathcal{L}_{\Sigma \setminus \signatureMinOf{A\land C}} 
\end{equation*}
Consequently, we also have
\begin{equation*}
    \beliefsOf{\kappa} \cap \mathcal{L}_{\Sigma \setminus (\signatureMinOf{A} \cup \signatureMinOf{C})} \subseteq \beliefsOf{\kappa} \cap \mathcal{L}_{\Sigma \setminus \signatureMinOf{A\land C}} \ .
\end{equation*}

\end{description}
We show that {~\ocfMarg~} violates \agm{2}, \agm{4}, 
\FARBE{\agm{7}} and \propCF.
\begin{description}

\item[{[}\normalfont \agm{2} is violated.{]}]
We have to show that  \( A \notin \beliefsOf{\kappa} \) \emph{does not} imply \( \beliefsOf{\kappa} \subseteq \beliefsOf{\forgetBy{\kappa}{A}}  \).
For demonstrating a violation of \agm{2} we consider the signature \( \Sigma=\{a,b\} \).
Let \( A= a \) and let \( \kappa \) be a ranking function such that \( \beliefsOf{\kappa}=Cn(b) \), i.e., choose \( \kappa \) such that \( \kappa^{-1}(0)=\{ ab,\ol{a}b \} \).
Note that we have  \( A \notin \beliefsOf{\kappa} \) and \( \signatureMinOf{A}=\{a\} \). 
From the latter and \eqref{eq:hash:proof:prop:ocfmarg-agm-postulates}, we obtain \( \beliefsOf{\forgetBy{\kappa}{A}}= \beliefsOf{\kappa} \cap \mathcal{L}_{\Sigma\setminus\signatureMinOf{A}} = \Cn_{\{b\}}(b) \).
Clearly, we have \( b \models_{\Sigma} a \land b  \), yet \( b \not \models_{\{b\}} a \land b  \).
This shows that we have \( \beliefsOf{\kappa} \not\subseteq \beliefsOf{\forgetBy{\kappa}{A}}  \).

\item[{[}\normalfont \agm{4} is violated.{]}]
We have to show that \( \beliefsOf{\kappa} \subseteq \Cn_{\Sigma}(\beliefsOf{\forgetBy{\kappa}{A}}\cup \{A\}) \) \emph{does not} hold.
For demonstrating a violation of \agm{4} we consider the signature \( \Sigma=\{a,b\} \).
Let \( A=a \) and let \( \kappa \) be a ranking function such that \( \beliefsOf{\kappa}=Cn(\neg a \land b) \), i.e., choose \( \kappa \) such that \( \kappa^{-1}(0)=\{ \ol{a}b \} \).
Note that we have  \( \signatureMinOf{A}=\{a\} \).  
From the latter and \eqref{eq:hash:proof:prop:ocfmarg-agm-postulates}, we obtain \( \beliefsOf{\forgetBy{\kappa}{A}} = \beliefsOf{\kappa} \cap \mathcal{L}_{\Sigma\setminus\signatureMinOf{A}} = \Cn_{\{b\}}(b) \).
Hence, we have that \( \Cn_{\Sigma}( \beliefsOf{\forgetBy{\kappa}{A}} \cup \{A\})=\Cn_{\Sigma}(a\land b) \).
This shows that \agm{4} is violated, because we have \( \neg a\land b \in \beliefsOf{\kappa} \) and \( \neg a\land b \notin \Cn_{\Sigma}( \beliefsOf{\forgetBy{\kappa}{A}} \cup \{A\} ) \).

\item[{[}\normalfont \agm{7} is violated.{]}]
We have to show that \( A \notin \beliefsOf{\forgetBy{\kappa}{A\land C}} \) \emph{does not} imply    \( \beliefsOf{\forgetBy{\kappa}{A\land C}} \subseteq \beliefsOf{\forgetBy{\kappa}{A}} \).
For demonstrating a violation of \agm{7} we consider the signature \( \Sigma=\{a,b,c\} \).
Let \( A=a\lor c \) and \( C=a \) be two formulas from \( \mathcal{L}_{\Sigma} \). 
Note that \( A\land C \equiv a \) holds and that we have \( \signatureMinOf{A}=\{a,c\} \) and \( \signatureMinOf{C}=\{a\} \) and \( \signatureMinOf{A\land C}=\signatureMinOf{a}=\{a\} \).
Furthermore, let \( \kappa \) be a ranking function with \( \beliefsOf{\kappa}=\Cn_{\Sigma}(a\land b\land c) \), i.e., choose \( \kappa \) such that \( \kappa^{-1}(0)=\{ abc \} \).
From \eqref{eq:hash:proof:prop:ocfmarg-agm-postulates}, we obtain:
\begin{align*}
    \beliefsOf{\forgetBy{\kappa}{A}}  & = \beliefsOf{\kappa} \cap \mathcal{L}_{\Sigma \setminus \signatureMinOf{A}} = \Cn_{\{b\}} (b)  \\
    \beliefsOf{\forgetBy{\kappa}{A\land C}}  & = \beliefsOf{\kappa} \cap \mathcal{L}_{\Sigma \setminus \signatureMinOf{A\land C}}   = \Cn_{\{b,c\}} (b\land c) 
\end{align*}
From the equations above, we obtain that \(  \beliefsOf{\forgetBy{\kappa}{A\land C}} \not\subseteq \beliefsOf{\forgetBy{\kappa}{A}} \) hold, because, e.g., we have \( b\land c \in \beliefsOf{\forgetBy{\kappa}{A\land C}}  \), but \( b \land c \notin \beliefsOf{\forgetBy{\kappa}{A}}  \).

\item[{[}\normalfont \propCF is violated.{]}]
We have to show that
\begin{equation*}
    \beliefsOf{\forgetBy{\Psi}{A\land C}}=\begin{cases}
        \beliefsOf{\forgetBy{\Psi}{A}} \text{ or} &  \\
        \beliefsOf{\forgetBy{\Psi}{C}} \text{ or} &  \\
        \beliefsOf{\forgetBy{\Psi}{A}} \cap \beliefsOf{\forgetBy{\Psi}{C}} & 
    \end{cases}
\end{equation*}
 \emph{does not} hold.
For demonstrating a violation of \propCF we consider the signature \( \Sigma=\{a,b,c\} \).
Let \( A=a\lor c \) and \( C=a\lor \neg c \), and let \( \kappa \) be a ranking function such that \( \beliefsOf{\kappa}=Cn( a \land b \land c) \), i.e., choose \( \kappa \) such that \( \kappa^{-1}(0)=\{ abc \} \).
Note that \( A\land C \equiv a \) holds and that we have \( \signatureMinOf{A}=\{a,c\} \) and \( \signatureMinOf{C}=\{a,c\} \) and \( \signatureMinOf{A\land C}=\signatureMinOf{a}=\{a\} \).
From \eqref{eq:hash:proof:prop:ocfmarg-agm-postulates}, we obtain:
\begin{align*}
    \beliefsOf{\forgetBy{\kappa}{A}}  & = \beliefsOf{\kappa} \cap \mathcal{L}_{\Sigma \setminus \signatureMinOf{A}} = \Cn_{\{b\}} (b)  \\
    \beliefsOf{\forgetBy{\kappa}{C}}  & = \beliefsOf{\kappa} \cap \mathcal{L}_{\Sigma \setminus \signatureMinOf{C}} = \Cn_{\{b\}} (b)  \\
    \beliefsOf{\forgetBy{\kappa}{A\land C}}  & = \beliefsOf{\kappa} \cap \mathcal{L}_{\Sigma \setminus \signatureMinOf{A\land C}}   = \Cn_{\{b,c\}} (b\land c) 
\end{align*}
From the equations above, we obtain \( \beliefsOf{\forgetBy{\kappa}{A}} \cap \beliefsOf{\forgetBy{\kappa}{C}} = \beliefsOf{\forgetBy{\kappa}{A}}   = \Cn_{\{b\}} (b)\).
Note that we have \( b\land c \in  \beliefsOf{\forgetBy{\kappa}{A\land C}} \), yet \( b\land c \notin  \beliefsOf{\forgetBy{\kappa}{A}} \).
Thus, we obtain that \( \beliefsOf{\forgetBy{\Psi}{A\land C}} \neq \beliefsOf{\forgetBy{\kappa}{A}} \) and  \( \beliefsOf{\forgetBy{\Psi}{A\land C}} \neq \beliefsOf{\forgetBy{\kappa}{C}} \) and  \( \beliefsOf{\forgetBy{\Psi}{A\land C}} \neq \beliefsOf{\forgetBy{\kappa}{A}} \cap \beliefsOf{\forgetBy{\kappa}{C}} \).\qedhere
\end{description}
}

\end{proof}

Forgetting by \ocfMarg ensures that no new beliefs are added but fails to recover the original beliefs when the negation of $A$ was believed before.
This information is completely lost under forgetting by \ocfMarg. 
However, a bit surprisingly, forgetting by \ocfMarg\ respects a kind of coherence, as expressed by \agm{7}, even without the explicit prerequisite $ \beliefsOf{\forgetBy{\kappa}{AB}} \not\models A$.

\begin{proposition}
\label{prop:ocfmarg-asp-postulates}
\FARBE{Forgetting by \ocfMarg  fulfills \propW,
\propWCSigma, \propSCSigma, \propCPSigma,
\propWE,} \propE, and \propOI.
\end{proposition}

\begin{proof} %
	We show all properties for general marginalizations $\margsub{\kappa}{\Sigma'}$, this yields immediately the statement of the proposition for the forgetting operator as a special case of marginalization.

\FARBE{\propW is clear} because of $\margsub{\kappa}{\Sigma'} (AB) = \kappa(AB)$ for $A,B \in \mathcal{L}_{\Sigma'}$.

\FARBE{For \propCPSigma, consider an OCF $\kappa$ over $\Sigma$, and let $A \in \cL_{\Sigma}$. We have to show
$
\beliefsOf{\forgetBy{\kappa}{A}}  
=_{\signatureOf{\kappa}}
\margsub{\beliefsOf{\kappa}}{\signatureOf{\kappa} \setminus \signatureMinOf{A}} 
$
which is obtained by:
\begin{align*}
\beliefsOf{\forgetBy{\kappa}{A}}  
&= \beliefsOf{\margsub{\kappa}{\signatureOf{\kappa} \setminus \signatureMinOf{A}}}  && \textrm{(Equation~\eqref{eq_forgetting_by_OCF_marginalization})}\\
&= \margsub{\beliefsOf{\kappa}}{\signatureOf{\kappa} \setminus \signatureMinOf{A}}  && \textrm{(Proposition~\ref{prop:kappa_synmag})}\\
&=_{\signatureOf{\kappa}}
 \margsub{\beliefsOf{\kappa}}{\signatureOf{\kappa} \setminus \signatureMinOf{A}}  && \textrm{(Equation~\eqref{eq_subseteqCNsig})}
\end{align*}
}
\FARBE{Thus,  \propWCSigma and \propSCSigma hold because   \propCPSigma\ holds
(Proposition~\ref{prop_relations_postulates_asp}).}

	\FARBE{OCF-Marginalization} \FARBE{fulfils} \propWE because $\Bel(\kappa_1) = \Bel(\kappa_2)$ \FARBE{means that} for all $\omega \in \Omega$ it holds that $\kappa_1(\omega) = 0$ if and only \FARBE{if $\kappa_2(\omega)=0$. Furthermore,}
	$\margsub{\kappa_1}{\Sigma'}(\omega') = 0$ if there exists a world $\omega$ with $\omega\models\omega^\prime$ and $\kappa_1(\omega) = 0$,
\FARBE{implying $\kappa_2(\omega) = 0$ and $\margsub{\kappa_2}{\Sigma'}(\omega') = 0$ and thus leading}
to $\Bel(\margsub{\kappa_1}{\Sigma^\prime}) = \Bel(\margsub{\kappa_2}{\Sigma'})$.
	
	\FARBE{For \propE,} let $\kappa_1 \cong \kappa_2$, and let $B,C \in \cL_{\Sigma'}$. 
\FARBE{The derivation}
$(C|B) \in \condOf{\margsub{\kappa_1}{\Sigma'}}$ iff $\margsub{\kappa_1}{\Sigma'}(BC) < \margsub{\kappa_1}{\Sigma'}(B\ol{C})$ iff $\kappa_1(BC) < \kappa_1(B\ol{C})$ \FARBE{holds} due to $B,C \in \cL_{\Sigma'}$.
	With $\kappa_1 \cong \kappa_2$, this holds iff $\kappa_2(BC) < \kappa_2(B\ol{C})$ iff $\margsub{\kappa_2}{\Sigma'}(BC) < \margsub{\kappa_2}{\Sigma'}(B\ol{C})$, i.e., $(C|B) \in \condOf{\margsub{\kappa_2}{\Sigma'}}$. 
	Therefore, $\condOf{\margsub{\kappa_1}{\Sigma'}} = \condOf{\margsub{\kappa_2}{\Sigma'}}$, and hence $\margsub{\kappa_1}{\Sigma'} \cong \margsub{\kappa_2}{\Sigma'}$.
	
	\FARBE{\ocfMarg} fulfills \propOI because the marginalization first to $A$ and then to $B$ is the same as the marginalization to $\{A,B\}$.
	In both cases the rank of a world is determined by taking the minimum rank of a world over the reduced signature $\Sigma \setminus (\Sigma_A \cup \Sigma_B)$.
\end{proof}

\FARBE{%
\begin{proposition}
 \label{prop_marg_eval_BE_und_DE}
 Forgetting by OCF-marginalization 
 \FARBE{satisfies 
 \propBE\ and
 \propDE.}
\end{proposition}
\begin{proof}
\FARBE{
	We start with proving that \propDE is satisfied.
\begin{description}
\item[{[}\normalfont \propDE is satisfied.{]}]

For showing \propDE, we have to show that
$ \beliefsOf{\forgetBy{(\kappa_1)}{A}} = \beliefsOf{\forgetBy{(\kappa_2)}{C}} $
for all formulas $A,C \in \cL$ and for all OCFs $\kappa_1, \kappa_2$ with \( A\equiv C \) and
$\kappa_1 \cong \kappa_2$.
First, note that $A \equiv C$ implies
\( \signatureMinOf{A} =  \signatureMinOf{C}\).
Furthermore, from $\kappa_1 \cong \kappa_2$ we get 
$ \beliefsOf{\kappa_1} = \beliefsOf{\kappa_2} $
by employing Proposition~\ref{prop_ocf_equiv_implies_bel_equal}.
Using $ \beliefsOf{\kappa_1} = \beliefsOf{\kappa_2} $ and applying 
Proposition~\ref{prop:kappa_synmag} twice yields the following sequence of equivalences:
\begin{align*}
	\beliefsOf{\forgetBy{(\kappa_1)}{A}}
	& = \beliefsOf{\margsub{(\kappa_1)}{\Sigma \setminus \signatureMinOf{A}}} \\
	&
	= \margsub{\beliefsOf{\kappa_1}}{\Sigma \setminus \signatureMinOf{A}} 
	= \margsub{\beliefsOf{\kappa_2}}{\Sigma \setminus \signatureMinOf{C}}\\
	&
	\FARBE{= \beliefsOf{\margsub{(\kappa_2)}{\Sigma \setminus \signatureMinOf{C}}}} 
	= \beliefsOf{\forgetBy{(\kappa_2)}{C}}
\end{align*}
This shows that \propDE is satisfied.

\item[{[}\normalfont \propBE is satisfied.{]}]
As given by Proposition~\ref{prop_relations_postulates_asp}, satisfaction of \propDE implies satisfaction of \propBE.
Hence, we have that \propBE is satisfied. \qedhere
\end{description}}
 \end{proof}}

\begin{proposition}
 \label{prop_marg_eval_persistence}
 Forgetting by OCF-marginalization satisfies Epistemic Persistence \FARBE{\propEP} \FARBE{ and hence \FARBE{Belief Persistence \propBP and also both Positive Persistence \propPP and Negative Persistence \propNP}.}
\end{proposition}
\begin{proof}
Let $\FARBE{\signatureOf{\Psi}} = \Sigma_1 \cupdisjoint \Sigma_2$ and $A \in \cL_{\Sigma_2}$. \FARBE{For \propEP, we} have to show that $\margsub{\kappa}{{\Sigma_1}} = \margsub{(\forgetBy{\kappa}{A})}{\Sigma_1}$, where $\forgetBy{\kappa}{A} = \margsub{\kappa}{{\Sigma \backslash \Sigma_A}}$. Since $\Sigma_A \subseteq \Sigma_2$, $\Sigma_1 \subseteq \Sigma \backslash \Sigma_A$ holds, and therefore $\margsub{(\forgetBy{\kappa}{A})} {\Sigma_1}= \margsub{(\margsub{\kappa}{{\Sigma \backslash \Sigma_A}})} {\Sigma_1} = \margsub{\kappa}{{\Sigma_1}}$.
\FARBE{Therefore, \FARBE{\propEP} is satisfied, and due to Proposition \ref{prop_epppnp}, also
\FARBE{\propBP, \propPP, and \propNP}
are fulfilled.}
\qedhere

 \end{proof}
 
\FARBE{Finally, straightforward calculations show that forgetting by OCF-marginalization also satisfies the OCF-specific property of \emph{Linear Equivalence}:}

\FARBE{
\begin{proposition}
 \label{prop_marg_eval_lineq}
 Forgetting by OCF-marginalization satisfies \propxE. 
\end{proposition}
}

\FARBE{
\begin{proof}
Let $ \kappa_1, \kappa_2 $ be OCFs over $ \Sigma $ with $\kappa_1 \xequiv \kappa_2$, and let $q$ be a rational number such that $\kappa_2 = q \cdot \kappa_1$.
For $A \in \cL_{\Sigma}$ we show
\(
(\kappa_2)^{\circ}_A = q \cdot (\kappa_1)^{\circ}_A
\)
for forgetting by OCF-marginalization (cf.\ Equation~\eqref{eq_forgetting_by_OCF_marginalization}) by the following derivation,
where $\omega' \in \Omega_{\Sigma \setminus \signatureMinOf{A}}$:
\begin{align*}
(\kappa_2)^{\circ}_A(\omega')
   &= \margsub{(\kappa_2)}{\Sigma \setminus \signatureMinOf{A}}(\omega')\\
   &=  \min\{ \kappa_2(\omega) \mid \omega\in\Omega_\Sigma \text{ and } \omega\models\omega' \}\\
   &=  \min\{ q \cdot \kappa_1(\omega) \mid \omega\in\Omega_\Sigma \text{ and } \omega\models\omega' \}\\
   &=  q \cdot \min\{ \kappa_1(\omega) \mid \omega\in\Omega_\Sigma \text{ and } \omega\models\omega' \}\\
   &= q \cdot \margsub{(\kappa_1)}{\Sigma \setminus \signatureMinOf{A}}(\omega')\\
   &= q \cdot (\kappa_1)^{\circ}_A(\omega')
   \qedhere
\end{align*}
\end{proof}
}

\subsection{Forgetting by Lifted OCF-Marginalization}
\label{subsec:marginalization_lifted}
In this \FARBE{section,} we will evaluate \FARBE{lifted OCF-marginalization}
\FARBE{as a} forgetting operator. More precisely,  we consider the operator
 \FARBE{$\forgetBy{\kappa}{A} = \margANDembed{\kappa}{\Sigma\setminus\signatureMinOf{A}}{\Sigma}$
given in Equation~\eqref{eq_forgetting_by_lifted_OCF_marginalization}.}
Recall that marginalization of an OCF \( \kappa \) may yield \FARBE{an OCF} over a proper subsignature, but \FARBE{lifted OCF-marginalization} of \( \kappa \) yields always \FARBE{an OCF} over the same signature.
\FARBE{The following observation states that lifted OCF-marginalization
$\forgetBy{\kappa}{A}(\omega)$ is given by applying $\kappa$ to the
$\Sigma\setminus\signatureMinOf{A}$-part of $\omega$.
\begin{proposition}\label{prop_basics_lifted_marginalization_01}
    Let $ \Sigma$ be a signature, let $\kappa$ be an OCF over $\Sigma$, and let $A \in \cL_{\Sigma}$. Then for every $\omega \in \Omega_\Sigma$ we have:  
\begin{align*}
       \forgetBy{\kappa}{A}(\omega) = \kappa(\omega^{\Sigma\setminus\signatureMinOf{A}})
\end{align*}
\end{proposition}
\begin{proof}
Using Equation \eqref{eq_forgetting_by_OCF_marginalization} for
OCF-marginalization and
Equation
\eqref{eq_forgetting_by_lifted_OCF_marginalization} for
lifted OCF-marginalization yields the following derivation:
\begin{align*}
\kappa^{\circ}_A(\omega)
   &= \margANDembed{\kappa}{\Sigma\setminus\signatureMinOf{A}}{\Sigma}(\omega)\\
   &= \margsub{\kappa}{\Sigma \setminus \signatureMinOf{A}}(\omega^{\Sigma\setminus\signatureMinOf{A}})\\
   &=  \min\{ \kappa(\hat{\omega}) \mid \hat{\omega}\in\Omega_\Sigma \text{ and } \hat{\omega} \models \omega^{\Sigma\setminus\signatureMinOf{A}} \}\\
   &= \kappa(\omega^{\Sigma\setminus\signatureMinOf{A}})
   \qedhere
\end{align*}
\end{proof}
}

    Before evaluating forgetting by lifted \FARBE{\ocfMarg,} we characterize \FARBE{the impact of forgetting a formula} \( A \) \FARBE{semantically.}    
\begin{proposition}\label{prop:basics_lifted_marginalization}
    Let $ \Sigma$ be a signature, let $\kappa$ be an OCF over $\Sigma$, and let $A \in \cL_{\Sigma}$.  
    Then the following statements hold:
    \begin{equation*}
        \ModBelOf{\forgetBy{\kappa}{A}} = \{ \omega_1 \in \Omega_\Sigma \mid \exists\omega_2 \in \ModBelOf{\kappa}.\ \omega_1^{\Sigma\setminus\signatureMinOf{A}}=\omega_2^{\Sigma\setminus\signatureMinOf{A}} \}
    \end{equation*}
\end{proposition}
\begin{proof}
    \FARBE{This is a} consequence of Proposition \ref{prop:es_synmag} and the proof of Proposition \ref{prop:variableelemination_model_syntactic}.
\end{proof}

The following propositions \FARBE{present} which AGM-inspired forgetting postulates are satisfied \FARBE{or violated, respectively,} by forgetting via lifted marginalization. %
\begin{proposition}\label{prop:liftedmarg_agmpostualtes}
The following statements hold for forgetting by lifted \FARBE{\ocfMarg}:
\begin{itemize}
    \item \agm{1}, \agm{3} and \agm{5} are \FARBE{satisfied.}
    \item \agm{2}, \agm{4}, \agm{6}, \agm{7} and \propCF\ are \FARBE{violated.}
\end{itemize}
\end{proposition}
\begin{proof}
    We start by \FARBE{addressing the postulates that}
    \FARBE{are satisfied:}
    \begin{description}    	
    	\item[{[}\normalfont \agm{5} is satisfied.{]}]
    	\FARBE{This is} given by the syntax-independence of OCF-marginalization and by the syntax-independence of subsequent lifting.

    	\item[{[}\normalfont \agm{1} is satisfied.{]}]
        \FARBE{We have to show that \( \beliefsOf{\forgetBy{\kappa}{A}} \subseteq \beliefsOf{\kappa} \) holds.
        From Proposition \ref{prop:basics_lifted_marginalization} we obtain that \( \ModBelOf{\kappa} \subseteq \ModBelOf{\forgetBy{\kappa}{A}} \) holds, yielding directly that \agm{1} is satisfied.}

        \item[{[}\normalfont \agm{3} is satisfied.{]}] 
        \FARBE{We have to show that \( A \in \beliefsOf{\forgetBy{\Psi}{A}} \) implies \( A\equiv\top \).
        We show the contraposition of the statement.
        Let \( A\in\cL_{\Sigma} \) be a non-tautological formula. }
        Clearly, if \( \ModBelOf{\forgetBy{\kappa}{A}} = \Omega_\Sigma \), then \( A\notin \beliefsOf{\forgetBy{\kappa}{A}} \). 
        In the following we assume \( \ModBelOf{\forgetBy{\kappa}{A}} \neq \Omega_\Sigma \).
        If \( A\equiv\bot \), then we obtain \( \signatureMinOf{A}=\emptyset \) and  thus, \( \ModBelOf{\kappa} = \ModBelOf{\forgetBy{\kappa}{A}} \) by Proposition \ref{prop:basics_lifted_marginalization}. As \( \ModBelOf{\kappa}\neq \emptyset \), we obtain \( A\notin\beliefsOf{\forgetBy{\kappa}{A}} \). 
        If \( A\not\equiv\bot \), then we have \( \signatureMinOf{A}\neq\emptyset \). 
        Because of  \( \signatureMinOf{A}\neq\emptyset \) and  because of Proposition \ref{prop:sigmin_semantically}, there exists \( \omega\in\Mod(\ol{A}) \) such \FARBE{that}
        \( \omega\in\ModBelOf{\kappa} \) or there exists \( \omega_2\in\ModBelOf{\kappa}\cap\Mod(A) \) such that \( \omega^{\Sigma\setminus\signatureMinOf{A}}=\omega_2^{\Sigma\setminus\signatureMinOf{A}} \).
        From Proposition \ref{prop:basics_lifted_marginalization}, we obtain \( \omega\in\ModBelOf{\forgetBy{\kappa}{A}} \).
        This last observation yields the desired result \( A\notin\beliefsOf{\forgetBy{\kappa}{A}} \).
    \end{description}
     We continue by \FARBE{addressing the postulates that}
     are violated by forgetting by lifted-marginalization.
     \begin{description}    	
     	\item[{[}\normalfont \agm{2} and \agm{4} are violated.{]}]
     	Let \( A \) be a formula with \( \signatureMinOf{A}\neq\emptyset \) and let \( \kappa \) be a ranking function such that \( \ModBelOf{\kappa} \subseteq \Mod(\ol{A}) \).
     	From Proposition~\ref{prop:basics_lifted_marginalization} \FARBE{we} obtain that \( \ModBelOf{\forgetBy{\kappa}{A}} \cap \Mod(A) \neq \emptyset \). 
     	Because of \( \ModBelOf{\kappa} \subseteq \Mod(\ol{A}) \) we have \( A\notin\beliefsOf{\kappa} \), and thus, \( \ModBelOf{\forgetBy{\kappa}{A}} \cap \Mod(A) \neq \emptyset \) implies violation of \agm{2}.
     	Moreover, because of \( \ModBelOf{\kappa} \cap \Mod(A) = \emptyset  \) and \( \ModBelOf{\forgetBy{\kappa}{A}} \cap \Mod(A) \neq \emptyset \), we obtain that \agm{4} is violated.

     	\item[{[}\normalfont \agm{6} is violated.{]}]
         \FARBE{We have to show that \( \beliefsOf{\forgetBy{\kappa}{A}} \cap \beliefsOf{\forgetBy{\kappa}{C}} \subseteq \beliefsOf{\forgetBy{\kappa}{A\land C}} \) \emph{does not} hold.}
     	\FARBE{We assume} that \( \Sigma \) contains more than two elements, in particular let \( a,b,c\in\Sigma \). 
     	Now let \( A=a \) and \( C=b \), for which we have \( \signatureMinOf{A}=\{a\} \) and \( \signatureMinOf{C}=\{b\} \) and \( \signatureMinOf{A\land C}=\{a,b\} \). 
     	Furthermore, let \( \kappa \) be \FARBE{an OCF} over \( \Sigma \) such hat \( \ModBelOf{\kappa}=\{ \omega\in\Omega_\Sigma \mid \omega \models_\Sigma  a\land b\land c \} \).
     	\FARBE{By employing Proposition \ref{prop:kappa_synmag}, Proposition~\ref{prop:bsmargrepresentation_extended} and Proposition~\ref{prop:basics_lifted_marginalization}} we obtain that
     	\begin{equation}\label{eq:prop:liftedmarg_agmpostualtes:s1}\tag{\( \star \)}
     		\begin{aligned}
     			\ModBelOf{\forgetBy{\kappa}{A}} & =  \{  \omega\in\Omega_\Sigma \mid \omega\models b \land c \},  \\
     			\ModBelOf{\forgetBy{\kappa}{C}} & =  \{  \omega\in\Omega_\Sigma \mid \omega\models a \land c \} \text{ and}  \\
     			\ModBelOf{\forgetBy{\kappa}{A\land C}} & =  \{  \omega\in\Omega_\Sigma \mid \omega\models c \}  \\
     		\end{aligned}
     	\end{equation}
     	holds. 
         \FARBE{By considering \eqref{eq:prop:liftedmarg_agmpostualtes:s1}, one obtains that the interpretation \( \omega=\ol{a}\ol{b}c \) satisfies \( \omega \in \ModBelOf{\forgetBy{\kappa}{A\land C}} \).
     	However, from \eqref{eq:prop:liftedmarg_agmpostualtes:s1} we obtain that \( \ModBelOf{\forgetBy{\kappa}{A}}\cup \ModBelOf{\forgetBy{\kappa}{C}} \) does not contain \( \omega \), because \( \omega\not\models b \land c \) and  \( \omega\not\models a \land c \).
     	Consequently, we obtain that \( \ModBelOf{\forgetBy{\kappa}{A\land C}} \not\subseteq \ModBelOf{\forgetBy{\kappa}{A}} \cup \ModBelOf{\forgetBy{\kappa}{C}}  \) hold. 
             The latter is equivalent to \( \beliefsOf{\forgetBy{\kappa}{A}} \cap \beliefsOf{\forgetBy{\kappa}{C}} \not\subseteq \beliefsOf{\forgetBy{\kappa}{A\land C}} \), and thus, that \agm{6} is violated.}
\FARBE{\item[{[}\normalfont \propCF is violated.{]}]
Recall that according to Proposition~\ref{prop_relations_postulates_agm}, satisfaction of \propCF implies satisfaction of \agm{6}.
Hence, because \agm{6} is violated, we also have that \propCF is violated.}
     	\item[{[}\normalfont \agm{7} is violated.{]}]     	
         \FARBE{
             For showing that \agm{7} is violated, we have to show that \( A \notin \beliefsOf{\forgetBy{\kappa}{A\land C}} \) \emph{does not} imply    \( \beliefsOf{\forgetBy{\kappa}{A\land C}} \subseteq \beliefsOf{\forgetBy{\kappa}{A}} \).
             For demonstrating a violation of \agm{7} we consider the signature \( \Sigma=\{a,b,c\} \).            
     	Now let \( A=a\lor b \) and \( C=a \), for which we have \( C\equiv A\land C \), and \( \signatureMinOf{A}=\{a,b\} \) and \( \signatureMinOf{C}=\signatureMinOf{A\land C}=\{a\} \). 
        }
     	Furthermore, let \( \kappa \) be \FARBE{an OCF} over \( \Sigma \) such \FARBE{that \( \ModBelOf{\kappa}=\modelsOf{Cn(\neg a\land \neg b)} = \{ \ol{a}\ol{b}c, \ol{a}\ol{b}\ol{c} \} \)).
         By employing Proposition~\ref{prop:kappa_synmag_extended} we obtain the following for the marginalization of \( \kappa \) by \( A \), respectively, \(  A \land C \):
         \begin{align*}
             \beliefsOf{\margsub{\kappa}{\Sigma\setminus\signatureMinOf{A}}}  & = \beliefsOf{\kappa} \cap \mathcal{L}_{\Sigma \setminus \signatureMinOf{A}} = Cn_{\{c\}}(\top)\\
             \beliefsOf{\margsub{\kappa}{\Sigma\setminus\signatureMinOf{A\land C}}}  & = \beliefsOf{\kappa} \cap \mathcal{L}_{\Sigma \setminus \signatureMinOf{A\land C}} = Cn_{\{b,c\}}(\neg b)
         \end{align*}
         Semantically, we obtain the following:
         \begin{align*}
             \ModBelOf{\margsub{\kappa}{\Sigma\setminus\signatureMinOf{A}}}  & = \{ c, \ol{c} \}\\
             \ModBelOf{\margsub{\kappa}{\Sigma\setminus\signatureMinOf{A\land C}}}  & = \{\ol{b}c,\ol{b}\ol{c}\}
         \end{align*}
         Thus, by employing Proposition~\ref{prop:basics_lifted_marginalization}, we obtain the following for the forgetting by lifted \ocfMarg of \( A \), respectively, \( A\land C \):
         \begin{align*}
             \ModBelOf{\forgetBy{\kappa}{A}}  & = \{ abc, ab\ol{c},  a\ol{b}c, a\ol{b}\ol{c},  \ol{a}bc, \ol{a}b\ol{c},  \ol{a}\ol{b}c, \ol{a}\ol{b}\ol{c} \}\\
             \ModBelOf{\forgetBy{\kappa}{A}}  & = \{ a\ol{b}c, a\ol{b}\ol{c}, \ol{a}\ol{b}c, \ol{a}\ol{b}\ol{c} \}
         \end{align*}
         We observe that \( abc \in \ModBelOf{\forgetBy{\kappa}{A}}  \), yet \( \ModBelOf{\forgetBy{\kappa}{A}} \).
         }
     	This shows that \( \ModBelOf{\forgetBy{\kappa}{A}} \not\subseteq \ModBelOf{\forgetBy{\kappa}{A\land C}}  \) and thus, that \agm{7} is violated.
     	\qedhere
     \end{description}
\end{proof}

The next proposition states which of the ASP-inspired postulates are satisfied
\FARBE{or violated, respectively,} by lifted marginalization as forgetting.
\FARBE{\begin{proposition}
\label{prop:liftedmarg-asp-postulates}
    The following statements hold for forgetting by lifted marginalization
    \begin{itemize}
        \item \propWCSigma, \propSCSigma, \propCPSigma, \propWE, \propE, \propOI, \FARBE{\propBE and \propDE} are \FARBE{satisfied.}
        \item \propW \FARBE{is violated.}
    \end{itemize}
\end{proposition}
\begin{proof}
We show satisfaction, respectively the violation, of postulates:
\begin{description}
    \item[{[}\normalfont \propCPSigma is satisfied.{]}]     
        Let \( \Sigma=\signatureOf{\kappa} \).
        We have to show that \[ \FARBE{ \beliefsOf{\forgetBy{\kappa}{A}} \gleichCNsig{\Sigma} \margsub{\beliefsOf{\kappa}}{\Sigma\setminus\signatureMinOf{A}} } \] is satisfied.
    First, note that we have
    \begin{equation*}
        \beliefsOf{\margsub{\kappa}{\Sigma\setminus\signatureMinOf{A}}} = \beliefsOf{\kappa} \cap \mathcal{L}_{\Sigma \setminus \signatureMinOf{A}}
    \end{equation*}
     due to Proposition~\ref{prop:kappa_synmag_extended}.
     Then, by employing Proposition~\ref{prop:basics_lifted_marginalization}, we obtain:
     \begin{equation*}         
         \Mod_{\Sigma}(\margsub{\Bel(\kappa)}{\Sigma\setminus\signatureMinOf{A}})
         = \{ \omega \in \Omega_{\Sigma} \mid \kappa(\omega^{\Sigma\setminus\signatureMinOf{A}}) = 0 \}
     \end{equation*}
     Observe that the following chain of equivalences holds:
    \begin{align*}
        \Mod_{\Sigma}(\margsub{\Bel(\kappa)}{\Sigma\setminus\signatureMinOf{A}})
        &= \{ \omega \in \Omega_{\Sigma} \mid \kappa(\omega^{\Sigma\setminus\signatureMinOf{A}}) = 0 \} \\
        &= \{ \omega \in \Omega_{\Sigma} \mid \margsub{\kappa}{\Sigma\setminus\signatureMinOf{A}}(\omega^{\Sigma\setminus\signatureMinOf{A}}) = 0 \} \\
        &= \{ \omega \in \Omega_{\Sigma} \mid \margANDembed{\kappa}{\Sigma\setminus\signatureMinOf{A}}{\Sigma}(\omega) = 0\} \\
        &= \FARBE{\Mod_{\Sigma}(\Bel(\forgetBy{\kappa}{A}))}.
    \end{align*}
    Therefore, $\margsub{\Bel(\kappa)}{\Sigma\setminus\signatureMinOf{A}} =_{\Sigma} \beliefsOf{\forgetBy{\kappa}{A}}$ and \FARBE{\propCPSigma} is fulfilled.
    
    \item[{[}\normalfont \propWCSigma and \propSCSigma are satisfied.{]}]   
    By employing Proposition~\ref{prop_relations_postulates_asp}, we obtain the satisfaction of \FARBE{\propWCSigma} and \FARBE{\propSCSigma} from the satisfaction of \FARBE{\propCPSigma}.

    \FARBE{
        \item[{[}\normalfont \propWE is satisfied.{]}]   
        We have to show that
        \begin{equation*}
            \text{If } \beliefsOf{\kappa_1} = \beliefsOf{\kappa_2} \text{, then }  \beliefsOf{\forgetBy{(\kappa_1)}{A}} = \beliefsOf{\forgetBy{(\kappa_2)}{A}}
        \end{equation*}
        is satisfied.       
        From satisfaction of \propCPSigma and \( \beliefsOf{\kappa_1} = \beliefsOf{\kappa_2} \) one obtains directly that 
        \( \beliefsOf{\forgetBy{(\kappa_1)}{A}} = \margsub{\beliefsOf{\kappa_1}}{\Sigma\setminus\signatureMinOf{A}} = \beliefsOf{\forgetBy{(\kappa_2)}{A}} \) holds.
    }
    
    \item[{\FARBE{[\normalfont \propE is satisfied.]}}]       
        Let \( \Sigma=\signatureOf{\kappa_1}=\signatureOf{\kappa_2} \).
        We have to show that \[ \text{If $\kappa_1 \cong \kappa_2$, then $ \forgetBy{(\kappa_1)}{A} \cong \forgetBy{(\kappa_2)}{A} $}  \] is satisfied.
        \FARBE{By employing Proposition~\ref{prop_ocf_equiv_preserved_under_marginalization} and $\kappa_1 \cong \kappa_2$, we obtain 
        \begin{equation*}
        \margsub{(\kappa_1)}{\Sigma \setminus \signatureMinOf{A}} \cong \margsub{(\kappa_2)}{\Sigma \setminus \signatureMinOf{A}}.        
        \end{equation*}
        From the latter we obtain 
        $\embedsup{\margsub{(\kappa_1)}{\Sigma \setminus \signatureMinOf{A}}}{\Sigma} \cong \embedsup{\margsub{(\kappa_2)}{\Sigma \setminus \signatureMinOf{A}}}{\Sigma}$
        by employing Proposition~\ref{prop:isoToLifting}.
        This shows that we have 
        $ \forgetBy{(\kappa_1)}{A} \cong \forgetBy{(\kappa_2)}{A} $.}
    
    \item[{[}\normalfont \propBE is satisfied.{]}]  
    Let \( \Sigma=\signatureOf{\kappa_1}=\signatureOf{\kappa_2} \).
    We have to show that \[ \text{If $\kappa_1 \cong \kappa_2$, then $ \beliefsOf{\forgetBy{(\kappa_1)}{A}} = \beliefsOf{\forgetBy{(\kappa_2)}{A}} $}  \] is satisfied.
    Because \propE is satisfied, we obtain $ \forgetBy{(\kappa_1)}{A} \cong \forgetBy{(\kappa_2)}{A} $.
        Consequently, we have
        $ \beliefsOf{\forgetBy{(\kappa_1)}{A}} = \beliefsOf{\forgetBy{(\kappa_2)}{A}} $
        \FARBE{according to Proposition~\ref{prop_ocf_equiv_implies_bel_equal}
            as required.}
    
    \item[{[}\normalfont \propDE is satisfied.{]}]   
        By employing satisfaction of \agm{5}, we obtain the satisfaction of {\propBE} from the satisfaction of {\propDE}.

    \item[{\FARBE{[\normalfont \propOI is satisfied.]}}]   
    \FARBE{
        Let \( \Sigma=\signatureOf{\kappa} \).
        We have to show that \[ \forgetBy{(\forgetBy{\kappa}{A})}{B} \cong \forgetBy{(\forgetBy{\kappa}{B})}{A}  \] is satisfied.
        Moreover, let \( \Sigma'=\Sigma\setminus\signatureMinOf{A} \) and let \( \Sigma''=\Sigma\setminus\signatureMinOf{B} \).
        Recall that \( \forgetBy{(\forgetBy{\kappa}{A})}{B} = \margANDembedANDmargANDembed{\kappa}{\Sigma'}{\Sigma}{\Sigma''}{\Sigma} \) and \( \forgetBy{(\forgetBy{\kappa}{A})}{B} = \margANDembedANDmargANDembed{\kappa}{\Sigma''}{\Sigma}{\Sigma'}{\Sigma} \).
        From Proposition~\ref{prop:margeorderproposition}~(3), we obtain the following two equations:
        \begin{align*}
        	\margANDembedANDmarg{\kappa}{\Sigma'}{\Sigma}{\Sigma''} & = \margANDembed{\kappa}{(\Sigma'\cap\Sigma'')}{\Sigma''} \\
        	\margANDembedANDmarg{\kappa}{\Sigma''}{\Sigma}{\Sigma'} & = \margANDembed{\kappa}{(\Sigma'\cap\Sigma'')}{\Sigma'}
        \end{align*}         
        From the latter two equations, \( \Sigma' \subseteq \Sigma \), \( \Sigma'' \subseteq \Sigma \), and Proposition~\ref{prop:margeorderproposition}~(2), we obtain the following line of equivalences: 
        \begin{align*}
            \forgetBy{(\forgetBy{\kappa}{A})}{B} & =  \margANDembedANDmargANDembed{\kappa}{\Sigma'}{\Sigma}{\Sigma''}{\Sigma}  =
            \margANDembedANDembed{\kappa}{(\Sigma'\cap\Sigma'')}{\Sigma''}{\Sigma}  \stackrel{\text{\tiny Prop~\ref{prop:margeorderproposition}}}{=}
            \margANDembed{\kappa}{(\Sigma'\cap\Sigma'')}{(\Sigma''\cup\Sigma)}                                                                       \\
                                                 & = 
                                                 \margANDembed{\kappa}{(\Sigma'\cap\Sigma'')}{(\Sigma)}                                                                                                                                                                                                                                 \\
                                                 & = 
                                                 \margANDembed{\kappa}{(\Sigma'\cap\Sigma'')}{(\Sigma'\cup\Sigma)} \stackrel{\text{\tiny Prop~\ref{prop:margeorderproposition}}}{=}  \margANDembedANDembed{\kappa}{(\Sigma'\cap\Sigma'')}{\Sigma'}{\Sigma} =  \margANDembedANDmargANDembed{\kappa}{\Sigma''}{\Sigma}{\Sigma'}{\Sigma}\\
                                                 & = \forgetBy{(\forgetBy{\kappa}{B})}{A}
        \end{align*}
        This shows that \( \forgetBy{(\forgetBy{\kappa}{A})}{B} = \forgetBy{(\forgetBy{\kappa}{B})}{A} \) holds.
        The desired statement \( \forgetBy{(\forgetBy{\kappa}{A})}{B} \cong \forgetBy{(\forgetBy{\kappa}{B})}{A} \) is a consequence of the latter.
    } 
    
    \item[{[}\normalfont \propW is violated.{]}]   
    \FARBE{We have to show that 
    \begin{equation*}
        \kappa \nmmodels \forgetBy{\kappa}{A}
    \end{equation*}
     is violated.
    Let \( \Sigma=\{ a,b \} \) and let \( \kappa \) be \FARBE{an OCF} with \( \kappa({a}{b}) = 0 \) and \( \kappa(\ol{a}{b}) = \kappa({a}\ol{b}) = \kappa(\ol{a}\ol{b}) = 1 \).
    For \( \kappa \) holds  \( ({b}|\ol{a}) \notin \Cons(\kappa) \).
    When forgetting the signature element \( a \), one obtains \( \forgetBy{\kappa}{a}({a}{b}) =  \forgetBy{\kappa}{a}(\ol{a}{b}) = 0 \) and \( \forgetBy{\kappa}{a}({a}\ol{b}) = \forgetBy{\kappa}{a}(\ol{a}\ol{b}) = 1 \).
    We observe that \( ({b}|\ol{a}) \in \Cons(\forgetBy{\kappa}{a}) \) holds.
    Consequently, we have that \( \Cons(\forgetBy{\kappa}{a}) \) is not a subset of \( \Cons(\kappa) \).
    This yields \( \kappa \notnmmodels \forgetBy{\kappa}{A} \), which is a violation of \propW.
    }    
    \qedhere    
\end{description}
\end{proof}}

\FARBE{
Now we turn to investigating the Persistence properties. 

\begin{proposition}
 \label{prop_lm_persistence}
 Forgetting by lifted OCF-marginalization satisfies \propEP, and hence also \propPP, \propNP, and \propBP. 
\end{proposition}

\begin{proof}
 Let $ \signatureOf{\Psi} = \Sigma_1 \cupdisjoint \Sigma_2$ and $A \in \cL_{\Sigma_2}$. According to (\ref{eq_forgetting_by_lifted_OCF_marginalization}), we have to show that  $\margsub{\kappa}{{\Sigma_1}} = \margsub{(\margANDembed{\kappa}{\Sigma\setminus\signatureMinOf{A}}{\Sigma})}{\Sigma_1}$.

Note that due to $A \in \cL_{\Sigma_2}$ and since $\Sigma_1, \Sigma_2$ are disjoint, each $\omega \in \Omega_{\Sigma}$ can be written in the form $\omega = \omega^{\Sigma_1} \omega^{\Sigma_2\setminus\signatureMinOf{A}} \omega^{\signatureMinOf{A}}$. 
Let $\omega_1 \in \Omega_{\Sigma_1}$. By applying
\FARBE{lifted \ocfMarg (cf. Equations \eqref{def_marginalization_lifted} and \eqref{eq_forgetting_by_lifted_OCF_marginalization}),}
we have 
\begin{eqnarray*}
 \margsub{(\margANDembed{\kappa}{\Sigma\setminus\signatureMinOf{A}}{\Sigma})}{\Sigma_1} (\omega_1) 
 &=& \margANDembed{\kappa}{\Sigma\setminus\signatureMinOf{A}}{\Sigma}(\omega_1) \\
  &=& \min_{\omega \models \omega_1} \margANDembed{\kappa}{\Sigma\setminus\signatureMinOf{A}}{\Sigma}(\omega) \\
  &=& \min_{\omega \models \omega_1} \kappa(\omega^{\Sigma_2\setminus\signatureMinOf{A}}) \\
   &=& \min_{\omega \models \omega_1} \min_{\omega' \models \omega^{\Sigma_2\setminus\signatureMinOf{A}}} \kappa(\omega') \\
   &=& \min_{\omega, \omega' \in \Omega_{\Sigma}} \kappa(\omega_1 \omega^{\Sigma_2\setminus\signatureMinOf{A}} {\omega'}^{\signatureMinOf{A}}) \\
   &=& \min_{\omega \in \Omega_{\Sigma}} \kappa(\omega_1 \omega^{\Sigma_2}) \\
&=& \kappa(\omega_1) \\
&=& \margsub{\kappa}{{\Sigma_1}}(\omega_1), 
\end{eqnarray*}
which was to be shown. Hence, forgetting by lifted OCF-marginalization satisfies \propEP. 

According to Proposition \ref{prop_epppnp}, forgetting by lifted OCF-marginalization also satisfies \propPP, \propNP, and \propBP. 
\end{proof}
}

\FARBE{Analogously to the situation when applying \FARBE{OCF-marginalization,} also  forgetting by lifted OCF-marginalization satisfies  \emph{Linear Equivalence}:}

\FARBE{
\begin{proposition}
 \label{prop_lifted_marg_eval_lineq}
 Forgetting by lifted OCF-marginalization satisfies \propxE. 
\end{proposition}
}

\FARBE{
\begin{proof}
Let $ \kappa_1, \kappa_2 $ be OCFs over $ \Sigma $ with $\kappa_1 \xequiv \kappa_2$, and let $q$ be a rational number such that $\kappa_2 = q \cdot \kappa_1$.
For $A \in \cL_{\Sigma}$ we show
\(
(\kappa_2)^{\circ}_A = q \cdot (\kappa_1)^{\circ}_A
\)
for forgetting by lifted OCF-marginalization (cf.\ Equation~\eqref{eq_forgetting_by_lifted_OCF_marginalization}) by the following derivation:
\begin{align*}
(\kappa_2)^{\circ}_A
   &= \margANDembed{(\kappa_2)}{(\Sigma \setminus \signatureMinOf{A})}{\Sigma}\\
   &= (\margsub{(\kappa_2)}{(\Sigma \setminus \signatureMinOf{A})})_{\uparrow \Sigma}\\
   &= (q \cdot \margsub{(\kappa_1)}{(\Sigma \setminus \signatureMinOf{A})})_{\uparrow \Sigma}
   \quad\quad\quad
(\textrm{according to Proposition~\ref{prop_marg_eval_lineq}})
\\
   &= q \cdot \margANDembed{(\kappa_1)}{(\Sigma \setminus \signatureMinOf{A})}{\Sigma}\\
   &= q \cdot (\kappa_1)^{\circ}_A
   \qedhere
\end{align*}
\end{proof}
}

\subsection{Forgetting by OCF-Conditionalization}
\label{subsec:conditionalization}
Another form of forgetting is \ocfCond which restricts the models of a ranking function to a specific context, forgetting any model outside of the context. 
\FARBE{According to \eqref{eq_forgetting_by_OCF_condionalization}, \ocfCond is defined by $  \kappa^{\circ}_A = \kappa|\overline{A}$, i.e., $\kappa^{\circ}_A(\omega) = \kappa(\omega) - \kappa(\ol{A})$ for all $\omega \models \ol{A}$.}

\begin{figure}
    \centering    
    \begin{ocftable}{3}
    	\ocf{$ \kappa $}{
    		{$ \overline{a}{b} $},
    		{$ {a}\overline{b} , \overline{a}\overline{b} $},
    		{$ {a}{b} $}
    	}
    	\ocf{$ \kappa^\circ_a=\kappa{|}\overline{a} $}{
    	{$  $},
    	{$ \overline{a}{b} $},
    	{$ \overline{a}\overline{b} $}
    	}
    \end{ocftable}
    \caption{Counterexample for Conditionalization for (AMGes-1)}
    \label{tab:ce_cond_1}
\end{figure}
\begin{proposition}
\label{prop:conditionalization-agm-postulates}
\OcfCond does not fulfill \agm{1}, but fulfills \agm{2}, \agm{3}, \agm{4}, \agm{5}, \agm{6}, \agm{7}, and \propCF.
\end{proposition}
\begin{proof}
\FARBE{We show the satisfaction, respectively, the violation, of postulates:}
\begin{description}
    \item[{[\normalfont \agm{1} is violated.]}]    
    We have to show that $ \beliefsOf{\forgetBy{\kappa}{A}} \subseteq \beliefsOf{\kappa} $ is violated.
    Let $\kappa$ be the OCF over $\Sigma=\{a,b\}$ with $\kappa({a}{b})=0$, $\kappa({a}\overline{b})=\kappa(\overline{a}\,\overline{b})=1$, and $\kappa(\overline{a}{b})=2$ as shown in \autoref{tab:ce_cond_1}.
    The forgetting of $\kappa$ by $a$ results in $\forgetBy{\kappa}{a}=\kappa{|}\overline{a} : \Mod(\ol{a}) \to \naturals $ with $\forgetBy{\kappa}{a}(\overline{a}\,\overline{b})=0$ and $\forgetBy{\kappa}{a}(\overline{a}{b})=1$ (cf. \autoref{tab:ce_cond_1}).
    We get \FARBE{$\beliefsOf{\kappa} = \Cn(ab)$ but $\beliefsOf{\forgetBy{\kappa}{a}} = \Cn(\overline{a}\,\overline{b})$. Hence obviously,  \agm{1} is not fulfilled.}
    
    \item[{[}\normalfont \agm{2} is satisfied.{]}]    
    We have to show that if $ A \notin \beliefsOf{\kappa} \text{, then } \beliefsOf{\kappa} \subseteq \beliefsOf{\forgetBy{\kappa}{A}} $.
    \FARBE{$A \not\in \beliefsOf{\kappa}$ means that $\kappa(\notA) = 0$. This implies $\kappa^{\circ}_A(\omega) = \kappa(\omega)$ for all $\omega \models \ol{A}$.
        According to \eqref{eq_bel_th}, this gives us
        $ \beliefsOf{\kappa} = Th(\{ \omega \, | \, \kappa(\omega) = 0 \})$, and
        $ \beliefsOf{\forgetBy{\kappa}{A}} = Th(\{ \omega \models \ol{A} \, | \, \kappa(\omega) = 0 \})$. Because the $\Th$-operator is antitone (see Lemma \ref{Lemma_Th}), we obtain
        $ \beliefsOf{\kappa} \subseteq \beliefsOf{\forgetBy{\kappa}{A}}$,} and hence, \agm{2} is fulfilled.
    
    \item[{[}\normalfont \agm{3} is satisfied.{]}]    
    We have to show that $  A \in \beliefsOf{\forgetBy{\kappa}{A}} $ implies $ A\equiv\top $.
        For a ranking function \( \kappa \), forgetting by OCF-conditionalition of \( A \) is \( \forgetBy{\kappa}{A} = \kappa|\ol{A} \).
        If \( A  \in \beliefsOf{\forgetBy{\kappa}{A}} \) holds, then also \( \{ \omega \FARBE{\models \ol{A}} \mid \kappa|\ol{A}(\omega) = 0 \} \subseteq \Mod(A) \) \FARBE{ holds.
        However, \( \{ \omega \models \ol{A} \mid \kappa|\ol{A}(\omega) = 0 \} \subseteq \Mod(\ol{A}) \) which is disjoint from \(  \Mod(A) \).
        Hence, \( \{ \omega \models \ol{A} \mid \kappa|\ol{A}(\omega) = 0 \} = \emptyset \), which is possible only if \( \Mod(\ol{A}) = \emptyset \).
        Therefore, we obtain %
        }
        \( A\equiv\top \).

    \item[{[}\normalfont \agm{4} is satisfied.{]}]
    We have to show that $ \beliefsOf{\kappa} \subseteq \Cn(\beliefsOf{\forgetBy{\kappa}{A}}\cup \{A\}) $ holds.
    \agm{4} is fulfilled because the forgetting of $A$ by OCF-conditionalization results in a ranking function over the models of $\notA$.
    So it is always the case that \FARBE{$  \notA \in \beliefsOf{\forgetBy{\kappa}{A}}$ holds.
        By this we get $ \Cn(\beliefsOf{\forgetBy{\kappa}{A}} \cup \{A\}) = \cL_{\Sigma}$, and hence  $\beliefsOf{\kappa} \subseteq \Cn(\beliefsOf{\forgetBy{\kappa}{A}} \cup \{A\})$ holds for all $\kappa$.}
    
    \item[{[}\normalfont \agm{5} is satisfied.{]}]    
    We have to show that  $ A\equiv C $ implies $  \beliefsOf{\forgetBy{\kappa}{A}} = \beliefsOf{\forgetBy{\kappa}{C}} $.
    Clearly, this property is satisfied, because forgetting by OCF-conditionalition is defined semantically.
    
    \item[{[\normalfont \agm{6} is satisfied.]}]    
    Because \propCF implies \agm{6} according to Proposition~\ref{prop_relations_postulates_agm}, \agm{6} is fulfilled as well.
    
    \item[{[\normalfont \agm{7} is satisfied.]}]    
    We have to show that $  A \notin \beliefsOf{\forgetBy{\Psi}{A\land C}} $ implies $ \beliefsOf{\forgetBy{\Psi}{A\land C}} \subseteq \beliefsOf{\forgetBy{\Psi}{A}} $.
    \FARBE{For \agm{7}, we first conclude $\kappa^{\circ}_{A\land C}(\notA) = 0$ from $ A \not\in\beliefsOf{\kappa^{\circ}_{A\land C}} $. This yields $\kappa(\notA) = \kappa(\notA \vee \ol{C}) = \min \{\kappa(\notA), \kappa(\ol{C})\}$.
        From
        $ \beliefsOf{\forgetBy{\kappa}{A}} = Th(\{ \omega \models \ol{A} \, | \, \kappa(\omega) =  \kappa(\notA)\})$ and
        $ \beliefsOf{\forgetBy{\kappa}{A\land C}} = Th(\{ \omega \models \ol{A} \vee \ol{C} \, | \, \kappa(\omega) = \kappa(\ol{A} \vee \ol{C}) = \kappa(\ol{A})\})$
        according to \eqref{eq_bel_th},
        we conclude with Lemma \ref{Lemma_Th} that
        $ \beliefsOf{\forgetBy{\kappa}{A\land C}} \subseteq  \beliefsOf{\forgetBy{\kappa}{A}}$ holds. }
    Thus, \agm{7} is fulfilled.
    
    \item[{[\normalfont \propCF is satisfied.]}]   
    We have to show that $ \beliefsOf{\forgetBy{\kappa}{A\land C}}=\beliefsOf{\forgetBy{\kappa}{A}}$ or $ \beliefsOf{\forgetBy{\kappa}{A\land C}}=\beliefsOf{\forgetBy{\kappa}{C}} $ or $ \beliefsOf{\forgetBy{\kappa}{A\land C}}=\beliefsOf{\forgetBy{\kappa}{A}}\cap \beliefsOf{\forgetBy{\kappa}{C}} $ holds.
    The forgetting by OCF-conditionalition of $A \land C$ for a ranking function \( \kappa \) is  $\forgetBy{\kappa}{AC} = \kappa|(\notA\lor\notC)$, i.e., the conditionalization of \( \kappa \) by $(\notA\lor\notC)$.
    \FARBE{%
        \FARBE{Let $\omega$ be a minimal world in $\forgetBy{\kappa}{AC}$, i.e. \( \omega \models \ol{A} \lor \ol{C} \) and $\forgetBy{\kappa}{AC}(\omega) = 0$. 
        Because of $\kappa|(\notA\lor\notC)(\omega) = \kappa(\omega) - \kappa(\notA\lor\notC)$, we have $\kappa(\omega) = \kappa(\notA\lor\notC) = \min\{\kappa(\notA), \kappa(\notC)\}$. Hence, $\kappa(\omega) = \kappa(\notA)$ or $\kappa(\omega) = \kappa(\notC)$ for all such $\omega$.
        Note that for \( \omega\models\ol{A} \) we have 
        	\( \kappa(\omega) = \kappa(\notA) \text{ iff } \forgetBy{\kappa}{A}(\omega) = \kappa|\notA(\omega) = \kappa(\omega) - \kappa(\notA) = 0 \).
         Analogously, for \( \omega\models\ol{C} \), we have $\kappa(\omega) = \kappa(\notC)$ iff $\forgetBy{\kappa}{C}(\omega) = \kappa|\notC(\omega) = \kappa(\omega) - \kappa(\notC) = 0$.}
        Therefore, $\omega$ would also be minimal in $\forgetBy{\kappa}{A}$ or in $\forgetBy{\kappa}{C}$, respectively, depending on the value of $\kappa(\notA\lor\notC)$.
        We can now distinguish three cases for these minimal worlds: Either $\kappa(\notA) < \kappa(\notC)$ and $(\forgetBy{\kappa}{AC})^{-1}(0) = (\forgetBy{\kappa}{A})^{-1}(0)$, or $\kappa(\notC) < \kappa(\notA)$ and $(\forgetBy{\kappa}{AC})^{-1}(0) = (\forgetBy{\kappa}{C})^{-1}(0)$, or $\kappa(\notA) = \kappa(\notC)$ and $(\forgetBy{\kappa}{AC})^{-1}(0) = (\forgetBy{\kappa}{A})^{-1}(0) \cup (\forgetBy{\kappa}{C})^{-1}(0)$.
        According to Proposition~\ref{prop:agm-postulates-ocf-characterization}, it follows that \propCF\ is fulfilled.\qedhere
    }
\end{description}
\end{proof}

\OcfCond fulfills almost all AGMes postulates.
Only \agm{1} cannot be fulfilled because it is possible that some of the most plausible \FARBE{worlds} do not refer to the context after forgetting and hence are removed during \ocfCond.

\begin{proposition}\label{prop:conditionalization_ASPPostulates}
	\FARBE{Forgetting by OCF-conditionalization} fulfills \propE, \propOI, \FARBE{\propDE, and \propBE,} but does not fulfill \propW, \propWCSigma, \propSCSigma, \propCPSigma, \propWE, and \propBP.
\end{proposition}

\begin{figure}[tb]%
    \centering%
    \begin{ocftable}{3}
        \ocf{$\kappa$}{
            {${a}\overline{b}{c}$, $\overline{a}\overline{b}{c}$, $\overline{a}\overline{b}\overline{c}$},
            {${a}\overline{b}\overline{c}$},
            {${a}{b}{c}$, ${a}{b}\overline{c}$, $\overline{a}{b}{c}$, $\overline{a}{b}\overline{c}$}
        }
        \ocf{$\forgetBy{\kappa}{b}=\kappa|\overline{b}$}{
            {},
            {${a}\overline{b}{c}$, $\overline{a}\overline{b}{c}$, $\overline{a}\overline{b}\overline{c}$},
            {${a}\overline{b}\overline{c}$}
        }
    \end{ocftable}
	\caption{Counterexample for \propW for \FARBE{forgetting by OCF-conditionalization}}
	\label{tab:ce_w_cond}
\end{figure}

\begin{figure}[tb]%
    \centering%
    \begin{ocftable}{3}
        \ocf{$\kappa$}{
            {$\overline{a}\overline{b}$},
            {${a}{b}$, ${a}\overline{b}$},
            {$\overline{a}{b}$}
        }
        \ocf{$\forgetBy{\kappa}{\overline{a}} = \kappa|a$}{
            {},
            {},
            {${a}{b}$, ${a}\overline{b}$}
        }
    \end{ocftable}
    \caption{\FARBE{Counterexample for \FARBE{\propWCSigma\ and \propSCSigma} for forgetting by OCF-conditionalization}}
    \label{tab:ce_wc-sc_cond}
\end{figure}

\begin{figure}[tb]%
	\centering%
    \begin{ocftable}{3}
        \ocf{$\kappa_1$}{
            {${a}\overline{b}$},
            {${a}{b}$},
            {$\overline{a}{b}$, $\overline{a}\overline{b}$}
        }
        \ocf{$\kappa_2$}{
            {${a}{b}$},
            {${a}\overline{b}$},
            {$\overline{a}{b}$, $\overline{a}\overline{b}$}
        }
        \ocf{$\forgetBy{(\kappa_1)}{\overline{a}} = \kappa_1|a$}{
            {},
            {${a}\overline{b}$},
            {${a}{b}$}
        }
        \ocf{$\forgetBy{(\kappa_2)}{\overline{a}} = \kappa_2|a$}{
            {},
            {${a}{b}$},
            {${a}\overline{b}$}
        }
    \end{ocftable}
	\caption{Counterexample for \propWE for \FARBE{forgetting by OCF-conditionalization}}
	\label{tab:ce_we_cond}
\end{figure}

\begin{figure}%
	\centering%
	\begin{ocftable}{3}
		\ocf{$\kappa$}{
			{$a \notb,\:\nota b$},
			{$\nota \notb$},
			{$ab$}
		}
		\ocf{$\forgetBy{\kappa}{b} = \kappa | \notb$}{
			{},
			{$a \notb$},
			{$\nota \notb$},
		}
	\end{ocftable}
	\caption{Counterxample for \propBP for forgetting by OCF-conditionalization.}
	\label{fig:ex_BP_cond}
\end{figure}

\begin{proof}
\FARBE{\textbf{Part I:}}	For ease of reading, we show
\FARBE{the properties
\propE\ and \propOI}
to hold for general conditionalizations $\kappa|A$, \FARBE{implying directly that they also
hold for forgetting by OCF-conditionalization (cf. Equations~\eqref{eq_def_condionalization} and \eqref{eq_forgetting_by_OCF_condionalization}).}

\FARBE{For    \propE, \FARBE{we
show that}
$\kappa_1 \cong \kappa_2$  implies $\kappa_1|A \cong \kappa_2|A$.
Due to Proposition~\ref{prop:inf_equiv},
$\kappa_1 \cong \kappa_2$ implies}
that for all $\omega_1,\omega_2 \in \Omega$ we have $\kappa_1(\omega_1) \leq \kappa_1(\omega_2)$ iff $\kappa_2(\omega_1) \leq \kappa_2(\omega_2)$
\FARBE{and thus in particular}
$\kappa_1(\omega^\prime_1) \leq \kappa_1(\omega^\prime_2)$ iff $\kappa_2(\omega^\prime_1) \leq \kappa_2(\omega^\prime_2)$ for all $\omega^\prime_1,\omega^\prime_2 \in \Mod(A)$.
\FARBE{Subtracting the same value on both sides of an inequation yields that,}
for all $\omega^\prime_1,\omega^\prime_2 \in \Mod(A)$, it is the case that $\kappa_1(\omega^\prime_1) -\kappa_1(A) \leq \kappa_1(\omega^\prime_2) -\kappa_1(A)$ iff $\kappa_2(\omega^\prime_1) -\kappa_2(A) \leq \kappa_2(\omega^\prime_2) -\kappa_2(A)$.
\FARBE{Thus,}
for $\omega^\prime_1,\omega^\prime_2 \in \Mod(A)$ we get $\kappa_1|A(\omega^\prime_1) \leq \kappa_1|A(\omega^\prime_2)$ iff $\kappa_2|A(\omega^\prime_1) \leq \kappa_2|A(\omega^\prime_2)$,
        and hence to $\kappa_1|A \cong \kappa_2|A$
\FARBE{according to Proposition~\ref{prop:inf_equiv}.}        
	
	\FARBE{For \propOI, \FARBE{we
show} $(\kappa|A)|B = (\kappa|B)|A$.
Let} $\kappa^\prime$ be the ranking function that is obtained by the conditionalization of $\kappa $ by $A$, thus \FARBE{for $\omega \models A$, we have} $\kappa^\prime (\omega) = \kappa|A (\omega) = \kappa(\omega) - \kappa(A)$.
	By conditioning $\kappa^\prime$ with $B$ we get $(\kappa|A)|B(\omega) = \kappa^\prime(\omega) - \kappa^\prime(B) = \kappa(\omega) - \kappa(A) - \kappa^\prime(B)$ for models of $AB$.
	With $\kappa^{\prime\prime}=\kappa|B$ we have $(\kappa|B)|A(\omega) = \kappa^{\prime\prime}(\omega) - \kappa^{\prime\prime}(A) = \kappa(\omega) - \kappa(B) - \kappa^{\prime\prime}(A)$ for models of $AB$.
	Since $\kappa^\prime$ is a ranking function over the models of $A$
\FARBE{we have}
$\kappa^\prime(B)=\min_{\{ \omega\in\modelsOf{A} \mid \omega\models B \}} \kappa(\omega) - \kappa(A) = \kappa(AB) - \kappa(A)$.
	This leads to $(\kappa|A)|B(\omega) = \kappa(\omega) - \kappa(A) - \kappa^\prime(B) = \kappa(\omega) - \kappa(A) - \kappa(AB) + \kappa(A) = \kappa(\omega) - \kappa(AB)$ for models of $AB$.
	With an analogous
\FARBE{derivation,} %
we get $\kappa^{\prime\prime}(A) = \kappa(AB) - \kappa(B)$ and $(\kappa|B)|A(\omega) = \kappa(\omega) - \kappa(AB)$ for models of $AB$.
	Thus, we get $(\kappa|A)|B(\omega)=(\kappa|B)|A(\omega)$ for all $\omega \models AB$, \FARBE{and hence $(\kappa|A)|B = (\kappa|B)|A$ as required.}
	
	For \propDE, we consider two OCFs $\kappa_1, \kappa_2$ with $\kappa_1 \cong \kappa_2$ and two propositions $A,C$ with $A \equiv C$. We have to show that \FARBE{$\beliefsOf{\forgetBy{(\kappa_1)}{A}} = \beliefsOf{\forgetBy{(\kappa_2)}{C}}$,} which means showing that $\kappa_1|\notA(\omega) = 0$ iff $\kappa_2|\notC(\omega) = 0$ for all $\omega$ (considering only models of $\notA \equiv \notC$). From Definition~\ref{def:conditionalization} we can derive that $\kappa_1|\notA(\omega) = 0$ implies $\kappa_1(\omega) = \kappa_1(\notA)$, meaning that $\omega$ is a minimal model of $\notA$ in $\kappa_1$. Because $\kappa_1$ and $\kappa_2$ are equivalent, there cannot be a more plausible model of $\notA \equiv \notC$ in $\kappa_2$ than $\omega$. Therefore, $\kappa_2(\omega) = \kappa_2(\notC)$ and $\kappa_2|\notC(\omega) = 0$ must hold as well. Hence, $\kappa_1|\notA(\omega) = 0$ implies $\kappa_2|\notC(\omega) = 0$. The implication in the other direction can be shown analogously.
		
	According to Proposition~\ref{prop_relations_postulates_asp}, \propDE implies \propBE. Therefore, \propBE
\FARBE{holds} %
as well.

\smallskip

\noindent
\FARBE{\textbf{Part II:}
We now show violation of the other postuates.}

\FARBE{For \propW, let}
	$\kappa$ be a ranking function as shown in Figure~\ref{tab:ce_w_cond}.
	The forgetting of $b$ leads to the conditionalization of $\kappa$ to $\overline{b}$, $\forgetBy{\kappa}{{b}}=\kappa|\overline{b}$, and to a removal of all the most plausible worlds.
	As a result we get $(\overline{c}|a) \in \Cons(\forgetBy{\kappa}{{b}})$ but $(\overline{c}|a) \not\in \Cons(\kappa)$ so that \propW is not fulfilled.

\FARBE{For  \propWCSigma\ or \propSCSigma, let}
	$\kappa$ be a ranking function over $\Sigma=\{a,b\}$ with $\kappa(\overline{a}{b})=0$, $\kappa({a}\overline{b})=\kappa({a}{b}) = 1$ and $\kappa(\overline{a}\,\overline{b})=2$
\FARBE{as shown in Figure~\ref{tab:ce_wc-sc_cond}.}
        By conditionalizing $\kappa$ to $a$ we get $\kappa|a = \forgetBy{\kappa}{\overline{a}}$ and $\forgetBy{\kappa}{\overline{a}}(ab) = \forgetBy{\kappa}{\overline{a}}(a\overline{b}) = 0$.
	We have $\Mod_{\{a,b\}}(\margsub{\beliefsOf{\kappa}}{\{b\}}) = \Mod_{\{a,b\}}(\margsub{\Cn(\ol{a}b)}{\{b\}}) = \{ab, \ol{a}b\}$, but after the conditionalization the most plausible worlds are $\Mod_{\{a,b\}}(\forgetBy{\kappa}{\ol{a}}) = \Mod_{\{a,b\}}(\Cn(a)) = \{ab, a \ol{b}\}$.
Clearly,
\FARBE{$\{ab, \ol{a}b\} \nsubseteq_{\{a,b\}} \{ab, a \ol{b}\}$} and
\FARBE{$\{ab, a \ol{b}\} \nsubseteq_{\{a,b\}}  \{ab, \ol{a}b\}$,}
\FARBE{and also
$\Cn(\{ab, \ol{a}b\}) \nsubseteq_{\{a,b\}} \Cn(\{ab, a \ol{b}\})$ and
$\Cn(\{ab, a \ol{b}\} \nsubseteq_{\{a,b\}}  \Cn(\{ab, \ol{a}b\})$.
Thus, because
\(\margsub{\beliefsOf{\kappa}}{\{a,b\} \setminus \{a\}} = \Cn(\{ab, \ol{a}b\}\)
and 
\(\beliefsOf{\kappa|a} = \Cn(\{ab, a\ol{b}\})\),}
neither \propWCSigma\ nor \propSCSigma\ is fulfilled.
\FARBE{According to %
 Proposition~\ref{prop_relations_postulates_asp}, \propCPSigma\ is not fulfilled either.}

\FARBE{For \propWE, consider $\kappa_1, \kappa_2$ as shown in
	Figure~\ref{tab:ce_we_cond}.
We have} $ \beliefsOf{\kappa_1}=\beliefsOf{\kappa_2}= \Cn(\ol{a})$.
	By conditionalization of $\kappa_1, \kappa_2$ \FARBE{to $a$, all worlds} of the most plausible level are deleted
\FARBE{in $\kappa_1|a = \FARBE{\forgetBy{(\kappa_1)}{\ol{a}}}$  and in  $\kappa_2|a = \FARBE{\forgetBy{(\kappa_2)}{\ol{a}}}$.}
	The worlds of the second plausibility level form the new beliefs, yielding $b \! \in \! \beliefsOf{\kappa_1|a} = \Cn(ab)$, but $b \! \not\in \! \beliefsOf{\kappa_2|a}=\Cn(a\ol{b})$.
\FARBE{Thus, forgetting by OCF-conditionalization does not fulfill \propWE.}        

\FARBE{For \propBP, consider $\kappa$ as shown in Figure~\ref{fig:ex_BP_cond}.}
We can split the signature of $\kappa$ into $\Sigma_1 \cupdisjoint \Sigma_2$ with $\Sigma_1 = \{a\}$ and $\Sigma_2 = \{b\}$ and we want to forget $b \in \cL_{\Sigma_2}$, which results in $\forgetBy{\kappa}{b} = \kappa|\notb$. If we now marginalize \FARBE{the belief sets of $\kappa$ and of} $\forgetBy{\kappa}{b}$ to $\Sigma_1 = \{a\}$, we obtain $\margsub{\beliefsOf{\kappa}}{\{a\}} = \margsub{\Cn(ab)}{\{a\}} = \Cn(a)$ and $\margsub{\beliefsOf{\forgetBy{\kappa}{b}}}{\{a\}} = \margsub{\Cn(\nota \notb)}{\{a\}} = \Cn(\nota)$, respectively, with the help of Proposition~\ref{prop:bsmargrepresentation_extended}. Clearly, $\margsub{\beliefsOf{\kappa}}{\{a\}} \neq \margsub{\beliefsOf{\forgetBy{\kappa}{b}}}{\{a\}}$, which means that \propBP is not fulfilled.
\end{proof}

\begin{proposition}
 \label{prop_cond_eval_pp}
 \FARBE{Forgetting by OCF-conditionalization does not satisfy \propPP, \propNP, nor \propEP.}
\end{proposition}
\begin{proof}
 Let $\kappa, \kappa'$ be OCFs such that \FARBE{$\signatureOf{\kappa} = \Sigma, \signatureOf{\kappa'} \subseteq \Sigma' \subseteq \Sigma$, and $\FARBE{\signatureMinOf{A}} \subseteq \Sigma \backslash \Sigma'$. %
 The} following counterexample shows that
\FARBE{forgetting by OCF-conditionalization, $\forgetBy{\kappa}{A} = \kappa|\overline{A}$
(see Equation~\eqref{eq_forgetting_by_OCF_condionalization}),}
does not satisfy \propPP, i.e., we have $\kappa \nmmodels \kappa'$, but not $\kappa|\overline{A} \nmmodels \kappa'$.

\FARBE{Let} $\Sigma = \{a,b,c\}, \Sigma' = \{b,c\}$, and $\kappa, \kappa'$ be defined as given \FARBE{by
the following table:} 
 \[
  \begin{array}{c@{\hspace{3mm}}c@{\hspace{5mm}}c@{\hspace{3mm}}c|c@{\hspace{3mm}}c||c@{\hspace{3mm}}c}
  \toprule
  \omega & \kappa(\omega) & \omega & \kappa(\omega) &  \omega' & \kappa'(\omega') & \omega & \kappa|\overline{a}(\omega)  \\ \midrule
   abc & 0  & \ol{a}bc & 6 & bc & 0 &  \ol{a}bc & 3\\
   ab\ol{c} & 1  & \ol{a}b\ol{c} & 5 & b\ol{c} & 1 & \ol{a}b\ol{c} & 2\\
   a\ol{b}c & 1  & \ol{a}\ol{b}c & 4 & \ol{b}c & 1 & \ol{a}\ol{b}c & 1\\
   a\ol{b}\ol{c} & 2  & \ol{a}\ol{b}\ol{c} & 3 & \ol{b}\ol{c} & 2 & \ol{a}\ol{b}\ol{c} & 0\\ \bottomrule
  \end{array}
 \]
\FARBE{Observe that} $\kappa' = \margsub{\kappa}{\Sigma'}$, so we clearly have $\kappa \nmmodels \kappa'$. The last two columns show the result of forgetting $a$ \FARBE{in $\kappa$}, i.e., $\kappa|\overline{a}$ where $\kappa(\ol{a})=3$, and it is easy to see that $\kappa|\overline{a} \not\nmmodels \kappa'$ holds. 

The counterexample in the next table shows that \propNP\ is not satisfied, i.e., here we have $\kappa \not\nmmodels \kappa'$, but
\FARBE{$\kappa|\overline{a} \nmmodels \kappa'$:}  %
\[
  \begin{array}{c@{\hspace{3mm}}c@{\hspace{5mm}}c@{\hspace{3mm}}c|c@{\hspace{3mm}}c||c@{\hspace{3mm}}c}
  \toprule
  \omega & \kappa(\omega) & \omega & \kappa(\omega) &  \omega' & \kappa'(\omega') & \omega & \kappa|\overline{a}(\omega)  \\ \midrule
   abc & 2  & \ol{a}bc & 3 & bc & 0 &  \ol{a}bc & 0\\
   ab\ol{c} & 1  & \ol{a}b\ol{c} & 4 & b\ol{c} & 1 & \ol{a}b\ol{c} & 1\\
   a\ol{b}c & 1  & \ol{a}\ol{b}c & 4 & \ol{b}c & 1 & \ol{a}\ol{b}c & 1\\
   a\ol{b}\ol{c} & 0  & \ol{a}\ol{b}\ol{c} & 5 & \ol{b}\ol{c} & 2 & \ol{a}\ol{b}\ol{c} & 2\\ \bottomrule
  \end{array}
 \]
\FARBE{\FARBE{According} to Proposition \ref{prop_epppnp}, also \propEP
\FARBE{is not} %
satisfied.}
\end{proof}

\FARBE{Finally, it is straightforward to check that  forgetting by  OCF-conditionalization satisfies  \emph{Linear Equivalence}:}

\FARBE{
\begin{proposition}
 \label{prop_cond_eval_lineq}
 Forgetting by OCF-conditionalization satisfies \propxE. 
\end{proposition}
}

\FARBE{
\begin{proof}
Let $ \kappa_1, \kappa_2 $ be OCFs over $ \Sigma $ with $\kappa_1 \xequiv \kappa_2$, and let $q$ be a rational number such that $\kappa_2 = q \cdot \kappa_1$.
For $A \in \cL_{\Sigma}$ we show
\(
(\kappa_2)^{\circ}_A = q \cdot (\kappa_1)^{\circ}_A
\)
for forgetting by OCF-conditionalization (cf.\ Equation~\eqref{eq_forgetting_by_OCF_condionalization}) by the following derivation,
where $\omega \in \Mod(A)$:
\begin{align*}
(\kappa_2)^{\circ}_A(\omega)
   &= \kappa_2|\overline{A} (\omega)\\
   &= \kappa_2(\omega) - \kappa_2(\overline{A})\\
   &= q \cdot \kappa_1(\omega) - q \cdot \kappa_1(\overline{A})\\
   &= q \cdot (\kappa_1(\omega) - \kappa_1(\overline{A}))\\
   &= q \cdot \kappa_1|\overline{A} (\omega)\\
   &= q \cdot (\kappa_1)^{\circ}_A(\omega) \qedhere
\end{align*}
\end{proof}
}

\section{Evaluation: \FARBE{Forgetting by Strategic  C-Contractions}}
\label{sec_evaluation_ocf_c-contractions} 

C-contractions provide a general framework for performing contractions in the sense of Definition \ref{def:contraction}, with a free parameter $\gamma$ that can be specified in various ways, according to Proposition \ref{prop_c-contr_gamma}. If this parameter is chosen quite arbitrarily, we can hardly expect any general properties to hold. Therefore, we will focus our evaluation on the different subclasses of c-contractions defined by strategies in Section \ref{sec:instantiating_subclasses}: c-ignorations, c-revocations, as well as minimal and non-minimal c-contractions. 
\FARBE{So, in the following, when we say that, e.g., c-revocations satisfy a specific postulate this implies that all involved contraction operations are defined via a selection strategy that satisfies \postRevoke\ \FARBE{(see Section~}\FARBE{\ref{sec:instantiating_subclasses})}.}
Since examples may serve as counterexamples to \FARBE{various} postulates and for several classes of c-contractions, we collect all examples for this section in \FARBE{the \hyperref[app_examples_c-contractions]{Appendix}}.

\subsection{\FARBE{Commonalities of All Subclasses}}
\label{subsec_eval_stratccontr_general}
We start with considering postulates that are satisfied by all subclasses of c-contractions enumerated above. The most basic postulate that a contraction has to satisfy is $\kappa  \contractOpSelect A \not\models A$, i.e., \agm{3}. For c-contractions, this can be verified easily.

\begin{proposition}
 \label{prop_ccontr_agmes3}
 All considered subclasses---c-revocations, c-ignorations, minimal and non-minimal c-contractions---satisfy \agm{3}. 
\end{proposition}
\begin{proof}
Let $\kappa$ be an OCF, let $A \in \cL$,  and let 
  $\contractOpSelect$   be a strategic c-contraction operation from one of the mentioned subclasses.
 According to \FARBE{Proposition~\ref{prop:agm-postulates-ocf-characterization}}, we have to show that $(\kappa \contractOpSelect A)(\ol{A}) = 0$ for $A \not\equiv \top$.
 \FARBE{Recall that, due to Proposition~\ref{prop_c-contr_gamma}, every c-contraction  $\kappa \contractOpSelect A$ is given by 
  		\begin{align*}
  		(\kappa \contractOpSelect A)(\omega) =  - \kappa(\ol{A}) + \kappa(\omega) + \begin{cases}
				\gamma & \text{if } \omega\models A\\
				0 & \text{if } \omega\models \ol{A}
			\end{cases}
		\end{align*}
with $\gamma \geq \kappa(\ol{A}) - \kappa(A)$.
}
Hence \FARBE{$(\kappa \contractOpSelect A)(\ol{A}) = \min_{\omega\models \ol{A}} (-\kappa(\notA) + \kappa(\omega)) = - \kappa(\notA) + \kappa(\notA) =0$}. 
\end{proof}

Strategies for c-contractions are also useful to ensure less trivial properties. 

\begin{proposition}
 \label{prop_c-contractions_BE}
\FARBE{All considered subclasses---}
C-revocations, c-ignorations, minimal and non-minimal c-contractions\FARBE{---satisfy} 
\propBE. 
 \end{proposition}%
\begin{proof}
  Let $\kappa_1, \kappa_2$ be two equivalent OCFs, let $A \in \cL$,  and let 
  $\contractOpSelect$   be a strategic c-contraction operation from one of the mentioned subclasses. 
  \FARBE{We make use of the characterization of \propBE\ given in \FARBE{Proposition~\ref{prop:asp-postulates-ocf-characterization}}. 
      Because c-contractions are signature-preserving, it is sufficient to}
  show that $(\kappa_1 \contractOpSelect A)(\omega) = 0$ iff $(\kappa_2 \contractOpSelect A)(\omega) = 0$. 
 
  According to Proposition \ref{prop_c-contr_gamma}, for each $i \in \{1,2\}$, $\kappa_i \contractOpSelect A$ has the form 
  		\begin{align*}
  		(\kappa_i \contractOpSelect A)(\omega) =  - \kappa_i(\ol{A}) + \kappa_i(\omega) + \begin{cases}
				\gamma_i & \text{if } \omega\models A\\
				0 & \text{if } \omega\models \ol{A}
			\end{cases}
		\end{align*}
with $\gamma_i \geq \kappa_i(\ol{A}) - \kappa_i(A)$. 
\FARBE{Consequently, $(\kappa_i \contractOpSelect A)(\omega) = 0$ is characterized as follows:
\begin{description}
    \item[\normalfont Case 1:] If $\omega \models A$, then $(\kappa_i \contractOpSelect A)(\omega) = 0$ iff $\kappa_i (\omega) = \kappa_i(\ol{A}) - \gamma_i$.
    \item[\normalfont Case 2:] If $\omega \models \ol{A}$, then $(\kappa_i \contractOpSelect A)(\omega) = 0$ iff  $\kappa_i (\omega) = \kappa_i(\ol{A})$.
    \end{description}
}
In \FARBE{the case of $\omega \models A$, i.e., Case 1}, we have $\kappa_i (\omega) \leq \kappa_i(A)$, due to $\gamma_i \geq \kappa_i(\ol{A}) - \kappa_i(A)$. 
This can happen only when $\kappa_i (\omega) = \kappa_i(A)$ and $\gamma_i = \kappa_i(\ol{A}) - \kappa_i(A)$, i.e., only when $\contractOpSelect$ is a c-ignoration. In all other cases, only Case 2 applies. 

\FARBE{Because $\kappa_1, \kappa_2$ are equivalent,}
Proposition \ref{prop_ocf_equiv_minmodel} yields $\kappa_1(\omega) = \kappa_1(A)$ iff $\kappa_2(\omega) = \kappa_2(A)$ and $\kappa_1(\omega) = \kappa_1(\ol{A})$ iff $\kappa_2(\omega) = \kappa_2(\ol{A})$.
\FARBE{We obtain that} $(\kappa_1 \contractOpSelect A)(\omega) = 0$ iff $(\kappa_2 \contractOpSelect A)(\omega) = 0$, \FARBE{which was to be proven}. 
\end{proof}

By definition, c-revocations, c-ignorations, minimal and non-minimal c-contractions all satisfy the syntax independence property \postSyntaxIndependent\ (see Definitions \ref{def_cignoration_crevocation} and \ref{def_minimal_nonminimal_ccontraction}, as well as Proposition \ref{prop_ignore_min_si}). Therefore, the following proposition is an immediate consequence of Proposition \ref{prop_c-contractions_BE} and \FARBE{Proposition~\ref{prop_relations_postulates_mixed}}:

\begin{proposition}
 \label{prop_c-contractions_DE_AGM5}
 All considered subclasses---c-revocations, c-ignorations, minimal and non-minimal c-contractions---satisfy \propDE and \agm{5}. 
\end{proposition}

Note that  it is crucial to choose selection strategies within one and the same subclass here. For instance, if we consider two equivalent OCFs $\kappa_1, \kappa_2$ and c-contractions by the same proposition $A$, but apply c-ignoration for $\kappa_1$ and c-revocation for $\kappa_2$, the resulting contractions would fail to satisfy \propBE.

\FARBE{
Finally, we consider the \propxE postulate for the different subclasses of c-contractions. According to Proposition \ref{prop_xequiv}, we just have to check whether the respective selection strategies satisfy \postxE. This can be seen immiately for c-ignorations and minimal c-contractions. For c-revocations and non-minimal c-contractions, the strategies have to be chosen correspondingly.
}
\FARBE{
\begin{proposition}
 \label{prop_xequiv_ccontr}
 C-Ignorations and minimal c-contractions satisfy  \propxE. If the respective selection strategies fulfill \postxE, then also c-revocations and non-minimal c-contractions satisfy \propxE.
\end{proposition}
}

\FARBE{
In the following subsections, we consider the remaining postulates for each subclass separately. Since the type of strategy is clear for each class, we omit the $\selectionStrategyContractProp$-symbol. We start with considering c-ignorations.

}

\subsection{\FARBE{Forgetting by } C-Ignorations}
\label{subsec:ignorance}

Ignoration is %
a special kind of contraction that postulates
explicitly undecidedness between $A$ and $\neg A$.
In the framework of ranking functions, a c-ignoration is a special kind of c-contraction \cite{BeierleKernIsbernerSauerwaldBockRagni2019KIzeitschrift} which makes at least one model of $A$ and one of $\neg A$ maximally plausible so that afterwards we believe neither \FARBE{$A$ nor $\neg A$}.
\FARBE{Technically, from (\ref{eq:short_propositional_c_contraction_cond2_gamma}) and $\gamma = \kappa(\ol{A}) - \kappa(A)$, we immediately derive that c-ignorations have the following form: 
  		\begin{align}
  		\label{eq_c-ignorations_short}
  		(\kappa - A)(\omega) =   \kappa(\omega) - \begin{cases}
				\kappa(A) & \text{if } \omega\models A\\
				\kappa(\ol{A}) & \text{if } \omega\models \ol{A}
			\end{cases} \ ,
		\end{align}
which will be useful for the following investigations. 
The next proposition summarizes which of the postulates hold for c-ignorations, and which cannot be expected in general. 
}

\begin{proposition}
	\label{prop:c-ignoration}
    \FARBE{The following statements hold:}
\begin{enumerate}
 \item \FARBE{c-Ignorations fulfill \agm{1} , \agm{3}, and \agm{5}, 
 but do not fulfill \agm{2}, \agm{4}, \agm{6}, \agm{7}, and \propCF.}
 \item c-Ignorations do not fulfill  \propW, \propWCSigma, \propSCSigma, \propCPSigma, \propWE, \propE, \propOI, \propPP, \propNP, \propBP, and \propEP.
\end{enumerate}

\end{proposition}

\begin{proof}
    \FARBE{We start with proving the statements in 1.}
\begin{enumerate}
 \item  \FARBE{As \FARBE{for} \agm{1}, according to \FARBE{Proposition~\ref{prop:agm-postulates-ocf-characterization}}, we have to show that $\kappa(\omega) = 0$ implies $(\kappa-A)(\omega) = 0$. This can be seen easily from (\ref{eq_c-ignorations_short}): If $\kappa(\omega) = 0$ and $\omega \models A$, we have $0 = \kappa(\omega) = \kappa(A)$, hence $(\kappa - A)(\omega) = 0$. Analogously, if $\kappa(\omega) = 0$ and $\omega \models \ol{A}$, we have $0 = \kappa(\omega) = \kappa(\ol{A})$, hence $(\kappa - A)(\omega) = 0$.}
     \FARBE{Satisfaction of \agm{3} and \agm{5} is given by Proposition~\ref{prop_ccontr_agmes3} and Proposition~\ref{prop_c-contractions_DE_AGM5}.}
Example~\ref{ex_agmes24_ignore_nonmin} provides a counterexample for  \agm{2} and \agm{4}, and 
Example~\ref{ex_agmes67W_ignore_nonmin} provides a counterexample to \agm{6} and \agm{7}.
\FARBE{Because \agm{6} is violated, we obtain a  violation of \propCF from Proposition~\ref{prop_relations_postulates_agm}.
}
\item Counterexamples to the postulates are provided as follows: 
\FARBE{Example~\ref{ex_W-wCsem-sCsem_ignore_min} for \propW, \propWCSigma, and \propSCSigma,}
\FARBE{
Example~\ref{ex_wE_min} for \propWE, 
 Example~\ref{ex_E_min} for \propE, 
}
Example~\ref{ex_OI_ignore_min} for \propOI, 
Example~\ref{ex_PP_ignore_min} for \propPP, 
Example~\ref{ex_NP_ignore_min} for \propNP,
Example~\ref{ex_BP_ignore_revoc_min_nonmin} for \propBP.
\FARBE{Since neither \propWCSigma\ nor \propSCSigma\ are satisfied, also \propCPSigma\ cannot be satisfied, according to Proposition~\ref{prop_relations_postulates_asp}}.
Similarly, since neither \propPP\ nor \propNP\ are satisfied, also \propEP\ cannot be satisfied, according to Proposition~\ref{prop_epppnp}.
\qedhere
\end{enumerate}
\end{proof}

\subsection{\FARBE{Forgetting by } C-Revocations}
\label{subsec:results_strat_crevocations}

\FARBE{Revocations are complementary to ignorations, \FARBE{as} they enforce that $\neg A$ is believed after forgetting $A$ (see Section \ref{sec:kindsOfForgetting}). Strategic c-revocations provide an elegant way to implement revocations that are based on the general schema of c-contractions. According to Proposition \ref{prop_c-contr_gamma} and Definition \ref{def_cignoration_crevocation}, c-revocations have the form 
		\begin{align}\label{eq_crevocation_gamma}
			(\contract{\kappa}{A})(\omega) =  - \kappa(\ol{A}) + \kappa(\omega) + \begin{cases}
				\gamma & \text{if } \omega\models A\\
				0 & \text{if } \omega\models \ol{A}
			\end{cases}
		\end{align}
with $\gamma > \kappa(\ol{A}) - \kappa(A)$. 

The next proposition summarizes our results on evaluating c-revocations regarding the postulates.
}

\begin{proposition}
\label{prop_results_crevocations}
\FARBE{The following statements hold:}
\begin{enumerate}
 \item C-revocations 
 do not fulfill \agm{1},  but  fulfill \FARBE{\agm{2}}, \agm{3}, \agm{4}, \agm{5}, \agm{6}, \agm{7}, and \propCF. 
 \item  C-revocations do not satisfy \propW, \propWCSigma, \propSCSigma, \propCPSigma, \propWE, \propE, \propOI, \propPP, \propNP, \propEP, and \propBP.
\end{enumerate}
\end{proposition}

\begin{proof}
Throughout the proof, we make use of the equivalent formalizations of the postulates, as stated in \FARBE{Proposition~\ref{prop:agm-postulates-ocf-characterization}}.  \FARBE{It is crucial to observe that for c-revocations, $(\kappa - A)(\omega) = 0$ iff $\omega \models \ol{A}$ and $\kappa(\omega) = \kappa(\ol{A})$, because the case $\omega \models A$ and $\kappa(\omega) = \kappa(\ol{A}) - \gamma$ cannot occur because of $\gamma > \kappa(\ol{A}) - \kappa(A)$.}

\begin{enumerate}
 \item  Example~\ref{ex_agmes1_revoc_nonmin} is a counterexample to \agm{1}. 

 \FARBE{Satisfaction of \agm{3} and \agm{5} is given by Proposition~\ref{prop_ccontr_agmes3} and Proposition~\ref{prop_c-contractions_DE_AGM5}.}
 
 For \agm{2}, in the case $\kappa(\ol{A}) = 0$, this immediately yields $\kappa(\omega) = 0$ from $(\kappa - A)(\omega) = 0$, hence \agm{2} is fulfilled. \agm{4} is trivially satisfied because for all $\omega \models A$, $(\kappa - A)(\omega) > 0$. 
 
 For 
  \propCF, two propositions $A,C$ have to be considered.
 We have $(\kappa - AC)(\omega) = 0$ iff $\omega \models \ol{A} \vee \ol{C}$ and $\kappa(\omega) = \kappa(\ol{A} \vee \ol{C}) = \min \{\kappa(\ol{A}),\kappa(\ol{C})\} $. This is equivalent to (case 1) $\omega \models \ol{A}$ and $\kappa(\omega) =  \min \{\kappa(\ol{A}),\kappa(\ol{C})\} $, or 
 (case 2) $\omega \models \ol{C}$ and $ \kappa(\omega) =  \min \{\kappa(\ol{A}),\kappa(\ol{C})\} $.
 In case~1, we have $\kappa(\ol{A}) \leq \kappa(\omega)$, so 
 $\kappa(\omega) =  \min \{\kappa(\ol{A}),\kappa(\ol{C})\} $ means $\kappa(\omega) = \kappa(\ol{A})$. Analogously, in case~2, it must be that $\kappa(\omega) = \kappa(\ol{C})$.
\FARBE{Note that in case~1, $\omega \models \ol{A}$ and $\kappa(\omega)=\kappa(\ol{A})$ together are equivalent to $(\kappa-A)(\omega)=0$. Analogously, in case~2 we have $(\kappa-C)(\omega)=0$.
 If for all $\omega$, case~1 holds, $(\kappa-AC)(\omega) = 0$ iff $(\kappa-A)(\omega)=0$. 
 If for all $\omega$, case~2 holds, $(\kappa-AC)(\omega) = 0$ iff $(\kappa-C)(\omega)=0$. 
 If neither case holds for all $\omega$, $(\kappa-AC)(\omega)=0$ is equivalent to $(\kappa-A)(\omega)=0$ or $(\kappa-C)(\omega)=0$. This shows \propCF.}
 
 \FARBE{As \propCF ist fulfilled, it follows from Proposition~\ref{prop_relations_postulates_agm} that \agm{6} is fulfilled as well.}
 
 \FARBE{For \agm{7}, consider again two propositions $A,C$, and let $(\kappa-AC)(\ol{A})=0$. Then 
 $ \kappa(\ol{A})  = \kappa(\ol{A}\lor\ol{C}) = \min\{\kappa(\ol{A}),\kappa(\ol{C})\}$. 
 We have to show that $(\kappa-A)(\omega)=0$ implies $(\kappa-AC)(\omega)=0$.  
 Moreover, we have $(\kappa-A)(\omega)=0$ iff $\omega\models\ol{A}$ and $\kappa(\omega)=\kappa(\ol{A})$.
 In this case, because of $\omega \models \ol{A}\lor\ol{C}$ and $\kappa(\omega)=\kappa(\ol{A})=\kappa(\ol{A}\lor\ol{C})$, it must be that $(\kappa-AC)(\omega)=0$. Hence, \agm{7} holds.}

 \item Counterexamples to the postulates are provided as follows: 
  \FARBE{Example~\ref{ex_W-wCsem-sCsem_revoc_nonmin} for  \propW, \propWCSigma , and \propSCSigma , }
\FARBE{Example~\ref{ex_wE_nonmin} for  \propWE,}
 Example~\ref{ex_E_revoc} for  \propE,
 Example~\ref{ex_OI_revoc_nonmin} for  \propOI, 
 Example~\ref{ex_PP-NP_revoc_nonmin} for  \propPP\ and   \propNP,
Example~\ref{ex_BP_ignore_revoc_min_nonmin} for  \propBP. 
 
 \FARBE{Since neither \propWCSigma nor \propSCSigma are satisfied, also  \propCPSigma cannot be satisfied, according to Proposition~\ref{prop_relations_postulates_asp}}. 
Similarly, since neither \propPP\ nor \propNP\ are satisfied, also \propEP\ cannot be satisfied, according to  Proposition \ref{prop_epppnp}.
 \end{enumerate}
\end{proof}

\subsection{Forgetting by Minimal C-Contractions}
\label{subsec:minimal_c_contraction}

According to Proposition \ref{prop_c-contr_gamma} and Definition \ref{def_minimal_nonminimal_ccontraction}, minimal c-contractions have the form 
		\begin{align}\label{eq_cmin_gamma}
			(\contract{\kappa}{A})(\omega) =   \kappa(\omega) - \begin{cases}
				0 & \text{if } \omega\models A\\
				 \kappa(\ol{A})  & \text{if } \omega\models \ol{A}
			\end{cases}
		\end{align}
		
The next proposition summarizes our findings on minimal c-contractions. 

\begin{proposition}\label{prop:min-c-contraction}
\label{prop:properties_minimal_alpha_contraction}
\FARBE{The following statements hold:}
\begin{enumerate}
 \item Minimal c-contractions fulfill \agm{1} to \agm{7} and \propCF.
 \item 	Minimal 
		c-contractions violate \propW, \propWCSigma, \propSCSigma, \propCPSigma, \propWE, \propE, \propOI, \propPP, \propNP, \propBP, \propEP.
\end{enumerate}
\end{proposition}

\begin{proof}
\begin{enumerate}
 \item The proof %
 for \agm{1} to \agm{7}
 can be derived from the results given in \cite[Thm. 8]{KernIsbernerBockSauerwaldBeierle2017},
 and this implies also \propCF, due to Proposition~\ref{prop_relations_postulates_agm}.
\item Counterexamples to the postulates are provided as follows:
Example~\ref{ex_W-wCsem-sCsem_ignore_min} for \propW, \propWCSigma, \propSCSigma;
\FARBE{
Example~\ref{ex_wE_min} for \propWE;
Example~\ref{ex_E_min} for \propE;
}
Example~\ref{ex_OI_ignore_min} for \propOI;
Example~\ref{ex_PP_ignore_min} for \propPP; 
Example~\ref{ex_NP_ignore_min} for \propNP; 
Example~\ref{ex_BP_ignore_revoc_min_nonmin} for \propBP.

\FARBE{Since neither \propWCSigma\ nor \propSCSigma\ are satisfied, also \propCPSigma\ cannot be satisfied, according to Proposition~\ref{prop_relations_postulates_asp}}. 
Similarly, since neither \propPP\ nor \propNP\ are satisfied, also \propEP\ cannot be satisfied, according to Proposition \ref{prop_epppnp}.
\end{enumerate}

	\end{proof}
 
\subsection{Forgetting by Non-Minimal C-Contractions}
\label{subsec:nonminimal_c_contraction}

Non-minimal c-contractions are strategic c-contractions where the impact parameter~$\gamma$ chosen by the selection strategy always differs from $\kappa(\ol{A})$, i.e. $\gamma = \selectionStrategyContractProp(\kappa, A) \neq \kappa(\ol{A})$ for every OCF $\kappa$ and every proposition $A$.
According to Proposition \ref{prop_c-contr_gamma} and Definition \ref{def_minimal_nonminimal_ccontraction}, non-minimal c-contractions have the form 
\begin{align}\label{eq:nonmin-c-contraction-gamma}
	(\contract{\kappa}{A})(\omega) =  - \kappa(\ol{A}) + \kappa(\omega) + \begin{cases}
		\gamma & \text{if } \omega\models A\\
		0 & \text{if } \omega\models \ol{A}
	\end{cases}
\end{align}
with $\gamma \geq \kappa(\ol{A}) - \kappa(A)$ and $\gamma \neq \kappa(\ol{A})$. 

The next proposition summarizes our results on evaluating non-minimal c-contractions regarding the postulates.

\begin{proposition}\label{prop:nonmin-c-contraction-postulates}
\FARBE{The following statements hold:}
	\begin{enumerate}
		\item Non-minimal c-contractions do not fulfill \agm{1}, \agm{2}, \agm{4}, \agm{6}, \agm{7}, and \propCF.
		\item Non-minimal c-contractions do not fulfill \propW, \propWCSigma, \propSCSigma, \propCPSigma, \propWE, \propE, \propOI, \propPP, \propNP, \propBP, and \propEP.
	\end{enumerate}
\end{proposition}

\begin{proof}
	Counterexamples to the postulates are provided as follows:		
	\begin{enumerate}
		\item%
		Example~\ref{ex_agmes1_revoc_nonmin} for \agm{1};
		Example~\ref{ex_agmes24_ignore_nonmin} for \agm{2} and \agm{4};
		Example~\ref{ex_agmes67W_ignore_nonmin} for \agm{6} and \agm{7}.
		
		As \agm{6} is not satisfied, \propCF cannot be satisfied according to Proposition~\ref{prop_relations_postulates_agm}.
		\item%
		\FARBE{Example~\ref{ex_W-wCsem-sCsem_revoc_nonmin} for \propW, \propWCSigma, and \propSCSigma;}
		Example~\ref{ex_wE_nonmin} for \propWE;
		\FARBE{Example~\ref{ex_E_revoc} for \propE;}
		Example~\ref{ex_OI_revoc_nonmin} for \propOI;
		Example~\ref{ex_PP-NP_revoc_nonmin} for \propPP and \propNP;
		Example~\ref{ex_BP_ignore_revoc_min_nonmin} for \propBP.
		
		\FARBE{Since neither \propWCSigma\ nor \propSCSigma\ are satisfied, also \propCPSigma\ cannot be satisfied, according to Proposition~\ref{prop_relations_postulates_asp}}. 
		Similarly, since neither \propPP nor \propNP are satisfied, also \propEP cannot be satisfied, according to Proposition~\ref{prop_epppnp}. \qedhere
	\end{enumerate}
\end{proof}	

\section{Related Work and Discussion}
\label{sec:related_work}

\FARBE{The term \emph{forgetting} has been used in a large variety of different approaches in knowledge representation and beyond~
    \FARBE{\cite{Delgrande17,KS_FangLiuDitmarsch2016,KS_EiterWang2008,KS_Delgrande2017,KS_ErdemFerraris2007,KS_PavlikAnderson2005,KS_RajaratnamLevesquePagnuccoThielscher2014,KS_Sauerwald2019,KS_SuSattarLvZhang2014,KS_DitmarschHerzigLangMarquis2009}.}  
Conceptually, there is no common \FARBE{understanding} on what forgetting exactly is in the context of knowledge and epistemic states~\cite{KernIsbernerEiter2018}.
The \FARBE{closest and most relevant} work to this article is the work by Delgrande on forgetting~\cite{Delgrande17}, which proposes a \FARBE{technical} notion of forgetting and discusses on the nature of forgetting.

Delgrande provides \FARBE{``a knowledge level account of forgetting''}, whereby forgetting is treated as belief change operator that amounts to a reduction in the language~\cite{Delgrande17}.
The approach has been deeply developed in \FARBE{that  and many other papers (see, e.g.,~\cite{KS_LinReiter1994}),} and is rooted in \emph{forgetting a fact \( p \)} by Boole~\cite{KS_Boole1854}. Technically, forgetting an atomic fact \( p \) in a formula \( \psi \) amounts to
\begin{equation*}
	\mathit{forget}(\psi,p) = \psi[p/\top] \lor \psi[p/\bot]
\end{equation*}
whereby \( \psi[p/t] \) denotes the result of simultaneously substituting all occurrences of \( p \) in \( \psi \) by \( t\in\{\top,\bot\} \).
Delgrande extends the approach of forgetting an atom \( p \) in a formula \( \psi \) to forgetting \( p \) in arbitrary sets of formulas \( \Gamma \) over some signature \FARBE{\( \Sigma \)}:
\begin{equation*}
	\mathit{forget}(\Gamma,p) = \Cn_{\FARBE{\Sigma}}(\Gamma)\cap \mathcal{L}_{\Sigma\setminus\{p\}}
\end{equation*}
\FARBE{Semantically, }\FARBE{this corresponds to the operation of marginalization (see Section~\ref{sec:bg_marginlisation}).}
Delgrande provides a larger discussion on forgetting in comparison to other operations, including some of the operations, e.g., contraction, which we also consider here.
His major point is
that forgetting should stand purely for signature-reducing operations; other kinds of belief change operators should be considered as a different kind of operation~\cite{Delgrande17}.

\FARBE{While we basically agree that signature-reducing operations differ significantly from AGM belief change operations, we argue in this article that  signature-reducing operations are not enough to capture all facets of forgetting, and that ideas from AGM belief change theory enrich the landscape of forgetting, both conceptually and technically. Hence,}
\FARBE{we advocate here a \FARBE{broader} view on forgetting within the context of epistemic states.
One rationale is that the axiomatization of forgetting by Delgrande~\cite{Delgrande17} in an epistemic context, might lead to triviality results in this context~\cite{SauerwaldKernIsbernerBeckerBeierle2022SUM}.
However, this problem might be solvable via a reinterpretation of \FARBE{Delgrande's} postulates within an epistemic context.
In common-sense reasoning, in the tradition of knowledge representation and reasoning, and even in artificial intelligence as an intellectual endeavour, 
the notion of \emph{forgetting} should aim at
providing conceptualizations and implementations of human-like abilities.
Delgrande's proposal to denote only signature-reducing operations as forgetting seems to 
\FARBE{be too technical}
to capture what forgetting is from the perspective of, e.\,g., cognition or common sense.

This article acts as a witness for the variety of forgetting by showing that there are several substantially different forgetting operators, each implementing a certain kind of forgetting with different characteristics (cf. Figure~\ref{fig_evaluation_overview}).
The results in this article support Delgrande's view~\cite{Delgrande17} that signature-reducing operations on the one hand and other kinds of belief change operations on the other hand form groups of operations on their own.
In Figure~\ref{fig_evaluation_overview}, we see a divide in the satisfaction of postulates by language-reducing operations, like marginalization and conditionalization, and by other kinds of operations, like forms of contraction.
While the different viewpoints on forgetting are reflected in the typical postulates, some of the  ASP-inspired postulates are satisfied by belief-change inspired operators;
on the other side, language-reducing operations also satisfy some of the belief-change  inspired postulates.
It is a strength of the approach developed in this article that all kinds of operators can be studied in one common framework.
}

}

\section{Conclusions and Future Work}
\label{sec:conclusion}
\FARBE{
In this article,  \FARBE{we} presented a unifying epistemic framework for different kinds of intentional forgetting in epistemic states. Five general types of forgetting---\emph{Contraction}, \emph{Ignoration}, \emph{Revocation}, \emph{Marginalization}, and \emph{Conditionalization}---were \FARBE{conceptually} characterized and instantiated by seven concrete forgetting operations for Spohn’s ranking functions.
Our study imports postulates from logic programming and belief change theory into our framework.
By adapting existing postulates, introducing novel persistence and equivalence postulates, and evaluating all operators against this comprehensive set, we obtained a systematic overview of differences and commonalities between forgetting paradigms.

The evaluation confirms and sharpens known contrasts between logic-programming-inspired forgetting and AGM-style contraction, while revealing bridging points such as \emph{Belief Equivalence}~\propBE, satisfied by all considered operations. The notion of linear equivalence for ranking functions further demonstrates how the arithmetic structure of OCFs can preserve strong equivalence properties in the epistemic setting.

There are several obvious directions for future work: (i) extending the framework to more complex knowledge items, such as conditionals and rules; (ii) developing efficient algorithms for the proposed operations, with attention to scalability; and (iii) applying the framework in dynamic and multi-agent contexts, where principled forgetting supports relevance management, complexity control, and adaptability.%
}
\FARBE{Our future work also includes investigating the relationships of forgetting based
on marginalization and conditionalization to splitting techniques used in knowledge
representation and reasoning
\cite{Parikh99,Kern-IsbernerB17,KernIsbernerBeierleBrewka2020KR,HeyninckKernIsbernerMeyerHaldimannBeierle2023AAAI,KernIsbernerSezginBeierle2023AIJ,BeierleSpiegelHaldimannWilhelmHeyninckKernIsberner2024KR}, and how and whether these approaches can
profit from insights and results gained in the context of forgetting.}
\FARBE{Furthermore, we will continue our investigations to additionally consider non-intentional forgetting operators~\cite{BeierleKernIsbernerSauerwaldBockRagni2019KIzeitschrift,KS_FermeHerzigMartinez2025,KS_SauerwaldBeierle2019}.}

\medskip
\pagebreak[3]
\noindent\textbf{Acknowledgements}\
We thank Tanja Bock for her contributions to the conference papers \cite{KernIsbernerBockBeierleSauerwald2019a,KernIsbernerBockSauerwaldBeierle2019} where the first ideas for this line of research were presented and for her comments
on early drafts of this article.
The research reported here was supported by the Deutsche Forschungsgemeinschaft (DFG, German Research Foundation),
project 512363537,
grant BE-1700/12-1 awarded to Christoph Beierle and
grant KE 1413/15-1 awarded to Gabriele Kern-Isberner.
Kai Sauerwald was partially supported
by the Deutsche Forschungsgemeinschaft (DFG, German Research Foundation) project 465447331 awarded to Matthias Thimm.

%
\bibliographystyle{elsarticle-num} 
\addcontentsline{toc}{section}{Bibliography}
\bibliography{references.bib}

\clearpage
\appendix
\renewcommand{\thesection}{\Alph{section}}
\makeatletter
\renewcommand\@seccntformat[1]{%
    \appendixname\@seccntDot\hskip.5em%
}
\makeatother
\label{subsec:contraction}

\section{Counterexamples Used in Proofs in Section~\ref{sec_evaluation_ocf_c-contractions}}
\label{app_examples_c-contractions}
This appendix presents examples that serve as counterexamples relevant for various postulates and for several classes of c-contractions. These counterexamples are mainly used in proofs in Section~\ref{sec_evaluation_ocf_c-contractions}.

\begin{example}[\agm{1}; revocation, non-minimal]
\label{ex_agmes1_revoc_nonmin}

C-revocations and non-minimal c-contractions do not fulfill \agm{1}, as  we show by a counterexample. 
Let $\kappa$ be a ranking function over the signature $\Sigma=\{a,b\}$ as given in \autoref{tab:ce_agm1_rev_non-min}, and we c-contract $\kappa$ by $A = a$ according to Proposition \ref{prop_c-contr_gamma}, choosing $\gamma=2$. Since $\kappa(a) = 0$ and $\kappa(\ol{a}) = 1 < 2 = \gamma$, this contraction is both a c-revocation and a non-minimal c-contraction (cf.\ Section \ref{sec:instantiating_subclasses}). The contraction result is also shown in \autoref{tab:ce_agm1_rev_non-min}, and we find that $\kappa(ab) = 0$, but $\contract{\kappa}{a}(ab) = 1$. Hence, by \FARBE{Proposition~\ref{prop:agm-postulates-ocf-characterization}}, this c-contraction cannot fulfil \agm{1}. 

\end{example}

\begin{example}[\agm{2}, \agm{4}; ignoration, non-minimal]
\label{ex_agmes24_ignore_nonmin}
C-igno\-ra\-tions and non-minimal c-contractions do not fulfill \agm{2} and \agm{4}, as  we show by a counterexample. 
Let $\kappa$ be a ranking function over the signature $\Sigma=\{a,b\}$ as given in \autoref{tab:ce_con_2_4}, and we c-contract $\kappa$ by $A = a$ according to Proposition \ref{prop_c-contr_gamma}, choosing $\gamma=(-1)$. Since $\kappa(a) = 1$ and $\kappa(\ol{a}) = 0$, this contraction is both an ignoration ($\kappa(\ol{a}) - \kappa(a) = -1 = \gamma$) and a non-minimal c-contraction ($\kappa(\ol{a}) = 0 \neq -1 = \gamma$) (cf.\ Section \ref{sec:instantiating_subclasses}).
\FARBE{The contraction result is also shown in  \autoref{tab:ce_con_2_4}, and we find that $(\contract{\kappa}{a})(ab) = 0$, but $\kappa(ab) = 1$
. Hence, by \FARBE{Proposition~\ref{prop:agm-postulates-ocf-characterization}}, this c-contraction cannot fulfil \agm{2}.
Regarding \agm{4}, since for all $\omega \models a$ we have that $\kappa(\omega) > 0$, but $(\contract{\kappa}{a})(ab) = 0$, \agm{4} cannot be fulfilled due to \FARBE{Proposition~\ref{prop:agm-postulates-ocf-characterization}}.}
\end{example}

\begin{figure}
\FARBE{
	\centering
    \begin{ocftable}{3}
   		\ocf{$\kappa$}{
   		    {$\overline{a} b$},
   			{${a} \overline{b},\:\overline{a} \overline{b}$},
   			{$a b$}
   		}	
   		\ocf{$\kappa - a$}{
   			{${a} \overline{b}$},
   			{$a {b},\:\overline{a} {b}$},
   			{$\overline{a} \overline{b}$}
      	}	
        \end{ocftable} 
    	 \caption{\FARBE{Counterexample to \agm{1} for c-revocations and non-minimal c-contractions for Example~\ref{ex_agmes1_revoc_nonmin}}}
        \label{tab:ce_agm1_rev_non-min}
}
\end{figure}

\begin{figure}
\FARBE{
	\centering
   		\begin{ocftable}{3}
      		\ocf{$\kappa$}{
      			{${a} \overline{b}$},
       			{$a {b},\:\overline{a} b$},
       			{$\overline{a} \overline{b}$}
      		}
       		\ocf{$\kappa - a$}{
       			{},
       			{$a \overline{b},\:\overline{a} {b}$},
       			{${a} b,\:\overline{a} \overline{b}$}
      		}
   		\end{ocftable}
	\caption{\FARBE{Counterexample to \agm{2} and \agm{4} for c-ignorations and non-minimal c-contractions for Example~\ref{ex_agmes24_ignore_nonmin}}}
    \label{tab:ce_con_2_4}
}      
\end{figure}

\begin{example}[\propW, \propWCSigma, \propSCSigma; revocation, non-minimal]
\label{ex_W-wCsem-sCsem_revoc_nonmin}
Let $\kappa$ be a ranking function over $\Sigma=\{a,b\}$ with $\kappa(ab)=0$ and $\kappa(a\ol{b}) = \kappa(\ol{a}b) = \kappa(\ol{a}\ol{b}) = 1$. We c-contract $\kappa$ by $A=a$ according to Proposition \ref{prop_c-contr_gamma}, choosing $\gamma=2$. Since $\kappa(a)=0$ and $\kappa(\ol{a})=1 < 2=\gamma$, this contraction is both a c-revocation and a non-minimal c-contraction (cf. Section \ref{sec:instantiating_subclasses}). By contracting $\kappa$ with $A$ we get $\kappa-a(\ol{a}b)=\kappa-a(\ol{a}\ol{b}) = 0$, $\kappa-a(ab)=1$ and $\kappa-a(a\ol{b})=2$. \\
This leads to $(\ol{a}|b)\in\Cons(\kappa-a)$ but $(\ol{a}|b)\not\in\Cons(\kappa)$, so the c-contraction cannot fulfill \propW. \\
For \propSCSigma, we choose $\omega=ab$ since $\kappa(\omega^{\Sigma\setminus\signatureMinOf{A}}) = \kappa(b) = 0$. According to \FARBE{Proposition~\ref{prop:asp-postulates-ocf-characterization}}, we would need $(\kappa-a)(\omega) = 0$ in order to fulfill \propSCSigma. However, we have $(\kappa-a)(\omega) = 1$.\\
For \propWCSigma, we choose $\omega=\ol{a}\ol{b}$. It holds that $(\kappa-a)(\omega)=0$, but $\kappa(\omega^{\Sigma\setminus\signatureMinOf{A}}) = \kappa(\ol{b}) = 1$. So, due to \FARBE{Proposition~\ref{prop:asp-postulates-ocf-characterization}}, \propWCSigma cannot be fulfilled.
\end{example}

\begin{example}[\propOI; revocation, non-minimal]
\label{ex_OI_revoc_nonmin}
	\setlength{\parindent}{0pt}

We consider the OCF $\kappa$ over 	$\Sigma = \{a, b, c\}$, as given in Fig.~\ref{fig_OI_revoc_nonmin}. Let $A = a,\ B = b$, and we compute $(\kappa - a)- b$ vs.\ $(\kappa - b) - a$, where all contractions are c-revocations resp.\ non-minimal c-contractions. All (intermediate) results are also shown in Fig.~\ref{fig_OI_revoc_nonmin}.
Observe that $\kappa(a) = \kappa(b) = 0,\ \kappa(\overline{a}) = \kappa(\overline{b}) = 2$. 

First, we compute $(\kappa - a)- b$. For the first c-contraction $\kappa - a$, we choose $\gamma_{11} = 3 (> 2)$. We have $\kappa - a (b) = 0, \kappa - a(\ol{b}) = 3$, and we choose $\gamma_{12} = 4 (>3)$ to set up $(\kappa - a)- b$. 
For computing $(\kappa - b) - a$, we first choose $\gamma_{21} = 6 (>2)$ for $(\kappa - b)$, yielding $\kappa - b(a) = 0, \kappa - b(\ol{a}) = 6$. Then we choose $\gamma_{22} = 8 (>6)$ to obtain $(\kappa - b) - a$. 

It can be clearly seen that 
$(\kappa - a) - b  \neq  (\kappa - b) - a$, so that \propOI\ is not satisfied. Note that we even do not have equality on the level of belief sets, because $\Bel((\kappa - a) - b) = \Cn(a\overline{b}c)$, while $\Bel( (\kappa - b) - a) = \Cn(\overline{a}bc \lor \overline{a}\overline{b})$.

\begin{figure}
	\centering
	\begin{ocftable}{10}
		\ocf{$\kappa$}{
		    {},
			{$\overline{a} \overline{b} {c},\:\overline{a} \overline{b} \overline{c}$},
			{},
			{},
			{$\overline{a} {b} \overline{c}$},
			{},
			{${a} \overline{b} \overline{c}$},
			{${a} \overline{b} {c},\:\overline{a} {b} {c}$},
			{${a} {b} \overline{c}$},
			{${a} {b} {c}$}
		}
		\ocf{$\kappa - a$}{
			{},
			{},
			{},
			{$\overline{a} \overline{b} {c},\:\overline{a} \overline{b} \overline{c}$},
			{},
			{${a} \overline{b} \overline{c}$},
			{${a} \overline{b} {c},\:\overline{a} {b} \overline{c}$},
			{${a} {b} \overline{c}$},
			{${a} {b} {c}$},
			{$\overline{a} {b} {c}$}
		}
		\ocf{$(\kappa - a)-b$}{		
			{},
			{},
			{},
			{},
			{},
			{$\overline{a} b \overline{c}$},
			{${a} {b} \overline{c},\:\overline{a} \overline{b} {c},\:\overline{a} \overline{b} \overline{c}$},
			{${a} {b} {c}$},
			{${a} \overline{b} \overline{c},\:\overline{a} {b} {c}$},
			{$a \overline{b} c$}
		}
		\ocf{$\kappa - b$}{
			{$\overline{a} {b} \overline{c}$},
			{},
			{},
			{$\overline{a} {b} {c},\:\overline{a} \overline{b} c,\:\overline{a} \overline{b} \overline{c}$},
			{$a b \overline{c}$},
			{$a b c$},
			{},
			{},
			{$a \overline{b} \overline{c}$},
			{$a \overline{b} c$}
		}
		\ocf{$(\kappa - b)-a$}{
		    {},
			{},
			{$a b \overline{c}$},
			{$a b c$},
			{},
			{},
			{${a} \overline{b} \overline{c},\:\overline{a} {b} \overline{c}$},
			{${a} \overline{b} {c}$},
			{},
			{$\overline{a} b c,\:\overline{a} \overline{b} {c},\:\overline{a} \overline{b} \overline{c}$}
		}
	\end{ocftable}
		\caption{Counterexample to \propOI for c-revocations and non-minimal c-contractions for Example~\ref{ex_OI_revoc_nonmin}}
	\label{fig_OI_revoc_nonmin}
\end{figure}	

\end{example}

\begin{figure}
	\centering
	\begin{ocftable}{5}
		\ocf{$\kappa_1$}{
			{},
			{$\overline{a} \overline{b}$},
			{$\overline{a} b$},
			{$a \overline{b}$},
			{$a b$}
		}
		\ocf{$\kappa_2$}{
			{$\overline{a} \overline{b}$},
			{},
			{$\overline{a} b$},
			{$a \overline{b}$},
			{$a b$}
		}
		\ocf{$\kappa_1 - a$}{
			{},
			{},			
			{$a \overline{b}$},
			{$a b,\: \overline{a} \overline{b}$},
			{$\overline{a} b$}
		}
		\ocf{$\kappa_2 - a$}{
			{},
			{},
			{$a \overline{b},\: \overline{a} \overline{b}$},
			{$a b$},
			{$\overline{a} b$}
		}
	\end{ocftable}
	\caption{Counterexample to \propE for c-revocations for Example~\ref{ex_E_revoc}}
	\label{fig:ex_E_revoc}
\end{figure}

\begin{example}[\propE; revocation, non-minimal]
	\label{ex_E_revoc}
	Let $\kappa_1, \kappa_2$ be ranking functions as defined in Figure~\ref{fig:ex_E_revoc}. We have $\kappa_1 \cong \kappa_2$, but the c-contraction of both ranking functions with a and $\gamma = 3 > 2 = \kappa(\ol{a})$ (which is also a c-revocation as $\kappa(a) = 0$) leads to $(\kappa_1 - a) \ncong (\kappa_2 - a)$. As a counterexample, consider $\omega_1 = \ol{a}\ol{b}$ and $\omega_2 = ab$. We have $(\kappa_1 - a)(\ol{a}\ol{b}) \leq (\kappa_1 - a)(ab)$, but $(\kappa_2 - a)(\ol{a}\ol{b}) > (\kappa_2 - a)(ab)$. Therefore, according to Proposition \ref{prop:inf_equiv}, $(\kappa_1 - a)$ and $(\kappa_2 - a)$ cannot be equivalent.
\end{example}

\begin{figure}
	\centering
	\begin{ocftable}{3}
		\ocf{$\kappa$}{
			{$a\,c$},
			{$a\,\overline{c},\: \overline{a}\,c$},
			{$\overline{a}\,\overline{c}$}
		}
		\ocf{$\kappa - a$}{
			{},
			{$a\,c,\: \overline{a}\,c$},
			{$a\,\overline{c},\: \overline{a}\,\overline{c}$}
		}
		\ocf{$\kappa - c$}{		
			{},
			{$a\,c,\: a\,\overline{c}$},
			{$\overline{a}\,c,\: \overline{a}\,\overline{c}$}
		}
		\ocf{$\kappa - ac$}{
			{},
			{$a\,\overline{c},\: \overline{a}\,c$},
			{$a\,c,\: \overline{a}\,\overline{c}$}
		}
	\end{ocftable}
	\caption{Counterexample to \agm{6}, \agm{7}, and \propW for c-ignorations and non-minimal c-contractions for Example~\ref{ex_agmes67W_ignore_nonmin}}
	\label{fig:ex_agmes67W_ign}
\end{figure}

\begin{example}[\agm{6}, \agm{7}, \propW; ignoration, non-minimal]
    \label{ex_agmes67W_ignore_nonmin}
    \FARBE{
    Let $\kappa$ be a ranking function as defined in Figure~\ref{fig:ex_agmes67W_ign}. The ranking functions $(\kappa - a)$, $(\kappa - c)$ and $(\kappa - ac)$ are the results of c-ignorations of $a$, $c$, and $ac$, using $\gamma_{a} = -1 = \kappa(\ol{a}) - \kappa(a)$, $\gamma_{c} = -1 = \kappa(\ol{c}) - \kappa(c)$, and $\gamma_{ac} = -2 =  \kappa(\ol{a} \lor \ol{c}) - \kappa(ac)$ as the respective values for $\gamma$. Note that all these c-contractions are also non-minimal c-contractions.\\%
    For \agm{6}, observe that $(\kappa - ac)(ac) = 0$, but $(\kappa - a)(ac) = (\kappa-c)(ac) = 1 \neq 0$. This is a contradiction to Proposition~\ref{prop:agm-postulates-ocf-characterization} with $A=a$ and $C=c$. Hence, these c-ignorations do not fulfil \agm{6}.\\%
    For \agm{7}, we have $(\kappa - ac)(\overline{a}) = 0$. Therefore, $(\kappa - a)(\omega) = 0$ should imply $(\kappa - ac)(\omega) = 0$ according to Proposition~\ref{prop:agm-postulates-ocf-characterization} using $A=a$ and $C=c$, but this is not the case for $\omega = a \overline{c}$. Hence, \agm{7} is not fulfilled.\\%
    For \propW, note that $(\kappa - ac) \models (c|a)$, because $(\kappa - ac)(ac) = 0 < 1 = (\kappa - ac)(a\ol{c})$. However, $\kappa \not\models (c|a)$, because $\kappa(ac) = 2 > 1 = \kappa(a\ol{c})$. This contradicts Proposition~\ref{prop:asp-postulates-ocf-characterization} using $A=ac$, $B=a$, and $C=c$. Therefore, this c-ignoration does not fulfil \propW.
    }
\end{example}

\begin{example}[\propOI; ignoration, minimal]
\label{ex_OI_ignore_min}
	For \propOI consider the case where we have individual impacts for the contraction with $a$ and for the contraction with $b$ for both application orderings. For each step the impacts has been choosen so that there will be a minimal c-contraction. 
	\FARBE{	
	The impacts for the first contraction by $a$ were $\gamma_{11}=2$, and for the subsequent contraction with $b$: $\gamma_{12}=0$. 
	Since $\gamma_{11}=\kappa(\ol a) = \kappa(\ol a)-\kappa(a)$ and $\gamma_{12} = (\kappa-a)(\ol b) = (\kappa-a)(\ol b)-(\kappa-a)(b)$, both contractions are ignorations and minimal. 
	The impacts for the first contraction by $b$ were $\gamma_{22}=1$, and for the subsequent contraction with $a$: $\gamma_{21}=1$. 
	Since $\gamma_{22} = \kappa(\ol b) = \kappa(\ol b)-\kappa(b)$ and $\gamma_{21} = (\kappa-b)(\ol a)=(\kappa-b)(\ol a)-(\kappa-b)(a)$, these contractions are both ignorations and minimal.
	}
		The contraction first by $a$ and then by $b$ leads to the ranking function $(\kappa-a)-b$ shown on the left of Figure~\ref{tab:ce_oi_contr}, which is not equivalent to the ranking function $(\kappa-b)-a$ shown on the right, which is the result of first forgetting $b$ and then $a$.
\end{example}

\begin{figure}
	\centering
	\begin{ocftable}{3}
		\ocf{$\kappa$}{
		    {$\overline{a} b,\: \overline{a} \overline{b}$},
            {$a \overline{b}$},
            {$a b$}
		}
		\ocf{$\kappa - a$}{
			{},
            {$a \overline{b}$},
            {$a b,\: \overline{a} b,\: \overline{a} \overline{b}$}
		}
		\ocf{$(\kappa - a) - b$}{
			{},
            {$a \overline{b}$},
            {$a b,\: \overline{a} b,\: \overline{a} \overline{b}$}
		}
		\ocf{$\kappa - b$}{
			{$\overline{a} b$},
            {$\overline{a} \overline{b}$},
            {$a b,\: a \overline{b}$},
		}
		\ocf{$(\kappa - b) -a$}{
			{},
            {$\overline{a} {b}$},
            {$a b,\: {a} \overline{b},\: \overline{a} \overline{b}$}
		}
	\end{ocftable}
	\caption{Counterexample to \propOI for c-ignorations and minimal c-contractions for Example~\ref{ex_OI_ignore_min}}
	\label{tab:ce_oi_contr}
\end{figure}

\begin{example}[\propPP; ignoration, minimal]
\label{ex_PP_ignore_min}
For \propPP, consider the example given in Figure \ref{tab:pp_contr_ign}, where $\kappa$ is a ranking function over $\Sigma$ which describes a belief set $\psi$. $\kappa-a$ is a minimal c-contraction of $\kappa$ by $a$ and also a c-ignoration, since $\gamma=4=\kappa(\ol a)-\kappa(a) = \kappa(\ol a)$. We have to show that there exists a ranking function $\kappa'$ describing a belief set $\psi'$ with $\kappa\nmmodels\kappa'$, $\signatureOf{\psi'} \subset \Sigma'$ and $\signatureMinOf{A}\subset \Sigma\setminus\Sigma'$, but $\kappa-A\not\nmmodels\kappa'$. 
We set $\Sigma' = \{b,c\}$ and
$\kappa' = \margsub{\kappa}{\{b,c\}}$,
then clearly $\kappa \nmmodels \kappa'$, but $\kappa-a \not\nmmodels \kappa'$, because $\kappa'(bc) = 3 < 4 = \kappa'(b\ol{c})$, so $\kappa' \models (c|b)$, but $\kappa-a(bc) = 1 > 0 = \kappa-a(b\ol{c})$ and hence $\kappa-a \not\models (c|b)$. 
 
\begin{figure}
	\centering
	\begin{ocftable}{6}
		\ocf{$\kappa$}{
		    {$a b \overline{c},\:\overline{a} b c$},
			{$\overline{a} b \overline{c},\:\overline{a} \overline{b} c,\:\overline{a} \overline{b} \overline{c}$},
			{$a b c$},
			{},
			{},
			{$a \overline{b} {c},\:{a} \overline{b} \overline{c}$}
		}
		\ocf{$\kappa'$}{
		    {},
			{${b} \overline{c}$},
			{$b c$},
			{},
			{},
			{$ \overline{b} {c},\: \overline{b} \overline{c}$}
		}
		\ocf{$\kappa - a$}{
			{$a b \overline{c}$},
			{},
			{$a b c$},
			{},
			{$\overline{a} b c$},
			{$a \overline{b} {c},\:{a} \overline{b} \overline{c},\:\overline{a} b \overline{c},\:\overline{a} \overline{b} c,\:\overline{a} \overline{b} \overline{c}$}
		}
	
	\end{ocftable}
		\caption{Counterexample to \propPP for c-ignorations and minimal c-contractions for Example~\ref{ex_PP_ignore_min}}
	\label{tab:pp_contr_ign}
\end{figure}

\end{example}

\begin{example}[\propNP; ignoration, minimal]
\label{ex_NP_ignore_min}
The next example shows that also \propNP\ is not satisfied in general. Let $\kappa$ be a ranking function over $\Sigma=\{a,b,c\}$ which describes a belief set $\psi$ as given in Figure \ref{tab:ce_np_ign}. 
Again, $\kappa-a$ is a minimal c-contraction  and also a c-ignoration by $A=a$ and $\gamma=2$ ($\gamma=\kappa(\ol a)-\kappa(a) = \kappa(\ol a)$). We have to show that there exists a ranking function $\kappa'$ describing a belief set $\psi'$ with $\kappa-a\nmmodels\kappa'$, $\signatureOf{\psi'} \subset \Sigma'$ and $\signatureMinOf{A}\subset \Sigma\setminus\Sigma'$, but $\kappa\not\nmmodels\kappa'$. This time, consider $\Sigma'=\{b,c\}$ and $\kappa' = \margsub{(\kappa-a)}{\{b,c\}}$, then of course, $\kappa-a \nmmodels \kappa'$. But $\kappa \not\nmmodels \kappa'$, because $\kappa' \models (c|b)$ due to $\kappa'(bc) = 0 < 1 = \kappa'(b\ol{c})$, whereas $\kappa(bc) = 1 = \kappa(b\ol{c})$ and hence $\kappa \not\models (b|c)$.

\begin{figure}
	\centering
	\begin{ocftable}{4}
		\ocf{$\kappa$}{
		    {$\overline{a} b \overline{c}$},
			{$\overline{a} b {c},\:\overline{a} \overline{b} c,\:\overline{a} \overline{b} \overline{c}$},
			{$a b c,\: a b \overline{c}$},
			{$a \overline{b} c,\: a \overline{b} \overline{c}$}
		}
		\ocf{$\kappa'$}{
		    {},
			{},
			{$b c,\:b \overline{c}$},
			{$\overline{b} c,\:\overline{b} \overline{c}$}
		}
		\ocf{$\kappa - a$}{
			{},
			{},
			{${a} b c,\: a b \overline{c},\:\overline{a} b \overline{c}$},
			{$a \overline{b} c,\: a \overline{b} \overline{c},\:\overline{a} b c,\:\overline{a} \overline{b} c,\:\overline{a} \overline{b} \overline{c}$}
		}
	
	\end{ocftable}
		\caption{Counterexample to \propNP for c-ignorations and minimal c-contractions for Example~\ref{ex_NP_ignore_min}}
	\label{tab:ce_np_ign}
\end{figure}

\end{example}

\begin{figure}
	\centering
	\begin{ocftable}{4}
		\ocf{$\kappa$}{
			{$\overline{a} b$},
			{$\overline{a} \overline{b}$},
			{$a \overline{b}$},
			{$a b$}
		}
		\ocf{$\margsub{\kappa}{\Sigma_1}$}{
			{},
			{},
			{$\overline{b}$},
			{$b$}
		}
		\ocf{$\kappa_1$}{
			{},
			{},
			{$a \overline{b},\: \overline{a} b$},
			{$a b,\: \overline{a} \overline{b}$}
		}
		\ocf{$\margsub{\kappa_1}{\Sigma_1}$}{
			{},
			{},
			{},
			{$b,\:\overline{b}$}
		}
		\ocf{$\kappa_2$}{
			{},
			{$a \overline{b}$},
			{$a b,\: \overline{a} b$},
			{$\overline{a} \overline{b}$}
		}
		\ocf{$\margsub{\kappa_2}{\Sigma_1}$}{
			{},
			{},
			{$b$},
			{$\overline{b}$}
		}
	\end{ocftable}
	\caption{Counterexample to \propBP for ignorations, revocations, minimal and non-minimal c-contractions for Example~\ref{ex_BP_ignore_revoc_min_nonmin}}
	\label{fig:ex_BP_ignore_revoc_min_nonmin}
\end{figure}

\begin{example}[\propBP; ignoration, revocation, minimal, non-minimal]
	\label{ex_BP_ignore_revoc_min_nonmin}
	Let $\kappa$ be a ranking function over $\Sigma = \{a, b\}$ as defined in Figure~\ref{fig:ex_BP_ignore_revoc_min_nonmin}. Let $\Sigma_1 = \{b\}$ and $\Sigma_2 = \{a\}$. Using $\gamma_1 = \kappa(\ol{a}) - \kappa(a) = 2 - 0 = 2$, we obtain a c-ignoration of $a$, which is also a minimal c-contraction with $a$ and which leads to $\kappa_1$. If we use $\gamma_2 = 3 > \kappa(\ol{a}) - \kappa(a) = \kappa(\ol{a}) = 2$, we obtain a c-revocation of $a$, which is also a non-minimal c-contraction with $a$, resulting in $\kappa_2$.
	Note that $\beliefsOf{\margsub{\kappa}{\Sigma_1}} = \Cn(b)$, but $\beliefsOf{\margsub{\kappa_1}{\Sigma_1}} = \Cn(b \lor \overline{b}) = \Cn(\top) \neq \beliefsOf{\margsub{\kappa}{\Sigma_1}}$ and $\beliefsOf{\margsub{\kappa_2}{\Sigma_1}} = \Cn(\overline{b}) \neq \beliefsOf{\margsub{\kappa}{\Sigma_1}}$.
\end{example}

\begin{example}[\propW, \propWCSigma, \propSCSigma; ignoration, minimal]	
\label{ex_W-wCsem-sCsem_ignore_min}
Let $\kappa$ be a ranking function over $\Sigma = \{a,b\}$ with $\kappa(\ol a b) = 0,\, \kappa(ab)=\kappa(a\ol b) = 1$ and $\kappa(\ol a\ol b) = 2$. By c-contracting $\kappa$ with $A=b$ and $\gamma = 1$ we have a minimal c-ignoration ($\gamma = 1 = \kappa(\ol b)$) with the result: $\kappa-b(\ol ab) = \kappa-b(a\ol b) = 0$ and $\kappa-b(ab) = \kappa-b(\ol a\ol b) = 1$. \\
For \propW we have $\kappa-b \models (\ol b|a)$ so $(\ol b|a)\in C(\kappa-b)$. But $\kappa\not\models (\ol b|a)$ so $(\ol b|a) \not\in C(\kappa)$. \\
For \propSCSigma, we choose $\omega=\ol{a}\ol{b}$. Note that we have $\kappa(\omega^{\Sigma\setminus\signatureMinOf{A}}) = \kappa(\ol{a}) = 0$, but $(\kappa - b)(\ol{a}\ol{b}) = 1$. Therefore, according to \FARBE{Proposition~\ref{prop:asp-postulates-ocf-characterization}}, \propSCSigma is not fulfilled.\\
For \propWCSigma, we choose $\omega=a\ol{b}$. We have $(\kappa - b)(a\ol{b}) = 0$, but $\kappa(\omega^{\Sigma\setminus\signatureMinOf{A}}) = \kappa(a) = 1$. Therefore, according to \FARBE{Proposition~\ref{prop:asp-postulates-ocf-characterization}}, \propWCSigma is not fulfilled.
\end{example}

\begin{example}[\propWE; ignoration, minimal]
 \label{ex_wE_min}
	\FARBE{
\begin{figure}
	\centering
	\begin{ocftable}{4}
	
		\ocf{$\kappa_1$}{
		    {$a \overline{b}$},
			{$\overline{a} b$},
			{$\overline{a} \overline{b}$},
			{$a b$}
		}
		\ocf{$\kappa_2$}{
			{${a} \overline{b}$},
			{$\overline{a} \overline{b}$},
			{$\overline{a} {b}$},
			{$a b$}
		}
		\ocf{$\kappa_1 - a$}{
			{$a \overline{b}$},
			{},
			{$\overline{a} {b}$},
			{$a b,\:\overline{a} \overline{b}$}
		}
		\ocf{$\kappa_2 - a$}{
			{$a \overline{b}$},
			{},
			{$\overline{a} \overline{b}$},
			{$a b,\:\overline{a} {b}$}
		}
	\end{ocftable}
	\caption{Counterexample to \propWE for c-ignorations and minimal c-contractions for Example~\ref{ex_wE_min}}
	\label{tab:wE_min-c-con}
\end{figure}

			Let $ \kappa_1 $, $ \kappa_2 $ be ranking functions %
			as given in Figure \ref{tab:wE_min-c-con}.
			Note that $\kappa_2 $ differs from $ \kappa_1 $ only by an inversion of the ranks of $\ol{a}\ol{b}$ 		
			and $ \ol{a}b $.			
			We c-contract $\kappa_1$ and $\kappa_2$ with $A=a$ and $\gamma=1$. Since $\gamma = \kappa_1(\ol a)-\kappa_1(a) = \kappa_2(\ol a)-\kappa_2(a)$ and $\gamma=\kappa_1(\ol a)=\kappa_2 (\ol a)$ both c-contractions are ignorations and minimal. The results of the c-contractions are also shown in Figure \ref{tab:wE_min-c-con}. Observe that $\kappa_1$ and $\kappa_2$ have the same lowermost layer, but $\kappa_1-a(\ol a\ol b) = 0$ and $\kappa_2-a(\ol a\ol b) = 1\neq 0$. So due to \FARBE{Proposition~\ref{prop:asp-postulates-ocf-characterization}} c-contractions do not fulfill \propWE.
}			
			\end{example}

\begin{figure}
	\centering
	\begin{ocftable}{4}
		\ocf{$\kappa_1$}{
		    {},
			{$\overline{a} \overline{b}$},
			{${a} \overline{b},\: \overline{a} b$},
			{$a b$}
		}
		\ocf{$\kappa_2$}{
			{$\overline{a} \overline{b}$},
			{},
			{${a} \overline{b},\: \overline{a} b$},
			{$a b$}
		}
		\ocf{$\kappa_1 - b$}{
			{},
			{},
			{$\overline{a} {b},\:\overline{a} \overline{b}$},
			{$a b,\:{a} \overline{b}$}
		}
		\ocf{$\kappa_2 - b$}{
			{},
			{$\overline{a} \overline{b}$},
			{$\overline{a} {b}$},
			{$a b,\:{a} \overline{b}$}
		}
	\end{ocftable}
	\caption{Counterexample to \propE for c-ignorations and minimal c-contractions for Example~\ref{ex_E_min}}
	\label{tab:E_min-c-con}
\end{figure}

\begin{example}[\propE; ignoration, minimal]
    \label{ex_E_min}
    \FARBE{
	Let $ \kappa_1 $, $ \kappa_2 $ be ranking functions %
	as given in Figure \ref{tab:E_min-c-con}. 
	$ \kappa_2 $ differs from $ \kappa_1 $ by having a different rank for $ \ol{a}\ol{b} $: $ \kappa_2(\ol{a}\ol{b})=3 $.				We c-contract $\kappa_1$ and $\kappa_2$ with $A=b$ and $\gamma = 1$. Since $\gamma = \kappa_1(\ol b)-\kappa_1(b) = \kappa_2(\ol b)-\kappa_2(b)$ and $\gamma=\kappa_1(\ol b)=\kappa_2 (\ol b)$ both c-contractions are ignorations and minimal. The results of the c-contractions are also shown in Figure \ref{tab:E_min-c-con}. Observe that $\kappa_1$ and $\kappa_2$ are equivalent so $C(\kappa_1)=C(\kappa_2)$, but $\kappa_1-b$ and $\kappa_2-b$ are not equivalent, since $(b|\ol a)\in C(\kappa_2-b)$ but $(b|\ol a)\not\in C(\kappa_1-b)$. So due to \FARBE{Proposition~\ref{prop:asp-postulates-ocf-characterization}} c-contractions do not fulfill \propE.
	}
\end{example}

\begin{example}[\propWE; revocation; non-minimal]
\label{ex_wE_nonmin}
Non-minimal c-contractions do not fulfil \propWE. In Figure~\ref{tab:ce_we_non_min_c_contr} we have two ranking functions, $\kappa_1, \kappa_2$, with the same worlds in the lowermost layer. We c-contract $\kappa_1$ and $\kappa_2$ with $A=a$ and $\gamma=4$ so we have two revocations and two non-minimal c-contractions ($\kappa_1(\ol A) = 2,\, \kappa_2(\ol A)=1$). Because $\kappa_1-a(\ol ab) = 0$ but $\kappa_2(\ol ab) = 1$ the lowermost layers of the c-contractions are not equal and due to \FARBE{Proposition~\ref{prop:asp-postulates-ocf-characterization}} c-contractions do not fulfill \propWE.
\end{example}

\begin{figure}
	\centering
	\begin{ocftable}{5}
		\ocf{$\kappa_1$}{
		    {},
		    {$\overline{a} \overline{b}$},
			{$\overline{a} b$},
			{$a \overline{b}$},
			{$a b$}
		}
		\ocf{$\kappa_2$}{
			{},
			{},
			{$\overline{a} b$},
			{$a \overline{b}, \: \overline{a} \overline{b}$},
			{$a b$}
		}
		\ocf{$\kappa_1 - a$}{
		    {},
			{$a \overline{b}$},
			{$a b$},
			{$\overline{a} \overline{b}$},
			{$\overline{a} b$}
		}
		\ocf{$\kappa_2 - a$}{
			{$a \overline{b}$},
			{$a b$},
			{},
			{$\overline{a} b$},
			{$\overline{a} \overline{b}$}
			}
	\end{ocftable}
	\caption{Counterexample to \propWE for c-revocations and non minimal c-contractions for Example~\ref{ex_wE_nonmin}}
	\label{tab:ce_we_non_min_c_contr}
\end{figure}

\begin{figure}
	\centering
	\begin{ocftable}{4}
		\ocf{$\kappa$}{
			{$\overline{a} b \overline{c}$},
			{$\overline{a} b c,\: \overline{a} \overline{b} c,\: \overline{a} \overline{b} \overline{c}$},
			{$a b c $},
			{$a b \overline{c},\: a \overline{b} c,\: a \overline{b} \overline{c}$}
		}
		\ocf{$\kappa - a$}{
			{$a b c$},
			{$a b \overline{c},\: a \overline{b} c,\: a \overline{b} \overline{c}$},
			{$\overline{a} b \overline{c}$},
			{$\overline{a} b c,\: \overline{a} \overline{b} c,\: \overline{a} \overline{b} \overline{c}$}
		}
		\ocf{$\margsub{\kappa}{\{b,c\}}$}{
			{},
			{},
			{$b c$},
			{$b \overline{c},\: \overline{b} c,\: \overline{b} \overline{c}$}
		}
		\ocf{$\margsub{(\kappa - a)}{\{b,c\}}$}{
			{},
			{},
			{$b \overline{c}$},
			{$bc,\: \overline{b} c,\: \overline{b} \overline{c}$}
		}
	\end{ocftable}
	\caption{Counterexample to \propPP and \propNP for c-revocations and non-minimal c-contractions for Example~\ref{ex_PP-NP_revoc_nonmin}}
	\label{fig:ex_PP-NP_revoc_nonmin}
\end{figure}

\begin{example}[\propPP, \propNP; revocation, non-minimal]
	\label{ex_PP-NP_revoc_nonmin}
    \FARBE{
	Let $\kappa$ be a ranking function over $\Sigma=\{a,b,c\}$ as defined in Figure~\ref{fig:ex_PP-NP_revoc_nonmin}. Using $\gamma = 4 > \kappa(\overline{a}) - \kappa(a) = 2 - 0 = 2$, we obtain $\kappa-a$, which is both a c-revocation of $a$ and a non-minimal c-contraction with $a$.\\%
	Now consider the marginalized ranking functions $\margsub{\kappa}{\{b,c\}}$ and $\margsub{(\kappa - a)}{\{b,c\}}$ over $\Sigma'=\{b,c\} = \Sigma \setminus \signatureOf{a}$. According to Proposition~\ref{prop_basics_marginalization_conditionalization}, $\kappa \nmmodels \margsub{\kappa}{\{b,c\}}$ and $(\kappa - a) \nmmodels \margsub{(\kappa - a)}{\{b,c\}}$.\\%
	For \propPP, let $\kappa' = \margsub{\kappa}{\{b,c\}}$. Note that $\kappa' \models (\overline{c}|b)$, because $\kappa'(b \overline{c}) = 0 < 1 = \kappa'(b c)$. However, $(\kappa - a) \not\models (\overline{c}|b)$ due to $(\kappa - a)(bc) = 0 < 1 = (\kappa - a)(b \overline{c})$, which means that $(\kappa - a) \notnmmodels \kappa'$. Hence, \propPP is not fulfilled.\\%
    For \propNP, let $\kappa' = \margsub{(\kappa - a)}{\{b,c\}}$. Note that $\kappa' \models (c|b)$, because $\kappa'(bc) = 0 < 1 = \kappa'(b \overline{c})$. However, $\kappa \not\models (c|b)$ due to $\kappa(b \overline{c}) = 0 < 1 = \kappa(b c)$. This means we have $\kappa \notnmmodels \kappa'$, but $(\kappa - a) \nmmodels \kappa'$. Hence, \propNP is not fulfilled.
    }
\end{example}

\begin{example}[\propNP does not imply \propW; used in Proposition~\ref{prop_npw}]
 \label{ex_np_not_implies_w}

\begin{figure}
	\centering
	\begin{ocftable}{4}
		\ocf{$\kappa$}{
			{$\overline{a} b \overline{c},\: \overline{a} \overline{b} c$},
			{$\overline{a} b c,\: \overline{a} \overline{b} \overline{c}$},
			{$a b \overline{c},\: a \overline{b} c$},
			{$a b c,\: a \overline{b} \overline{c}$}
		}
		\ocf{$\kappa - a$}{
			{$a b \overline{c},\: a \overline{b} c$},
			{$a b c,\: a \overline{b} \overline{c}$},
			{$\overline{a} b \overline{c},\: \overline{a} \overline{b} c$},
			{$\overline{a} b c,\: \overline{a} \overline{b} \overline{c}$}
		}
		\ocf{$\kappa'$}{
			{},
			{},
			{$b c,\: \overline{b} \overline{c}$},
			{$b \overline{c},\: \overline{b} c$}
		}
	\end{ocftable}
	\caption{Counterexample showing that \propNP does not imply \propW for Example~\ref{ex_np_not_implies_w}}
	\label{fig:ex_prop-np-w}
\end{figure}

Let $\kappa$ be a ranking function over $\Sigma = \{a,b,c\}$ as defined in Figure~\ref{fig:ex_prop-np-w}. We obtain the c-revocation $(\kappa - a)$ by using $\gamma = 4 > \kappa(\overline{a}) - \kappa(a) = 2$. Note that $\kappa \models a$ because $\kappa(a) < \kappa(\overline{a})$, while $(\kappa - a) \models \overline{a}$ because $(\kappa -a)(\overline{a}) < (\kappa-a)(a)$. Therefore, $\kappa \notnmmodels (\kappa - a)$, i.e.\ \propW does not hold.

Now consider the ranking function $\kappa'$ over $\Sigma' = \{b,c\} = \Sigma \setminus \signatureOf{a}$ in Figure~\ref{fig:ex_prop-np-w}. Since $\kappa(bc) = (\kappa-a)(bc) = 0 < 1 = \kappa(b \overline{c}) = (\kappa-a)(b \overline{c})$, both $\kappa \models (c|b)$ and $(\kappa - a) \models (c|b)$. However, $\kappa' \models (\overline{c}|b)$ because $\kappa'(b \overline{c}) = 0 < 1 = \kappa'(bc)$. Therefore, $\kappa \notnmmodels \kappa'$ and $(\kappa - a) \notnmmodels \kappa'$, which means that \propNP is fulfilled in this scenario.
\end{example}

\end{document}